%% file: main.tex
\documentclass[10pt,journal,compsoc]{IEEEtran}
\ifCLASSOPTIONcompsoc
  \usepackage[nocompress]{cite}
\else
  \usepackage{cite}
\fi
\ifCLASSINFOpdf
\else
\fi

\input{math_commands.tex}
\usepackage{url}
\usepackage{hyperref}

\usepackage{booktabs}       
\usepackage{amsfonts}       
\usepackage{nicefrac}       
\usepackage{microtype}      
\usepackage{xcolor}         
\usepackage{colortbl}
\definecolor{lightcyan}{rgb}{0.88,1,1}

\usepackage{amsmath}
\usepackage{amssymb}
\usepackage{amsthm}
\usepackage{dsfont}
\usepackage{mathtools}

\usepackage{algorithm,algorithmicx}
\usepackage{algpseudocode}

\usepackage[noEnd=false,indLines=true]{algpseudocodex}

\usepackage{wrapfig}

\usepackage{multirow}

\usepackage{subcaption}

\usepackage{enumitem}
\usepackage{boldline} 
\usepackage{tabularx} 
\usepackage{tabulary} 
\usepackage{multirow} 
\usepackage{diagbox}  
\usepackage{makecell} 
\usepackage{relsize}
\usepackage[normalem]{ulem}
\usepackage{textcomp}

\newcolumntype{C}{>{\centering\arraybackslash}X}
\newcolumntype{L}{>{\raggedright\arraybackslash}X}
\newcolumntype{R}{>{\raggedleft\arraybackslash}X}
\newcolumntype{J}{>{\justifying\arraybackslash}X}
\newcolumntype{P}[1]{>{\centering\arraybackslash}p{#1}}
\newcolumntype{M}[1]{>{\centering\arraybackslash}m{#1}}
\newcolumntype{B}[1]{>{\centering\arraybackslash}b{#1}}

\theoremstyle{plain}
\newtheorem{theorem}{Theorem}[section]

\newtheorem{definition}[theorem]{Definition}

\theoremstyle{remark}
\newtheorem{remark}[theorem]{Remark}

\begin{document}
%
\title{Reinforced Embodied Active Defense: \\ Exploiting Adaptive Interaction for Robust Visual Perception in Adversarial 3D Environments}
%
%
%

\author{Xiao~Yang,
        Lingxuan~Wu,
        Lizhong~Wang,
        Chengyang~Ying, 
        Hang~Su, 
        and~Jun~Zhu,~\IEEEmembership{Fellow,~IEEE}
        \IEEEcompsocitemizethanks{\IEEEcompsocthanksitem X. Yang, L. Wu, L. Wang, C. Ying, H. Su and J. Zhu are with Dept. of Comp. Sci. \& Tech., Institute for AI, BNRist Center, THBI Lab, Tsinghua-Bosch Joint Center for ML, Tsinghua University, Beijing, China. 
        Email: yangxiao19@tsinghua.org.cn, \{wlx23, ycy21\}@mails.tsinghua.edu.cn, wanglizhong99@outlook.com, \{suhangss, dcszj\}@tsinghua.edu.cn. 
        Xiao Yang and Lingxuan Wu had contributed equally to this work. Corresponding authors: Hang Su; Jun Zhu. Code is available at \url{https://github.com/thu-ml/EmbodiedActiveDefense}.
        }
        }
\markboth{IEEE TRANSACTIONS ON PATTERN ANALYSIS AND MACHINE INTELLIGENCE}%
{Shell \MakeLowercase{\textit{et al.}}: A Sample Article Using IEEEtran.cls for IEEE Journals}
%



\IEEEtitleabstractindextext{%
\begin{abstract}
Adversarial attacks in 3D environments have emerged as a critical threat to the reliability of visual perception systems, particularly in safety-sensitive applications such as identity verification and autonomous driving. These attacks employ adversarial patches and 3D objects to manipulate deep neural network (DNN) predictions by exploiting vulnerabilities within complex scenes. Existing defense mechanisms, such as adversarial training and purification, primarily employ passive strategies to enhance robustness. However, these approaches often rely on pre-defined assumptions about adversarial tactics, limiting their adaptability in dynamic 3D settings. To address these challenges, we introduce \textbf{Reinforced Embodied Active Defense (\textsc{Rein}-EAD)}, a proactive defense framework that leverages adaptive exploration and interaction with the environment to improve perception robustness in 3D adversarial contexts. By implementing a multi-step objective that balances immediate prediction accuracy with predictive entropy minimization, \textsc{Rein}-EAD optimizes defense strategies over a multi-step horizon. Additionally, \textsc{Rein}-EAD involves an uncertainty-oriented reward-shaping mechanism that facilitates efficient policy updates, thereby reducing computational overhead and supporting real-world applicability without the need for differentiable environments. Comprehensive experiments validate the effectiveness of \textsc{Rein}-EAD, demonstrating a substantial reduction in attack success rates while preserving standard accuracy across diverse tasks. Notably, \textsc{Rein}-EAD exhibits robust generalization to unseen and adaptive attacks, making it suitable for real-world complex tasks, including 3D object classification, face recognition and autonomous driving. By integrating proactive policy learning with embodied scene interaction, \textsc{Rein}-EAD establishes a scalable and adaptable approach for securing DNN-based perception systems in dynamic and adversarial 3D environments. 


\end{abstract}

}

\maketitle

\IEEEdisplaynontitleabstractindextext

%
\IEEEpeerreviewmaketitle

\IEEEraisesectionheading{\section{Introduction}\label{sec:introduction}}

\IEEEPARstart{A}{dversarial} attacks in 3D environments have become a significant threat to the security and reliability of visual perception systems~\cite{brown2017adversarial,sharif2016accessorize}. These attacks utilize carefully crafted perturbations, such as adversarial patches and 3D objects, strategically placed in physical scenes to manipulate deep neural network (DNN) predictions~\cite{sharif2016accessorize,zhu2023understanding}.  The consequences of such vulnerabilities are especially severe in safety-critical domains, such as identity verification~\cite{sharif2016accessorize,xiao2021improving,yang2024face3dadv} and autonomous driving~\cite{song2018physical,zhu2023understanding}, where erroneous predictions can severely compromise system integrity. Therefore, ensuring robust perception under adversarial conditions is critical for deploying these systems reliably in real-world applications. 


In response to these emergent threats, researchers have developed diverse defense strategies for enhancing the robustness of DNNs. Adversarial training \cite{madry2017towards,wu2019defending,rao2020adversarial} has emerged as a particularly effective approach \cite{gowal2021improving}, where adversarial examples are deliberately incorporated into the training data to enhance the model's resilience. Concurrently, input preprocessing techniques, such as adversarial purification, have been proposed to mitigate these perturbations \cite{xiang2021patchguard,liu2022segment,xu2023patchzero}.
However, these approaches predominantly belong to \textbf{passive defenses}, which exhibit vulnerability to unseen or adaptive attacks \cite{athalye2018obfuscated,tramer2020adaptive} that circumvent existing robustness measures, due to the presuppositions regarding the adversary's approaches. Moreover, these strategies often neglect the intrinsic physical context and associated understanding of the scene and objects in the 3D realm, weakening these defenses in real-world physical environments.

In contrast to the limitations of passive defenses, human active vision employs iterative refinement and error-correction mechanisms to effortlessly detect misplaced or inconsistent elements in complex 3D environments~\cite{thomas1999theories,elsayed2018adversarial}. Inspired by human active vision, our recent seminal work proposed a novel defense framework of {Embodied Active Defense (EAD)}, which incorporates a proactive policy network to replicate this dynamic process~\cite{wuembodied}. 
EAD actively contextualizes environmental information and leverages object consistency to address misaligned adversarial patches in 3D settings. The system continuously refines its scene understanding by integrating current and past observations, forming a more comprehensive representation of the environment. This proactive behavior enables the system to predict areas of uncertainty and adjust its actions accordingly, thereby improving the quality of the observations it collects. By synergizing proactive movement and iterative predictions, EAD enhances its scene comprehension and mitigates the impact of adversarial patches. 

Despite its potential, the current EAD framework faces several critical challenges that limit its effectiveness and applicability in real-world scenarios. First, EAD’s greedy informative exploration strategy prioritizes immediate, single-step information gain over long-term relevance, resulting in temporally inconsistent actions. This short-sighted approach often leads the agent to revisit previously explored viewpoints, diminishing exploration efficiency and increasing the likelihood of erroneous predictions. Furthermore, EAD’s reliance on training the proactive policy network with differentiable environment models introduces a misalignment, particularly when handling non-differentiable real-world physical dynamics, which limits the framework’s practical applicability. Additionally, learning through differentiable simulations is computationally intensive and susceptible to numerical instabilities, further undermining the system’s overall efficiency.


To address existing limitations and enhance the effectiveness, applicability, and efficiency of defense mechanisms, we propose the Reinforced EAD (\textsc{Rein}-EAD) framework. This framework enables agents to optimize for long-term outcomes by learning through trial-and-error interactions without requiring differentiable simulations. To address temporal inconsistencies in exploration, we introduce a generalized objective function that accumulates multi-step interactions. This objective balances prediction loss reduction with predictive entropy minimization over a multi-step horizon, allowing the system to account for temporal dependencies and prioritize long-term outcomes over immediate uncertainty reduction. Additionally, to eliminate differentiable constraints and ensure efficient convergence, we implement an uncertainty-oriented reward-shaping technique within reinforcement learning. This approach provides dense rewards at each step, guiding the agent to reduce perceptual uncertainty and minimize prediction errors. By fostering efficient updates, this reward structure supports stable convergence even in complex environments, enhancing the agent’s adaptability to dynamic changes and improving overall performance in uncertain and evolving scenarios. Finally, to address overfitting and computational burden of adversarial patch generation, we present Offline Adversarial Patch Approximation (OAPA) for generating adversary-agnostic patches. OAPA systematically characterizes the manifold of adversarial patterns from diverse attack strategies. By distilling fundamental features across attack strategies offline, OAPA substantially improves generalization across diverse attacks while reducing the computational overhead of online adversarial training.

Extensive experiments demonstrate that \textsc{Rein}-EAD offers prominent advantages over conventional passive defense mechanisms. First, \textsc{Rein}-EAD consistently outperforms state-of-the-art defense techniques, achieving a remarkable 95\% reduction in attack success rate across a diverse range of tasks. Notably, \textsc{Rein}-EAD maintains or even enhances standard accuracy by effectively leveraging instructive information suited for detecting target objects in dynamic 3D environments. Second, \textsc{Rein}-EAD demonstrates superior \textbf{generalization} compared to passive approaches. Its attack-agnostic strategies enable it to defend effectively against a broad spectrum of adversarial patches, including both unseen and adaptive attacks. Moreover, \textsc{Rein}-EAD exhibits strong \textbf{applicability} in complex and real-world scenarios, such as 3D object classification, face recognition and object detection for autonomous driving. The trial-and-error learning paradigm ensures stable and efficient policy updates, making \textsc{Rein}-EAD highly adaptable to real-world tasks.

To summarize, our contributions are as follows:
\begin{itemize}
\item First, we propose \textsc{Rein}-EAD that integrates multi-step accumulative interactions and policy learning into a cohesive framework. It optimizes a multi-step objective with an uncertainty-oriented reward shaping, promoting temporally consistent and informative exploration.

\item Second, we develop an adversary-agnostic defense strategy of  OAPA, which enables \textsc{Rein}-EAD to defend against diverse adversarial patches, including unseen and adaptive attacks, without relying on specific assumptions about the adversary's capabilities. 

\item Third, through extensive experiments, we demonstrate  \textsc{Rein}-EAD's superior \textbf{effectiveness} over state-of-the-art passive defenses across various settings, strong \textbf{generalization} against various unseen and adaptive attacks, and \textbf{adaptability} to complex real-world scenarios.
\end{itemize}

\section{Related Work}
In this section, we delve into the threat posed by adversarial patches in 3D environments and explore the corresponding defensive strategies.

\subsection{Adversarial Patches and Defenses}

Adversarial patches, initially devised to manipulate specific regions of an image to mislead image classifiers~\cite{brown2017adversarial}, have significantly evolved. They now deceive a broad spectrum of perception models~\cite{sharif2016accessorize,song2018physical}, including those operating within 3D environments~\cite{eykholt2018robust,zhu2023understanding,yang2023towards,yang2024face3dadv}.

To defend perception models against such adversarial patch attacks, various strategies have been proposed, ranging from empirical defenses~\cite{dziugaite2016study,naseer2019local,rao2020adversarial,xu2023patchzero} to certified defenses~\cite{zhang2019towards,xiang2021patchguard}. However, a critical review reveals that most contemporary defense mechanisms fall under what we term as \textbf{passive defenses}. These approaches rely on information derived from monocular observations and presupposed adversarial tactics to mitigate patch-based threats.

Within the passive defense paradigm, two primary approaches have gained prominence. Adversarial training~\cite{madry2017towards,wu2019defending,rao2020adversarial} strengthens the perception model by exposing it to adversarial examples during training, {boosting intrinsic robustness against adversarial perturbations in supervised learning~\cite{liu2021probabilistic,yu2022understanding} and semi-supervised learning~\cite{li2024dynamic}}. Alternatively, adversarial purification~\cite{naseer2019local,liu2022segment,xu2023patchzero} integrates an auxiliary purifier into the perception pipeline. This purifier first identifies adversarial patches within observations and then neutralizes or removes these perturbations. The amended observations are subsequently processed by the model, yielding a two-stage defense pipeline that mitigates adversarial influences before perception tasks commence.

While passive defenses have shown success in specific scenarios, their dependence on static, pre-defined mechanisms renders them vulnerable to adaptive attacks and restricts their effectiveness in dynamic, real-world environments. In contrast, active defense strategies exhibit a more adaptable approach, dynamically responding to evolving adversarial threats in complex 3D settings.

Our recent work introduces a pioneering defense framework termed {Embodied Active Defense (EAD)}, which employs a proactive policy network to simulate this dynamic response process~\cite{wuembodied}. EAD actively contextualizes environmental information and leverages object consistency to address adversarial patch misalignments in 3D environments, marking a significant advancement in adversarial defense.

\subsection{Embodied Perception} 
The paradigm of embodied perception \cite{aloimonos1988active,bajcsy1988active} represents a significant shift in the landscape of artificial intelligence and computer vision. This approach posits that perception is not merely a passive process of information reception, but rather an active, embodied experience where an agent can dynamically interact with and navigate its environment to optimize perceptual outcomes and enhance task performance.

The embodied perception framework has demonstrated remarkable versatility, finding applications across a diverse spectrum of computer vision tasks. In the domain of object detection, researchers have leveraged embodied agents to actively explore and analyze scenes, leading to more robust and context-aware detection systems~\cite{kotar2022interactron,ruan2023improving}. Similarly, in the field of 3D pose estimation, embodied approaches have enabled more accurate and adaptable solutions by allowing agents to actively seek optimal viewpoints for estimation~\cite{ci2023proactive}. Furthermore, the paradigm has proven invaluable in advancing 3D scene understanding, where embodied agents can navigate complex environments to build comprehensive spatial representations~\cite{ma2022sqa3d}. 

The power of embodied perception lies in its ability to mimic the active, exploratory nature of biological vision systems, allowing agents to overcome the limitations of static and passive perception. By dynamically adjusting their position, orientation, or focus in response to environmental cues and task demands, embodied agents can potentially gather more informative and less ambiguous sensory data, leading to improved performance across different perceptual tasks. Our novel integration of embodied perception with adversarial robustness opens up new avenues for research, potentially leading to more resilient and adaptable perception systems capable of maintaining high performance even in sophisticated adversarial attacks.

\section{Methodology}

\begin{figure*}
    \centering
    \includegraphics[width=0.99\linewidth]{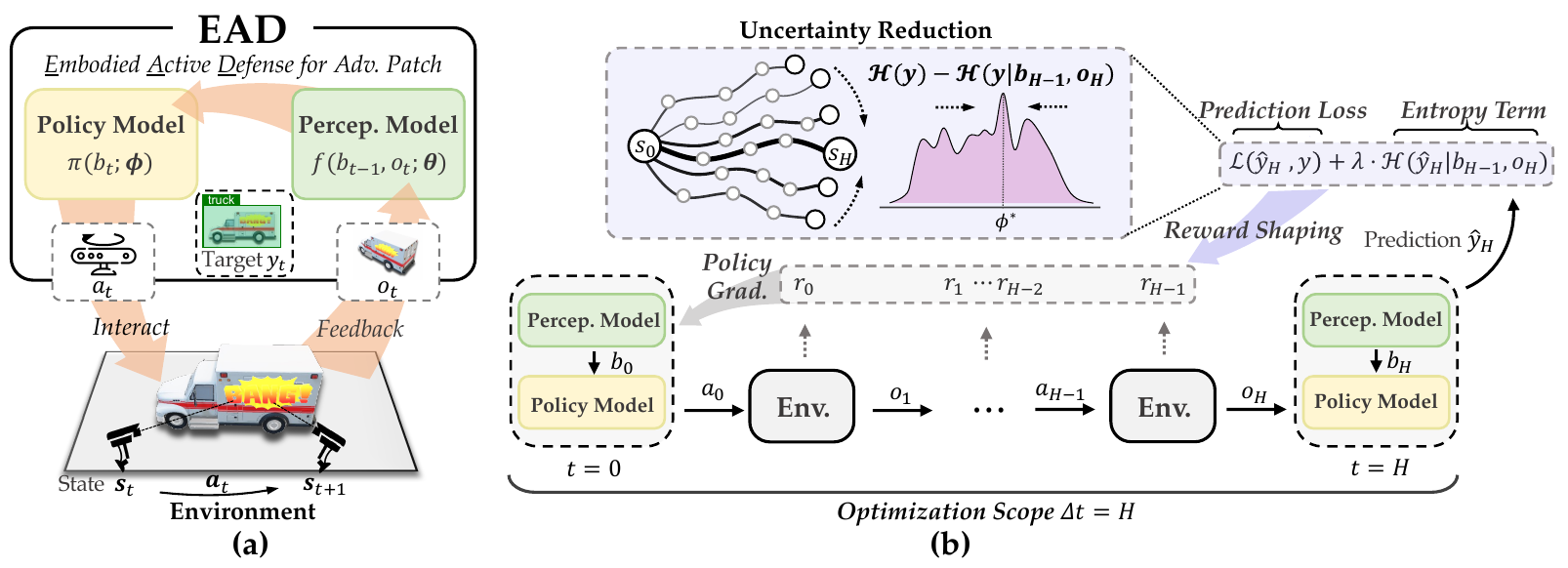}
     \vspace{-1em}
    \caption{An overview of EAD and \textsc{Rein}-EAD. (a) In EAD, the perception model refines the environment representation $b_t$ using observation $o_t$ and previous internal belief $b_{t-1}$, making task-specific prediction $y_t$. The policy model generates action $a_t$ based on $b_t$, minimizing perception uncertainty $H(y \mid b_{t-1}, o_t)$ over a single step. (b) \textsc{Rein}-EAD extends EAD by accumulating multi-step interactions, balancing prediction loss reduction and entropy minimization over horizon $H$. The policy is learned using model-free RL with dense rewards at each step, guiding the agent toward robust decision-making.}
    \label{fig:framework}
    \vspace{-1em}
\end{figure*}

\label{sec:ead}
We first introduce the Preliminary about Embodied Active Defense (EAD) in Sec.~\ref{subsec:ead}, which leverages recurrent feedback to counteract adversarial patches. Then, we provide a theoretical analysis of EAD in Sec.~\ref{subsec:uncertainty_analysis} for further exploration. 
In Sec.~\ref{subsec:reinforce-ead}, we propose \textsc{Rein}-EAD that incorporates multi-step interactions and policy learning. It effectively enhances the adaptability and resilience of the defense mechanism in complex and real-world environments.
Finally, we provide the adversary-agnostic defenses in Sec.~\ref{subsec:impl_tech}.

\subsection{Preliminary: Embodied Active Defense}
\label{subsec:ead}
Consider a scene $x \in \mathcal{X}$ with its associated ground-truth label $y \in \mathcal{Y}$. The perception model $f:\mathcal{O}\rightarrow \mathcal{Y}$ aims to predict the scene annotation $y$ based on the image observation $o_i \in \mathcal{O}$, where $o_i$ is derived from the scene $x$ and conditioned on the camera's state $s_i$ (encompassing camera's position and viewpoint). The function $\mathcal{L}(\cdot)$ represents a task-specific loss function, such as the cross-entropy loss.

Traditional passive defense strategies operate a single observation $o_i$ to counter adversarial patches, thus failing to capitalize on the rich contextual information obtainable through proactive exploration of the environment~\cite{ronneberger2015u,he2017mask}.
Formally, an adversarial patch $p$ is introduced into the observation $o_i$, resulting in the erroneous prediction of the perception model $f$. The generation of adversarial patches in 3D scenes~\cite{zhu2023understanding} typically optimizes:
\begin{equation}
\label{eqn:adv_objective}
\begin{aligned}
\max_{p} \mathbb{E}_{s_i} \mathcal{L}(f(A(p, o_i; s_i)), y), 
\end{aligned}
\end{equation}
where $A(\cdot)$ projects the 3D adversarial patch $p$ to the 2D camera observations $o_i$ according to the camera's state $s_i$.
This generalized 3D formulation encompasses the 2D case when $s_i$ is restricted to 2D transformations and patch locations, as in Brown~\etal~\cite{brown2017adversarial}.

For subsequent analysis, we define the set of deceptive adversarial patches $\mathcal{P}_{x}$ for scene $x$ as:
\begin{equation}
\label{eqn:adv_patch_set}
\mathcal{P}_{x}=\{ p \in [0,1]^{H_p\times W_p \times C}:  \mathbb{E}_{s_i}f(A(p,o_i;s_i))\neq y\},
\end{equation}
where $H_p$ and $W_p$ represent the height and width of the patch, respectively. In practice, approximating the solution set $\mathcal{P}_{x}$ involves employing specific optimization techniques \cite{madry2017towards,carlini2017towards} to solve the problem presented in Eq. (\ref{eqn:adv_objective}). This generalized formulation provides a robust foundation for analyzing the behavior and impact of adversarial patches in complex 3D environments. 

\noindent \textbf{The EAD framework}. Our recent work proposes EAD~\cite{wuembodied}, a paradigm that champions active scene engagement and iteratively leverages environmental feedback to enhance the robustness of perception systems against patch attacks. 
EAD comprises two recurrent models that emulate the intricate cerebral structure underpinning active human vision. The \textbf{perception model} $f(\cdot;\vtheta)$, parameterized by $\vtheta$, is meticulously crafted to facilitate sophisticated visual perception by fully harnessing the rich contextual information embedded within temporal observations from the external world. At each timestep $t$, the model ingeniously leverages the current observation $o_t$ and amalgamates it with the prevailing internal belief $b_{t-1}$ regarding the scene, thus constructing an enhanced representation of the surrounding environment $b_t$ by a recurrent paradigm. Simultaneously, the perception model generates a scene annotation $\hat{y}_t$ as:
\begin{equation}
     \{\hat{y}_t, b_{t}\} = f(o_t, b_{t-1}; \vtheta)\footnote{we set $f_y(o_t, b_{t-1}; \vtheta)=\hat{y}_t$ and $f_b(o_t, b_{t-1}; \vtheta)= b_{t}$ respectively.}.
\end{equation}
The subsequent \textbf{policy model} $\pi(\cdot;\vphi)$, parameterized by $\vphi$, serves to govern the visual control of movement. Formally, given the current collective environmental understanding $b_t$ meticulously sustained by the perception model, it derives the action $\va_t$  by sampling from the distribution $\pi(b_t; \vphi)$.

To formally characterize the EAD framework's interaction and proactive exploration within the environment (as illustrated in Fig.~\ref{fig:framework}), we extend the framework of the Partially-Observable Markov Decision Process (POMDP)~\cite{cassandra1994acting}. The interaction process under the scene $x$ is denoted by $\mathcal{M}(x) \coloneqq \langle \mathcal{S},\mathcal{A}, \mathcal{T}, \mathcal{O}, \mathcal{Z}\rangle$. Here, $\mathcal{S}$ and $\mathcal{A}$ represent the state and action spaces, respectively. For $\forall (s,a)\in \mathcal{S}\times\mathcal{A}$, the transition dynamic under the scene $x$ adheres to the Markovian property, satisfying $\mathcal{T}(\cdot\mid s, a, x)$. Due to the partially observed nature of the environment, the agent can not directly access the state $s$ but instead receives an observation $o$ sampled from the observation function $\mathcal{Z}(\cdot \mid s, x)$.
At each timestep $t$, EAD obtains an observation $o_t$ based on the current state $s_t$. This observation $o_t$ serves as crucial environmental feedback, enabling the refinement of the agent's understanding of the environment $b_t$ through the sophisticated perception model $f(\cdot;\vtheta)$. The recurrent perception mechanism employed by EAD is important for maintaining the stability of human vision \cite{thomas1999theories,kar2019evidence}. Rather than remaining static and passively assimilating observations, EAD leverages the policy model to execute actions $a_t$ sampled from the distribution $\pi(b_t;\vphi)$. The incorporation of the policy model enables EAD to determine the optimal action at each timestep, ensuring the acquisition of the most informative feedback from the scene.

\noindent \textbf{Training EAD against adversarial patches.} 
To equip the intricately constructed EAD model, we introduce a specialized learning algorithm designed for countering adversarial patches. Considering a data distribution $\mathcal{D}$ comprising paired data $(x, y)$, we examine adversarial patches $p \in \gP_{x}$ generated using \textbf{unknown attack techniques} that corrupt the observation $o_t$ into $o_t' = A(o_t,p;s_t)$.
The primary objective of EAD is to minimize the expected loss in the presence of adversarial patch threats. Consequently, the learning process of EAD to mitigate adversarial patches is formulated as an optimization problem of parameters $\vtheta$ and $\vphi$ as:
\begin{equation}
\label{eqn:objective}
\begin{gathered}    
    \min_{\vtheta,\vphi} \; \mathbb{E}_{(x,y)\sim \mathcal{D}, \tau \sim (\mathcal{M}(x),\pi), t} \Big[\sum_{p \in \mathcal{P}_{x}} \mathcal{L}(\hat{y}_t, y) \Big] ,\\ 
    \textrm{with}\; \{\hat{y}_t, b_{t}\} = f(A(o_t,p;s_t), b_{t-1};\vtheta), \; a_{t} \sim \pi(\cdot\mid b_{t}; \vphi)\\
    \textrm{s.t.} \quad o_{t} \sim \mathcal{Z}(\cdot\mid s_t, x), \quad s_{t} \sim \mathcal{T}(\cdot\mid s_{t-1}, a_{t-1}, x),
\end{gathered}
\end{equation}
where $\tau\coloneqq (o_0, a_0, o_1,\ldots, o_H)$ represents the collected trajectory with length $H$ and the probability $p_{\mathcal{M} (x),\pi}(\tau) = \rho_0(s_0)\prod_{t=0}^{H}\mathcal{T}(s_{t+1}\mid s_{t},a_t,x)\pi(a_t \mid b_t; \vphi)\mathcal{Z}(o_t \mid s_t)$;
while $\hat{y}_t$ signifies the model's prediction at timestep $t$, which adheres to a uniform distribution over $\{0, 1, 2, ..., H\}$.
Remarkably, the loss function $\mathcal{L}$ exhibits a task-agnostic nature, emphasizing the exceptional adaptability of the EAD framework. This inherent flexibility guarantees that EAD delivers robust defenses across a wide spectrum of perception tasks. 

\subsection{Theoretical Analysis of EAD}
\label{subsec:uncertainty_analysis}

To gain a deeper understanding of the model's behavior, we further examine a generalized instance of EAD in Eq. (\ref{eqn:objective}), where the agent employs the InfoNCE objective~\cite{oord2018representation}, simplified as:
\begin{equation}
\label{eqn:infonce_obj}
    \min_{\vtheta,\vphi} \; \mathbb{E}_{(x^{(j)},y^{(j)})} \left[ \frac{1}{K}\sum_{j=1}^K \log \frac{e^{- S(f_y(b_{t-1}^{(j)}, o_{t}^{(j)}; \vtheta), y^{(j)})}}{\displaystyle \frac{1}{K}\sum_{k=1}^K e^{- S(f_y(b_{t-1}^{(j)}, o_{t}^{(j)}; \vtheta), y^{(k)})}}\right],
\end{equation}
where ${(x^{(j)},y^{(j)})}_{j=1}^{K}$ denotes a data batch of size $K$ sampled from the distribution $ \mathcal{D}$, and $S: \mathcal{Y} \times \mathcal{Y} \rightarrow \mathbb{R}$ quantifies the similarity between the predicted scene annotation and the ground truth label. In embodied perception, this loss establishes a cross-modal correspondence between the observations and annotations like CLIP \cite{radford2021learning}. Moreover, we provide an information-theoretic interpretation of Eq.~(\ref{eqn:infonce_obj}) to elucidate its underlying principles.

\begin{theorem}[Proof in Appendix {A.1}]
    \label{thrm:eq_to_max_mutual_info}
   For mutual information between current observation $o_t$ and scene annotation $y$ conditioned on previous belief $b_{t-1}$, denoted as $I(o_t;y \mid b_{t-1})$, we can prove that our objective is a lower bound of this mutual information :
    \begin{equation}
        \begin{aligned}
         \label{eqn:objective_greedy_exploration}
        &\mathbb{E}_{(x^{(j)},y^{(j)})\sim \mathcal{D}} 
        \left[\frac{1}{K} \sum_{j=1}^K \log \frac{q_{\vtheta}(y^{(j)} \mid  b_{t-1}^{(j)}, o_{t}^{(j)})}{\frac{1}{K}\sum_{k=1}^K  q_{\vtheta}({y^{(k)}} \mid b_{t-1}^{(j)}, o_{t}^{(j)})}\right] \\ 
        \leq  & \mathbb{E}_{x} I(o_t;y \mid b_{t-1}) - \frac{\log(K)}{K} \\
        = & \mathbb{E}_{x} \left[\mathcal{H}(y \mid b_{t-1}) - \mathcal{H}(y \mid b_{t-1},o_t)\right] - \frac{\log(K)}{K},
        \end{aligned}
    \end{equation}
   where $q_{\vtheta}(y \mid  o_{1}, \cdots,  o_{t})$ represents the variational distribution approximating the true conditional distribution $p(y \mid o_1, \cdots, o_{t})$ with samples $(x^{(j)}, y^{(j)})_{j=1}^{K}$.
\end{theorem}
\begin{remark}
To elucidate the connection between the derived lower bound on conditional mutual information and the optimization objective, we reformulate the variational distribution $q_{\vtheta}(y \mid b_{t-1}, o_{t})$ by employing the similarity term from Eq.~(\ref{eqn:infonce_obj}), yielding $q_{\vtheta}(y \mid b_{t-1}, o_{t}) \coloneqq p(b_{t-1}, o_{t})e^{-S(f(b_{t-1}, o_t;\vtheta), y)}$. This reformulation equals the negative InfoNCE objective, as presented in Eq.~(\ref{eqn:infonce_obj}). It underscores that the EAD training procedure maximizes the conditional mutual information, guiding the agent to collect maximally informative observations $o_t$ for determining the task-designated annotation $y$. Note that the tightness of the derived lower bound improves as the batch size $K$ increases.
\end{remark}

The last equality in Eq. (\ref{eqn:objective_greedy_exploration}) follows from mutual information being equivalent to a reduction in conditional entropy. This reveals that the optimization guides the policy toward greedy informative exploration, defined as:
\begin{definition}[Greedy Informative Exploration]
\label{def:greedy_informative_exploration}
Greedy informative exploration, represented by $\pi^g$, refers to an action policy which, at any timestep $t$, chooses an action $a_t$ that maximizes the decrease in the conditional entropy of a random variable $y$ given a new observation $o_t$ obtained from executing action $a_t$. Formally,
\begin{equation}
\label{eqn:def_greedy}
\pi^g = \argmax_{\pi \in \Pi} \; [\mathcal{H}(y \mid b_{t-1}) - \mathcal{H}(y \mid b_{t-1},o_t)],
\end{equation}
where $\mathcal{H}(\cdot)$ represents the entropy, $\Pi$ denotes the space encompassing all feasible policies.
\end{definition}

\begin{remark}
The conditional entropy term $\mathcal{H}(y \mid b_{t-1})$ measures the uncertainty in target $y$, conditioned on the belief $b_{t-1}$ maintained up to the previous timestep. In contrast, $\mathcal{H}(y \mid b_{t-1},o_{t})$ quantifies the incorporated uncertainty in $y$ after  the observation $o_t$. While not guaranteed to yield globally optimal behavior over the entire trajectory, the \textbf{greedy informative exploration} strategy serves as an efficient baseline for rapid environmental learning through sequential actions and observations. 
\end{remark}

Through theoretical analysis, we demonstrate that the optimal policy $\pi_{\vphi}^*$ in the InfoNCE objective function (\ref{eqn:infonce_obj}) converges to a greedy informative exploration policy, combining mutual information with policy refinement. The perspective of information theory reveals that a well-trained EAD model naturally adopts a greedy informative policy, leveraging contextual information to address the high uncertainty \cite{smith2018understanding} in scenes containing adversarial patches.

\subsection{Reinforced Embodied Active Defense}
\label{subsec:reinforce-ead}
\begin{figure}[t]
    \centering
    \includegraphics[width=0.99\linewidth]{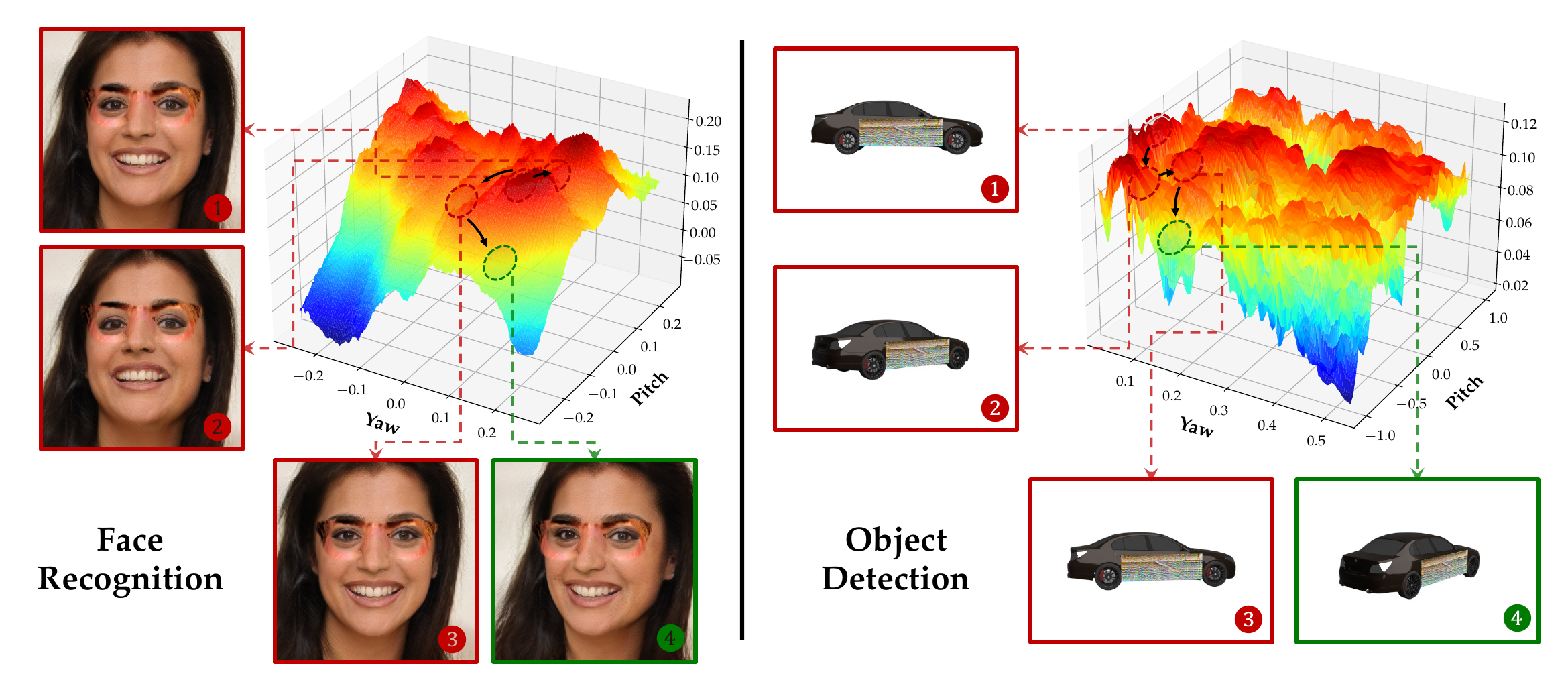}
    \caption{EAD's temporal inconsistency issue visualized on the loss landscape \wrt camera yaw and pitch.  The model often revisits similar states, limiting exploration and increasing vulnerability to adversarial impacts. These trajectories expose the drawbacks of a myopic greedy policy in dynamic settings.}  
    \label{fig:temp_inconsitentcy}
    \vspace{-1em}
\end{figure}

The previous EAD demonstrates significant potential but still faces challenges in three critical domains regarding effectiveness, applicability, and efficiency: (1) \textbf{Temporal inconsistency}: EAD's greedy informative often produces temporally inconsistent actions as illustrated in Fig.~\ref{fig:temp_inconsitentcy}. This inconsistency stems from the myopic nature of greedy policies, which optimize for immediate information gain without considering long-term relevance in embodied learning~\cite{zhang2021robust,ying2022towards}. Consequently, the agent may revert to previously explored viewpoints, reducing exploration effectiveness and potentially leading to erroneous predictions, especially when revisiting viewpoints with significant adversarial impacts~\cite{dong2022viewfool,ruan2023improving}. 
(2) \textbf{Limited applicability}: The reliance on differentiable dynamic models for learning EAD severely constrains its practical application. In real-world scenarios~\cite{sutton1991dyna,blondel2024elements}, accurately modeling environment dynamics in a differentiable manner is often infeasible due to the complexity and unpredictability of physical systems. (3) \textbf{Computational inefficiency and instability}: Learning via differentiable simulation is computationally demanding and prone to numerical instabilities. The process requires extensive computation to traverse the dynamics model and derive their gradients. Furthermore, it often suffers from vanishing gradients during optimization~\cite{bengio1994learning} due to insufficient gradient quality from the simulated dynamics~\cite{antonova2023rethinking}.

\subsubsection{Accumulative Interactions for Temporal Consistency}
\label{subsec:reinforce-ead-accu}
{Greedy exploration in the original EAD myopically minimizes uncertainty at each step in Eq.~(\ref{eqn:objective}) without considering the future effects of actions. Such a greedy approach often leads to temporal inconsistency and suboptimal outcomes from a holistic perspective. To address this limitation and improve the overall performance of EAD, we introduce accumulative informative exploration, which aims to minimize long-term uncertainty about the target variable through a sequence of interactions with the environment.}

\begin{definition}[Accumulative Informative Exploration]
    Accumulative informative exploration, denoted by $\pi^*$, refers to the policy that maximizes the reduction in the entropy of $y$ given a series of observation $o_{0:H}$ resulting from continuous interaction with the environment by executing actions $a_{0:H}$. Formally,
\begin{equation}
\pi^* = \argmax_{\pi \in \Pi} \; [\mathcal{H}(y) - \mathcal{H}(y \mid b_{H-1},o_{H})].
\end{equation}
\end{definition}

\begin{remark}
In contrast to the greedy informative exploration defined in Definition~\ref{def:greedy_informative_exploration}, the accumulative informative policy aims to minimize the uncertainty of the target $y$ through continuous interaction with the environment, rather than focusing solely on single-step. Greedy informative exploration can be considered a special case of accumulative informative exploration, limited to one-step transitions where $H=1$. From a long-term perspective, greedy informative exploration tends to myopically reduce step-wise uncertainty, often leading to sub-optimal outcomes compared to accumulative informative exploration.
\end{remark}

To achieve accumulative informative exploration, we propose a multi-step accumulative interaction objective that optimizes the policy over a horizon of $H$ steps, incorporating terms that encourage reaching belief states that minimize the prediction loss and penalize high-entropy predictions:
\begin{equation}
\label{eqn:multi_step_objective}
\begin{gathered}    
    \min_{\vtheta,\vphi} \; \mathbb{E}_{x,y, \tau } \sum_{p \in \mathcal{P}_{x}} \Big[\mathcal{L}(\hat{y}_H, y) + \lambda \cdot \mathcal{H}(\hat{y}_H \mid b_{H-1}, o_H) \Big], \\
    \textrm{with}\; \{\hat{y}_t, b_{t}\} = f(A(o_t,p;s_t), b_{t-1};\vtheta), \; a_{t} \sim \pi(\cdot\mid b_{t}; \vphi)\\
    \textrm{s.t.} \quad o_{t} \sim \mathcal{Z}(\cdot\mid s_t, x), \quad s_{t} \sim \mathcal{T}(\cdot\mid s_{t-1}, a_{t-1}, x),
\end{gathered}
\end{equation}
where $\mathbb{E}_{x,y, \tau }$ is the abbreviation of $\mathbb{E}_{(x,y)\sim \mathcal{D}, \tau \sim (\mathcal{M}(x),\pi)}$ following Eq.~(\ref{eqn:objective}), $\mathcal{L}(\hat{y}_H, y)$ represents the prediction loss in the step $H$, and $\mathcal{H}(\hat{y}_H \mid b_{t-1}, o_t)$ denotes the entropy of the predicted label at the step $H$. The sampled trajectories follow the same distribution with Eq.~(\ref{eqn:objective}). The entropy term serves as a regularizer, discouraging the agent from making high-entropy predictions that are characteristic of adversarial examples~\cite{smith2018understanding}.

The proposed multi-step interactions align with the definition of accumulative informative exploration, as it seeks to minimize the uncertainty of the target variable through a sequence of actions and observations. By incorporating the prediction loss and the entropy regularization term, the objective encourages the agent to reach belief states that are informative and robust, leading to resilience against adversarial perturbations.
Then we analyze the performance gap between the accumulative informative policy and the greedy one as below.


\begin{theorem}[Informative Policy Efficacy Inequality, Proof in Appendix {A.2}]
\label{thm:efficay_inequality}
Consider two policies interacting continuously with an environment over a time horizon $H$:
\begin{itemize}
\item $\pi^{g}$ denote the \emph{greedy informative policy}, resulting in observation sequence $o^{g}_{0:H}$ and belief sequence $b^{g}_{0:H}$.
\item $\pi^{*}$ denote the \emph{accumulative informative policy}, resulting in observation sequence $o^{*}_{0:H}$ and belief sequence $b^{*}_{0:H}$.
\end{itemize}
Define the \emph{trajectory information gain} from time $0$ to $H$ under a policy $\pi$ as the reduction in entropy of the variable $y$:
\begin{equation*}
\Delta \mathcal{H}_{\pi} \coloneqq \mathcal{H}(y) - \mathcal{H}\big(y \mid b_{H-1}, o_H\big).
\end{equation*}
Assume that the belief update function $f_b: (b_{t-1}, o_t) \mapsto b_t$ is bijective. Then the efficacy of $\pi^{*}$ relative to $\pi^{g}$ satisfies the following inequality:
\begin{equation}
\Delta \mathcal{H}_{\pi^{*}} \geq  \Delta \mathcal{H}_{\pi^{g}},
\end{equation}
where equality holds if and only if the problem exhibits optimal substructure.
\end{theorem}

\begin{remark}
As indicated by this inequality, the efficacy gap primarily arises from two factors: (1) the complexity of the exploration problem, which makes it difficult to guarantee an optimal structure to ensure the effectiveness of the greedy policy, and (2) the information loss during the belief update process, which encodes the previous belief $b_{t-1}$ and the current observation $o_t$ into the current belief $b_t$. The greedy policy does not account for information loss during belief updates through continuous interaction with the environment. Its formulation in Eq.~(\ref{eqn:def_greedy}) only ensures the selection of the most informative observation $o_t$ by taking action $a_{t-1}$ in a single step.
\end{remark}

By extending the greedy information exploration discussed in Sec.~\ref{subsec:uncertainty_analysis}, we can reformulate Eq.~(\ref{eqn:infonce_obj}) into a multi-step interactive form. The learned policy model $\pi_{\vphi}^*$ represents an accumulative informative policy, under the assumptions of unlimited model capacity and data samples. Additionally, Theorem \ref{thm:efficay_inequality} formally establishes the superiority of multi-step interactions, indicating that a well-trained multi-step model for contrastive tasks adopts an \textbf{accumulative informative policy} to continuously explore the environment. This model utilizes a series of contextual information from the environment to consistently reduce perceptual uncertainty caused by adversarial patches, providing a theoretical foundation for our multi-step accumulative objective in Eq.~(\ref{eqn:multi_step_objective}).



\subsubsection{Policy Learning for Real-World Applicability}
\label{subsec:reinforce-ead-free}
Despite the effectiveness in Sec.~\ref{subsec:reinforce-ead-accu}, \textsc{Rein}-EAD faces significant challenges in real-world applications. The unpredictability of physical systems makes it infeasible to model environment dynamics in a differentiable manner. Moreover, the inevitable computational overhead from differentiating the dynamics model and the numerical instability arising from approximations render it impractical for solving Eq.~(\ref{eqn:multi_step_objective}), especially over long horizons. These issues intensify with growing computational demands and cumulative gradient estimation errors, hindering the applicability and efficiency.

\begin{algorithm}[t]\small
\caption{Training \textsc{Rein}-EAD by Policy Learning}\label{alg:ead-rl}
\begin{algorithmic}[1]
   \Require Training data $\mathcal{D}$, number of iterations $M$, number of epochs $E$, loss function 
    $\mathcal{L}$, perception model $f(\cdot;\vtheta)$, policy model $\pi(\cdot;\vphi)$.
    \Ensure The parameters $\vtheta, \vphi$ of the learned EAD model. 
\For{iteration $\gets 0$ \textbf{to} $M - 1$}
    \LComment{Roll-out perception $f(\cdot;\vtheta)$ and policy $\pi(\cdot;\vphi)$ in the environment.}
    \State Collect set of augmented trajectory and label pairs $\mathcal{D}_{\tau} = \{(\tau, y)\}$ by running policy $\pi(\cdot;\vphi)$ and perception $f(\cdot;\vtheta)$ on $\mathcal{M}(x)$ with $(x, y) \sim \mathcal{D}$, where
    \begin{equation*}
        \tau = (o_0', b_0, a_0, r_0, o_1', b_1, \ldots)
    \end{equation*}
    \Statex
    \State Estimate advantages $\hat{A}_t$ using any advantage estimation algorithm

     \For{epoch $\gets 0$ \textbf{to} $E - 1$}
        \State $\vphi_{\text{old}} \gets \vphi$
        \ForAll{mini-batch $\mathcal{B}_{\tau} \in \mathcal{D}_{\tau}$} 
             \State Update $\vphi$ by maximizing estimated the PPO-Clip objective $\mathcal{J}_\text{policy}(\vphi)$ defined as:
            \begin{equation*}
                \frac{1}{|\mathcal{B}_{\tau}|}\sum_{\tau \in\mathcal{B}_{\tau}}\sum_{t}\min(R(\vphi)\hat{A}_t, \mathrm{clip}(R(\vphi), 1-\epsilon, 1+\epsilon) \hat{A}_t)
            \end{equation*} with one-step gradient ascent, where
            \begin{equation*}
                R(\vphi) = \frac{\pi(a_t \mid b_t,\vphi)}{\pi(a_t \mid b_t,\vphi_{\text{old}})}
            \end{equation*}\vspace{0pt}
            \State Update $\vtheta$ by minimize the estimated objective in Eq.~(\ref{eqn:multi_step_objective}), namely $\mathcal{J_{\text{percep}}(\vtheta)}$:
            \begin{equation*}
                \frac{1}{|\mathcal{B}_{\tau}|}\sum_{(\tau, y)\in\mathcal{B}_{\tau}}\sum_{t} \mathcal{L}(f(o_t, b_t; \vtheta), y) + \lambda \cdot \mathcal{H}(f(o_t, b_t; \vtheta))
            \end{equation*}
            with one-step gradient descent.
        \EndFor
    \EndFor
\EndFor
\end{algorithmic}

\end{algorithm}

To address these challenges and enhance the real-world applicability of our approach, we propose a policy learning method that incorporates an uncertainty-oriented reward-shaping technique within the reinforcement learning framework. By eliminating the need for differentiable dynamic models, 
our method enables the agent to directly learn a policy that maximizes the expected cumulative reward through trial-and-error interactions with the environment.  This flexibility allows the agent to efficiently adapt to changing dynamics or stochastic environments, which are prevalent in real-world scenarios~\cite{dosovitskiy2017carla}. 

Specifically, the uncertainty-oriented reward-shaping technique is designed to provide dense rewards at each step, guiding the agent to reduce perceptual uncertainty and minimize prediction errors. This approach addresses classical sparse reward strategy~\cite{sutton2018reinforcement} in Eq.~(\ref{eqn:multi_step_objective}), where the agent can only access the reward at the end of the episode. By incorporating a weighted combination of intermediate rewards, our method allows for more granular and informative feedback to the agent.
Formally, we define the dense reward $r_t$ as follows (proof of the equivalence in Appendix {A.3}):
\begin{equation}
    \label{eqn:reward}
    r_t = \mathcal{L}(\hat{y}_{t-1}, y) - \gamma \cdot \mathcal{L}(\hat{y}_{t}, y), \quad (t > 0)
\end{equation}
where $\gamma$ is the discount factor. This dense reward structure motivates the policy $\pi$ to seek new observations $o_t$ as feedback from the environment. The continuous feedback loop facilitated by the dense rewards enables the agent to efficiently adapt to new situations and make informed decisions in the face of uncertainty. Moreover, by fairly distributing the reward across each step, we alleviate the challenges of exploration and credit assignment~\cite{ng1999policy}, facilitating faster convergence and more efficient learning. 

As for the reinforcement learning backbone, we employ Proximal Policy Optimization (PPO)~\cite{schulman2017proximal} because of its learning efficiency and convergence stability. PPO enables stable policy updates by constraining the size of the policy change at each iteration based on the dense rewards provided by our uncertainty-oriented reward-shaping technique. This incremental learning process ensures that the agent maintains a stable trajectory towards reducing uncertainty and minimizing prediction errors. The detailed training procedure is in Algorithm~\ref{alg:ead-rl}.

\subsection{Adversary-agnostic Defense against Patch Attacks}
\label{subsec:impl_tech}

The computation of $\mathcal{P}_{x}$ typically necessitates the resolution of the inner maximization in Eq.~(\ref{eqn:objective}) by online adversarial training~\cite{madry2017towards}. However, this is not only computationally expensive ~\cite{wong2020fast} but also problematic as inadequate assumptions for characterizing adversaries can hinder the model's ability to generalize across diverse, unseen attacks~\cite{laidlaw2020perceptual}. 



\noindent \textbf{Offline adversarial patch approximation (OAPA).}   While the USAP approach~\cite{wuembodied} effectively learns an optimal informative strategy given sufficient training epochs, it is computationally expensive due to the need for extensive sampling to ensure the sampled manifold contains an adequate number of adversarial examples. To improve sampling efficiency while preserving the adversary-agnostic property, we introduce OAPA that employs the projected gradient technique to approximate the adversarial patch manifold $\gP_{x}$ before training the \textsc{Rein}-EAD model. OAPA systematically characterizes the manifold of adversarial patterns by pre-generating a surrogate set of patches $\Tilde{\gP} = {p_i}$, where each patch $p_i$ is derived through projected gradient ascent against the visual backbone. This offline approximation of the adversarial patch manifold allows the \textsc{Rein}-EAD model to learn a compact yet expressive representation of adversarial patterns, enabling it to effectively defend against previously unseen attacks.  Our empirical findings suggest that performing this offline approximation maximization is highly effective in developing models robust to a broad spectrum of attacks (refer to Sec.~\ref{sec:exp}). Additionally, because this maximization process occurs offline and only once before training, it substantially boosts training efficiency and renders it competitive with conventional training methods.

\section{Experiments}
\label{sec:exp}






In this section, we verify the effectiveness of \textsc{Rein}-EAD on different tasks, including face recognition in Sec.~\ref{subsec:fr}, 3D object classification in Sec.~\ref{subsec:classification} and object detection in Sec.~\ref{subsec:od}. 

\subsection{Evaluation on Face Recognition}
\label{subsec:fr}


\begin{table*}[t] 
\caption{The \textbf{Standard accuracy} (\%) and \textbf{attack success rates} (\%) on face recognition. $^\dagger$ denotes the methods using adversarial training. \colorbox{lightcyan}{Methods with light blue background} do not require a differentiable environment, while the rest do.}
\centering
\setlength\tabcolsep{7.5pt}
\renewcommand\arraystretch{1.25}
\label{tab:glass_xl_cannonical}
\begin{tabular}{c|c|cccc|cc|cc|cc}
\hline
\multirow{2}{*}{Method} & \multirow{2}{*}{Acc. (\%)} & \multicolumn{4}{c|}{White-box} & \multicolumn{2}{c|}{Transfer-based} & \multicolumn{2}{c|}{Query-based} & \multicolumn{2}{c}{Adaptive}  \\ 
\cline{3-12} 
& & MIM & EoT & GenAP & 3DAdv & Cos. & Softmax & NAttack & RGF & BPDA & Worst-case \\
\hline \hline
\multicolumn{12}{c}{\textit{Impersonation Attack}} \\
\hline 
Undefended & 88.86 & 100.0 & 100.0 & 99.00 & 89.00 & 28.00 & 23.00 & 100.0 & 100.0 & 100.0 & 100.0 \\ \hline
\rowcolor{lightcyan} JPEG & 89.98 & 99.00 & 100.0 & 99.00 & 93.00 & 33.00 & 33.00 & 96.00 & 94.00 & 99.00 & 100.0 \\ 
\rowcolor{lightcyan} LGS & 83.50 & 5.10 & 7.21 & 33.67 & 30.61 & 11.63 & 6.98 & 11.63 & 4.65 & 38.37 & 48.83 \\ 
SAC & 86.83 & 6.06 & 9.09 & 67.68 & 64.64 & 8.70 & 9.78 & 13.04 & 14.13 & 48.00 & 67.68 \\ 
PZ & 87.58 & 4.17 & 5.21 & 59.38 & 45.83 & 6.45 & 9.68 & 4.30 & 3.26 & 89.76 & 92.86 \\ \hline
SAC$^\dagger$ & 80.55 & 3.16 & 3.16 & 18.94 & 22.11 & 11.11 & 12.36 & 12.22 & 14.44 & 51.73 & 51.73\\ 
PZ$^\dagger$ & 85.85 & 3.13 & 3.16 & 19.14 & 27.34 & 8.24 & 5.00 & 10.58 & 9.41 & 61.01 & 98.99\\ 
DOA$^\dagger$ & 79.55 & 95.50 & 89.89 & 96.63 & 89.89 & 15.73 & 17.97 & 34.83 & 16.86 & 89.89 & 96.63 \\ 
{EAD} & \textbf{90.45} & 4.12 & 3.09 & 5.15 & \textbf{7.21} & 4.17 & 5.20 & 4.12 & 4.12 & 8.33 & 9.38\\
\rowcolor{lightcyan} \textbf{\textsc{Rein}-EAD} & 89.03 & \textbf{2.10} & \textbf{1.06} & \textbf{3.15} & 7.37 & \textbf{2.10} & \textbf{2.08} & \textbf{1.05} & \textbf{2.12} & \textbf{4.21} & \textbf{7.37}\\ \hline
\multicolumn{12}{c}{\textit{Dodging Attack}} \\
\hline 
Undefended & 88.86 & 100.0 & 100.0 & 99.00 & 98.00 & 44.00 & 35.00 & 96.00 & 96.00 & 100.0 & 100.0\\ \hline
\rowcolor{lightcyan} JPEG & 89.98 & 98.00 & 99.00 & 95.00 & 88.00 & 49.00 & 45.00 & 81.00 & 83.00 & 100.0 & 100.0 \\
\rowcolor{lightcyan} LGS & 83.50 & 49.47 & 52.63 & 74.00 & 77.89 & 22.11 & 21.05 & 18.95 & 20.00 & 78.92 & 78.92 \\
SAC & 86.83 & 73.46 & 73.20 & 92.85 & 78.57 & 40.80 & 36.84 & 55.26 & 50.00 & 65.22 & 92.85 \\
PZ & 87.58 & 6.89 & 8.04 & 58.44 & 57.14 & 41.67 & 28.34 & 28.33 & 31.67 & 88.89 & 90.00\\ \hline 
SAC$^\dagger$ & 80.55 & 78.78 & 78.57 & 79.59 & 85.85 & 47.46 & 43.54 & 55.93 & 62.71 & 85.02 & 85.85\\
PZ$^\dagger$ & 85.85 & 6.12 & 6.25 & 14.29 & 20.41 & 50.88 & 47.69 & 56.14 & 50.87 & 69.45 & 98.00 \\
DOA$^\dagger$ & 79.55 & 75.28 & 67.42 & 87.64 & 95.51 & 30.33 & 31.46 & 53.93 & 28.09 & 95.51 & 95.51 \\ 
EAD & \textbf{90.45} & \textbf{0.00} & \textbf{0.00} & \textbf{2.10} & 13.68 & 5.26 & 7.36 & 1.05 & \textbf{0.00} & 22.11 & 22.11\\
\rowcolor{lightcyan} \textbf{\textsc{Rein}-EAD} & 89.03 & 1.04 & 2.04 & 5.15 & \textbf{13.54} & \textbf{4.17} & \textbf{7.29} & \textbf{1.03} & \textbf{0.00} & \textbf{8.16} & \textbf{14.43} \\
\hline
\end{tabular}
\vspace{-1em}
\end{table*}

\subsubsection{Experimental Settings}

\textbf{Experimental environment.} To enable the unconstrained navigation and observation collection of \textsc{Rein}-EAD, we construct a manipulable simulation environment for both training and purposes. To align with specific vision tasks, we define the state as a combination of the camera’s yaw and pitch, while the action corresponds to the rotation of the camera\footnote{To prevent the agent from ``cheating'' by rotating to angles that conceal adversarial patches, specific constraints are imposed on the yaw and pitch.}. This definition establishes the transition function, with the core of the simulation environment revolving around the observation function, which generates a 2D image based on the camera’s state. As elaborated in Sec.~\ref{subsec:ead}, the training of original EAD requires differentiable environmental dynamics. To achieve a fair comparison, we first employ the advanced 3D generative model, EG3D~\cite{chan2022efficient}, which enables realistic differentiable rendering (refer to Appendix {C.1} for simulation fidelity). To further substantiate the superiority of \textsc{Rein}-EAD, we conduct training and testing of \textsc{Rein}-EAD in the same environment to ensure fair comparison.



 
\noindent \textbf{Evaluation metrics and protocols.} To rigorously validate the effectiveness of \textsc{Rein}-EAD, we conduct extensive experiments on CelebA-3D, which we meticulously reconstruct from 2D face images in CelebA into 3D representations by utilizing GAN inversion~\cite{zhu2016generative} with EG3D~\cite{chan2022efficient}. {To assess the \textbf{standard accuracy}, we sample 2,000 test pairs from the CelebA and follow the well-established evaluation protocol from LFW~\cite{LFWTech}. To comprehensively evaluate the robustness, we report the \textbf{attack success rate (ASR)} on 100 identity pairs, considering both impersonation and dodging attacks~\cite{yang2023adversarial} under various attack methods in both white-box and black-box settings.} The white-box attack methods emcompass MIM~\cite{dong2018boosting}, EoT~\cite{athalye2018synthesizing}, GenAP~\cite{xiao2021improving} and Face3Dadv (3DAdv)~\cite{yang2024face3dadv}. In the black-box attacks, we employ the transfer-based attack, targeting the surrogate models such as IResNet-18 CosFace~\cite{duta2021improved,wang2018cosface} and IResNet-50 Softmax with 3DAdv, and the query-based methods including NAttack~\cite{li2019nattack} and RGF~\cite{ghadimi2013stochastic}. Note that 3DAdv utilizes expectation over 3D transformations during the optimization, endowing it with inherent robustness to 3D viewpoint variation within a range of $\pm $15$^{\circ}$. More details are described in Appendices {\color{red}C.1} \& {\color{red}C.2}. 

\noindent \textbf{Implementation details.}
To extract discriminative visual features, we employ the pretrained IResNet-50 ArcFace~\cite{duta2021improved} as the visual backbone, leveraging its pretrained weights while keeping them frozen during subsequent training stages. To model the recurrent perception and policy components, we adopt a variant of the Decision Transformer~\cite{chen2021decision}, which effectively processes feature sequences extracted by the visual backbone to predict a normalized embedding for FR. We set the maximum horizon length to $H$~=~4 in EAD as default. In contrast to EAD, \textsc{Rein}-EAD eliminates the dependency on Back-Propagation Through Time (BPTT), allowing us to extend the horizon length to $H$~=~16 without VRAM constraints. Furthermore, we incorporate an additional MLP value head into \textsc{Rein}-EAD to facilitate advantage estimation for PPO. Further details on \textsc{Rein}-EAD can be found in Appendix {C.5}.

\begin{figure*}[t]
    \centering
 \includegraphics[width=0.99\linewidth]{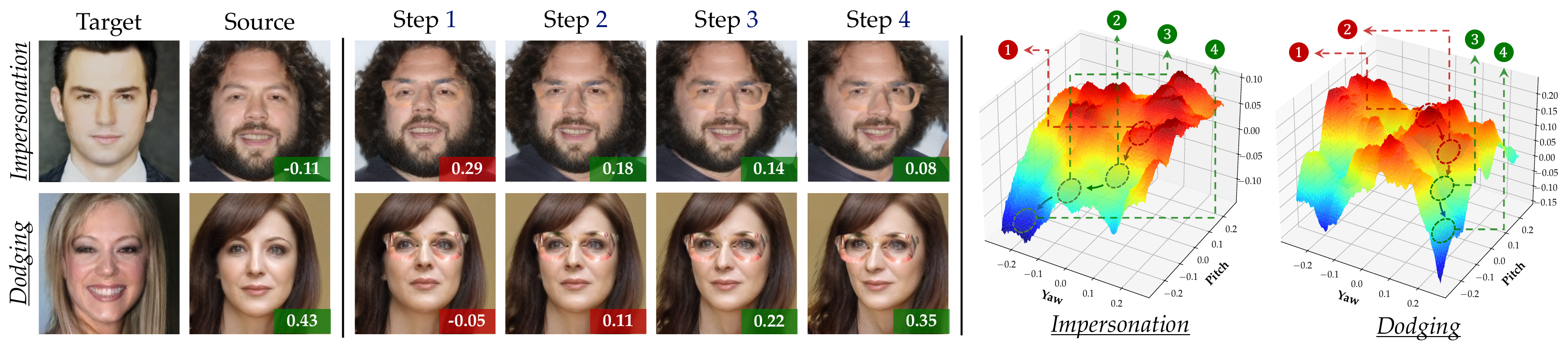}
 \vspace{-1em}
    \caption{\textbf{Qualitative results} of \textsc{Rein}-EAD. The first two columns present the original image pairs, and the subsequent columns depict the interactive inference steps taken by the model. The defensive trajectory of \textsc{Rein}-EAD is plotted on the loss landscape \wrt yaw and pitch of the camera, considering the IResNet-50 ArcFace as the target model~\cite{duta2021improved}. The adversarial glasses are generated with 3DAdv, which are robust to 3D viewpoint variation. The computed optimal threshold for distinguishing between positive and negative pairs is set to 0.19 from [-1, 1].}
    \label{fig:vis_efr_face}
    \vspace{-1em}
\end{figure*}

 \begin{figure*}[t]
    \centering
    \includegraphics[width=0.9\linewidth]{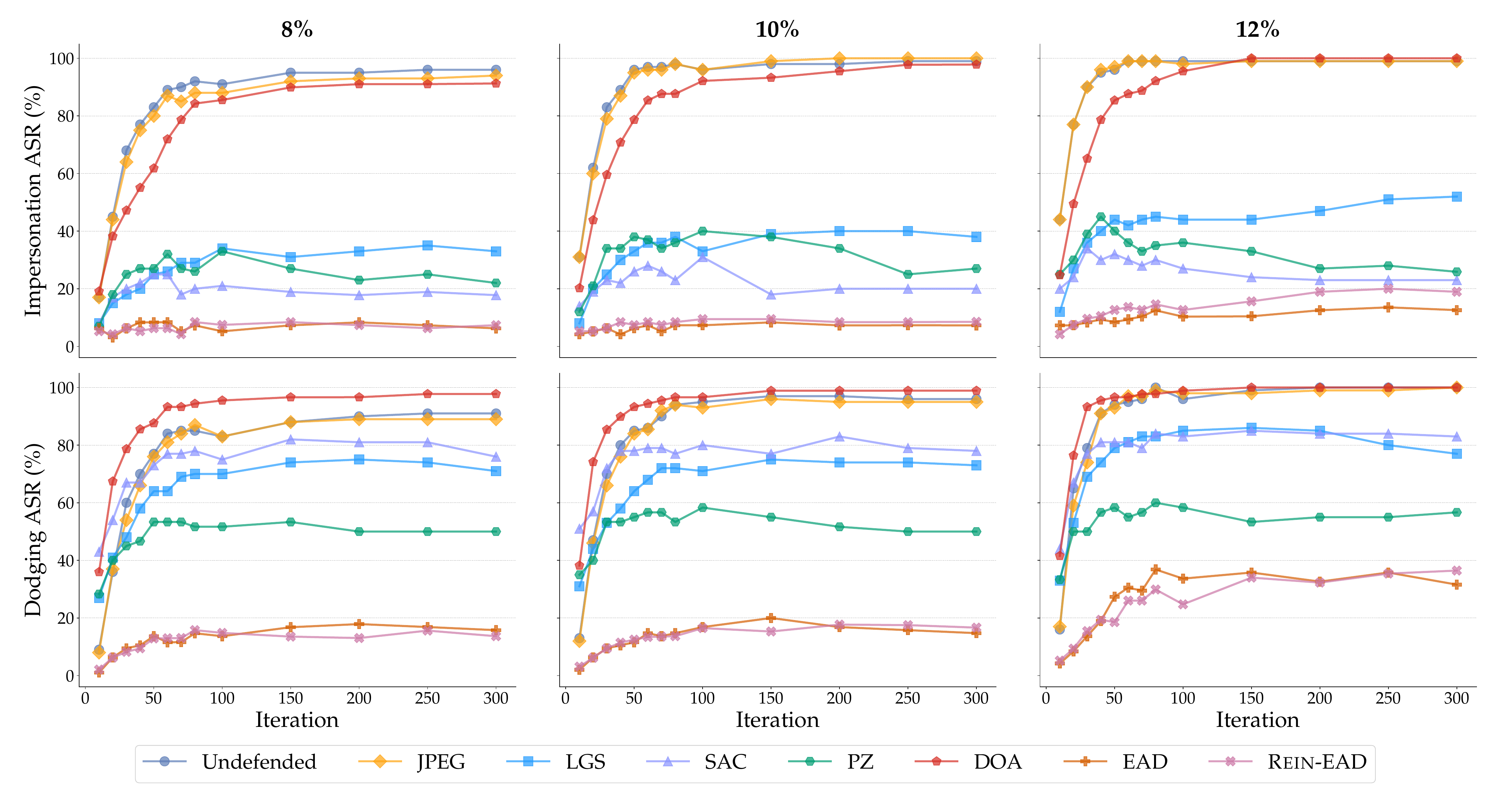}  
 \vspace{-1em}
    \caption{Comparative evaluation of defense methods across varying attack iterations with different adversarial patch sizes. Note that SAC, PZ and DOA are involved with adversarial training.}
    \label{fig:asr_iterations}
    \vspace{-1em}
\end{figure*}

\noindent \textbf{Defense baselines.}
To fully evaluate the effectiveness of \textsc{Rein}-EAD, we benchmark it against a diverse range of state-of-the-art defense methods. These baselines include adversarial training-based Defense against Occlusion Attacks (DOA)~\cite{wu2019defending}, and {purification-based methods like JPEG compression (JPEG)~\cite{dziugaite2016study}, local gradient smoothing (LGS)~\cite{naseer2019local}, segment and complete (SAC)~\cite{liu2022segment}, PatchZero (PZ)~\cite{xu2023patchzero}, Patch-Agnostic Defense (PAD)~\cite{jing2024pad} and DIFFender~\cite{kang2024diffender}}. For DOA, we employ rectangle-shaped PGD patch attacks~\cite{madry2017towards} with 10 iterations and a step size of ${2}/{255}$. Note that SAC and PZ require a patch segmenter to locate the area of adversarial patches. Therefore, we train the segmenter using patches of Gaussian noise to ensure the same adversarial-agnostic setting. Besides, we consider enhanced versions of SAC and PZ, denoted as SAC$^\dagger$ and PZ$^\dagger$, which involve training with adversarial patches generated using the EoT technique. More details are presented in Appendix {C.4}. 

\subsubsection{Experimental Results}

\noindent\textbf{Effectiveness of \textsc{Rein}-EAD.} Table~\ref{tab:glass_xl_cannonical} presents a comprehensive evaluation of both the {standard accuracy} and {robust performance} against diverse attacks under a white-box setting, with the adversarial patch size set to 8\% of the image size. Remarkably, our approach significantly outperforms previous state-of-the-art passive techniques that are agnostic to adversarial examples in both clean accuracy and defense efficacy. For instance, \textsc{Rein}-EAD reduces the attack success rate of 3DAdv by an impressive effect in both scenarios. Furthermore, \textsc{Rein}-EAD also improves the average attack rate reduction when faced with black-box and adaptive attacks compared with EAD by extending the informative greedy policy to an accumulative one. Notely, \textsc{Rein}-EAD even surpasses the performance of the undefended passive model regarding standard accuracy, effectively reconciling the trade-off between robustness and accuracy~\cite{su2018robustness} through embodied perception. Furthermore, our method even outstrips the baselines by incorporating adversarial examples during training. Although SAC$^\dagger$ and PZ$^\dagger$ are trained using patches generated via EoT~\cite{athalye2018synthesizing}, we still obtain superior performance,  highlighting the effectiveness of leveraging the environmental feedback in active defense. 

Fig.~\ref{fig:vis_efr_face} provides a visual illustration of the defense process executed by \textsc{Rein}-EAD. While \textsc{Rein}-EAD may be initially fooled by the adversarial patch, its subsequent active interactions with the environment progressively increase the similarity between the positive pair and decrease the similarity between the negative pair. Consequently, \textsc{Rein}-EAD effectively mitigates the impact of adversarial hallucination through the proactive acquisition of additional observations, as depicted in the corresponding loss landscapes.

\noindent \textbf{Effectiveness against adaptive attack.} 
While the deterministic and differentiable dynamic models could potentially enable backpropagation through the entire inference trajectory of EAD, the computational cost becomes prohibitive due to the rapid consumption of GPU memory as trajectory length $H$ increases. To overcome this, we first adopt an approach similar to the original strategy that approximates the true gradient by computing the expected gradient over a surrogate uniform superset policy distribution. This approximation (BPDA) necessitates an optimized patch to handle diverse action policies. Our adaptive attack implementation builds upon 3DAdv~\cite{yang2024face3dadv} leveraging 3D viewpoint variations. Moreover, we also evaluate more sophisticated adaptive attacks, including 1) attacking the whole pipeline using true gradients obtained by gradient checkpointing, and 2) targeting the perception and policy sub-modules independently. 
The \textbf{worst-case} performance of \textsc{Rein}-EAD across two adaptive attacks is presented in Table~\ref{tab:glass_xl_cannonical} for a reliable evaluation. Moreover, to benchmark other defense methods, we report their worst-case performance under a series of adaptive attacks. More details are presented in Appendix {C.3}. We can see that \textsc{Rein}-EAD maintains its robustness against the most potent adaptive attacks. This observation shows that the defensive capabilities of \textsc{Rein}-EAD stem from the synergistic integration of its policy and perception models, rather than relying on a short-cut strategy to neutralize adversarial patches from specific viewpoints. 

\begin{table*}[t]
\caption{The \textbf{standard accuracy} and \textbf{white-box impersonation attack success rates} on ablated models. For the model with stochastic policy, we report the mean and standard deviation across five independent runs to ensure reliability.}
\setlength\tabcolsep{9.3pt}
\renewcommand\arraystretch{1.25}
\centering
\begin{tabular}{c|c|c|ccccc}
\hline
\multirow{2}{*}{Category} & \multicolumn{1}{c|}{\multirow{2}{*}{Component}}        & \multirow{2}{*}{Acc. (\%)} & \multicolumn{5}{c}{Attack Success Rate (\%)}                                                                                                                                                                               \\ \cline{4-8} 
                        & &                         & MIM & {EoT}       &{GenAP}        & {3DAdv}     & Adaptive        \\ \hline \hline
          \multirow{3}{*}{Passive Perception} & Undefended & 88.86 & 100.0 & 100.0 & 99.00 & 98.00 & 98.00 \\ \cline{2-8} 
& \multirow{2}{*}{+ Random Movement}  & 90.38 & 4.17 & 5.05 & 8.33 & 76.77 & 76.77 \\ 
& & ($\pm$~0.12) & ($\pm$~2.28) & ($\pm$~1.35) & ($\pm$~2.21) & ($\pm$~3.34) & ($\pm$~3.34) \\ \hline
\multirow{4}{*}{EAD} & \multirow{2}{*}{+ Perception Model} & 90.22 & 18.13 & 18.62 & 22.19 & 30.77 & 31.13 \\ 
& & ($\pm$~0.31) & ($\pm$~4.64) & ($\pm$~2.24) & ($\pm$~3.97) & ($\pm$~1.81) & ($\pm$~3.01) \\ \cline{2-8}
& + Policy Model & 89.85 & 3.09 & 4.12 & 7.23 & 11.34 & 15.63 \\ \hline
\multirow{6}{*}{\textsc{Rein}-EAD} & \multirow{2}{*}{+ Multi-steps Interaction} & 89.02 & 2.12 & 3.19 & 5.33 & 12.77 & 6.38 \\ 
& & ($\pm$~0.18) & ($\pm$~0.08) & ($\pm$~0.44) & ($\pm$~0.87) & ($\pm$~1.13) & ($\pm$~0.80) \\ \cline{2-8}
& \multirow{2}{*}{{+ {OAPA (FGSM~\cite{szegedy2013intriguing})}}} & {88.95} & {3.22} & {3.03} & {5.15} & {13.18} & {5.14} \\ 
& & {($\pm$~0.19)} & {($\pm$~0.23)} & {($\pm$~0.41)} & {($\pm$~1.06)} & {($\pm$~1.11)} & {($\pm$~0.62)} \\ \cline{2-8}
& \multirow{2}{*}{+ {OAPA (PGD~\cite{madry2017towards})}} & 89.03 & \textbf{2.10} & \textbf{1.06} & \textbf{3.15} & \textbf{7.37} & \textbf{4.21} \\ 
& & ($\pm$~0.21) & ($\pm$~0.42) & ($\pm$~0.50) & ($\pm$~1.08) & ($\pm$~1.32) & ($\pm$~0.56) \\ \hline
\end{tabular}
\label{tab:ablation}
\vspace{-1em}
\end{table*}

\noindent \textbf{Generalization of \textsc{Rein}-EAD.}
As demonstrated in Table~\ref{tab:glass_xl_cannonical}, despite no prior knowledge of specific adversaries, ours exhibits remarkable generalization across various \textbf{unseen adversarial attack methods}. It is partially attributed to the inherent ability of \textsc{Rein}-EAD to dynamically interact with their environment, enabling them to adapt and respond to novel types of attacks. Additionally, we assess the models' resilience across a wide range of \textbf{patch sizes} and \textbf{attack iterations}. Despite being trained solely on patches constituting 10\% of the image, \textsc{Rein}-EAD consistently maintains a notably low attack success rate, even when subjected to larger patch size and increased attack iteration, as shown in Fig.~\ref{fig:asr_iterations}. This exceptional resilience can be attributed to \textsc{Rein}-EAD's primary reliance on environmental information, rather than solely depending on patterns of presupposed adversaries. By dynamically updating their perception, \textsc{Rein}-EAD can generalize well to different aspects. The details are available in Appendix {C.7}. 


\begin{figure}[t]
     \centering
     \includegraphics[width=0.99\linewidth]{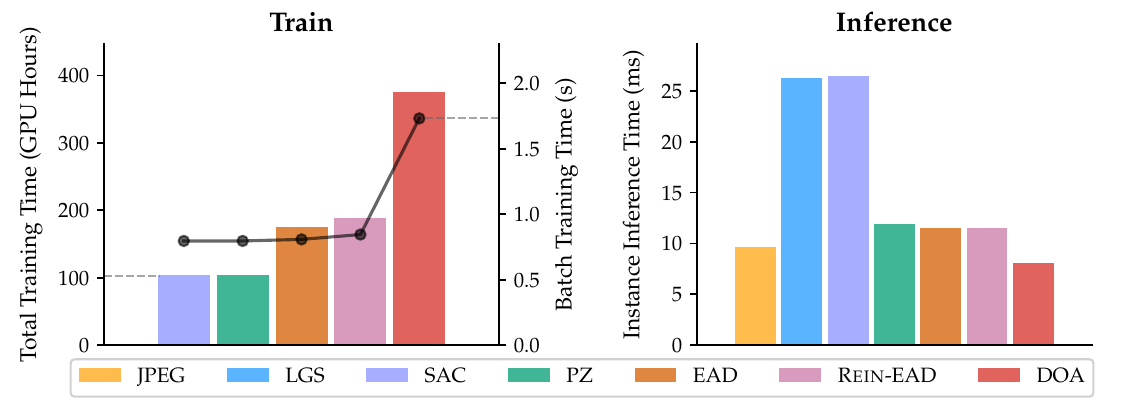}
        \caption{Comparative evaluation of {computational overhead} of defense methods.}
        \label{fig:computational_overhead}
        \vspace{-1em}
\end{figure}

\noindent \textbf{Computational overhead.} We further compare the computational overhead of \textsc{Rein}-EAD with passive defense baselines regarding both training and inference times. { The evaluation is conducted on an NVIDIA GeForce RTX 3090 Ti with a batch size of 64. SAC~\cite{liu2022segment} and PZ~\cite{xu2023patchzero} require two-stage segmenter training: initial training with pre-generated adversarial images followed by self-adversarial training. DOA~\cite{wu2019defending} necessitates feature extractor retraining. Our \textsc{Rein}-EAD approach involves offline and online phases without adversarial training.}
As indicated in Fig.~\ref{fig:computational_overhead}, although differential rendering employed by EAD imposes significant computational demands during the online training phase, the total training time of EAD effectively achieves an effective balance between the purely adversarial training method DOA and partially adversarial methods such as SAC and PZ. This efficiency primarily stems from our unique adversary-agnostic approach (OAPA), which eliminate the need to generate adversarial examples \emph{online}, thereby enhancing training efficiency. Despite the larger horizon of \textsc{Rein}-EAD, which inherently increases sampling time, it still demonstrates faster training than EAD. This advantage stems from \textsc{Rein}-EAD employed by the model-free approach avoids the computationally intensive process of backpropagation to obtain policy gradients, thus improving overall learning efficiency.  {Regarding model inference, the perception model accounts for 98.4\% of this processing time, while the lightweight policy MLP requires only 1.6\%. In total, \textsc{Rein}-EAD exhibits superior speed compared to other baselines, such as LGS and SAC.} This advantage is attributed to the reliance of LGS and SAC on CPU-intensive, rule-based image preprocessing techniques, which inevitably reduces their inference efficiency.  { More details about the computational overhead are provided in Appendix {C.6}.}

\begin{figure}
    \centering
    \vspace{-0.5em}
    \includegraphics[width=0.99\linewidth]{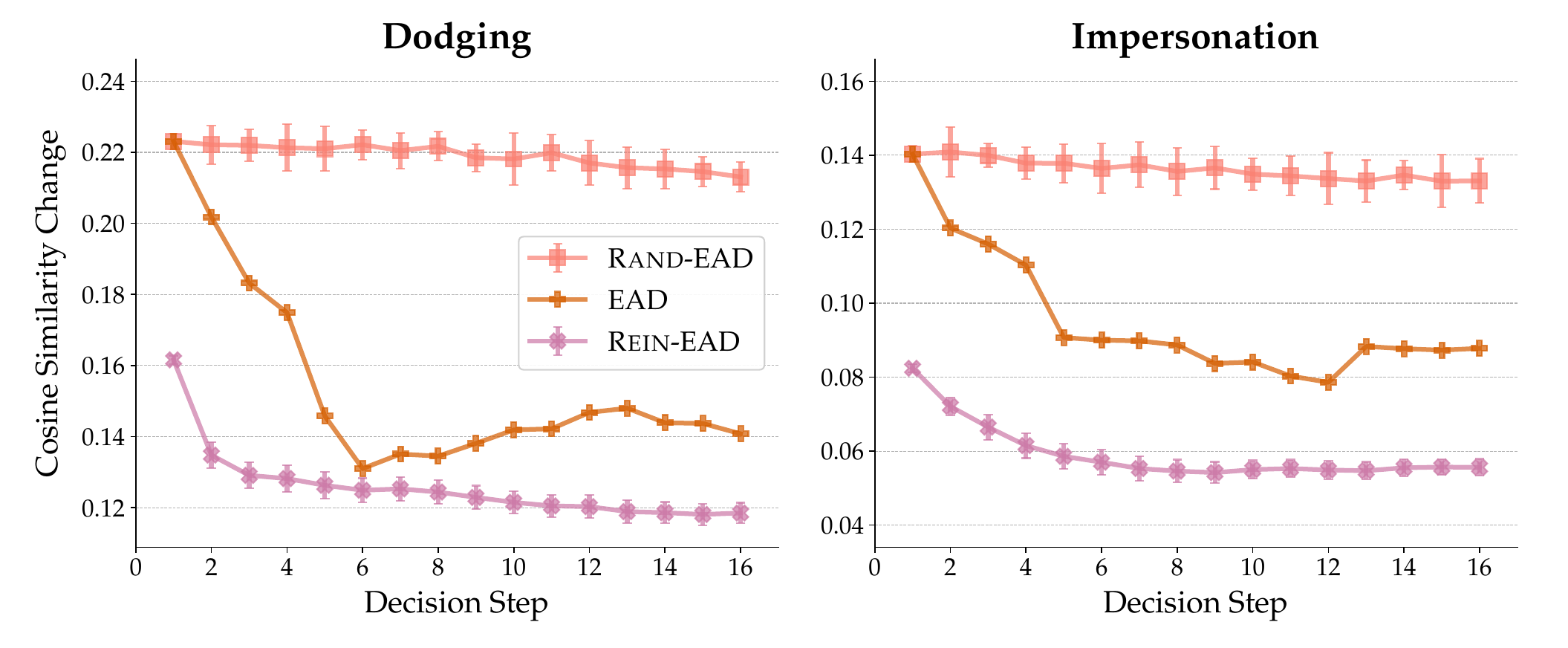}
    \caption{Performance variation along different decision steps.}
    \label{fig:sim_step}
    \vspace{-1.5em}
\end{figure}

\subsubsection{Ablation Study}




\begin{figure*}[ht]
    \centering
    \includegraphics[width=0.99\linewidth]{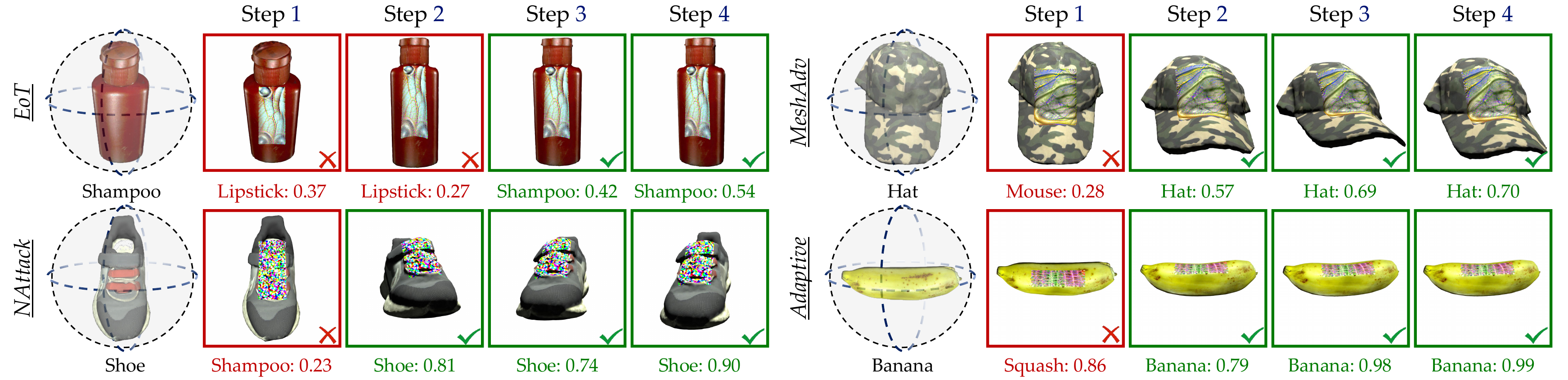}  
    \caption{Qualitative results of \textsc{Rein}-EAD on dynamic OmniObject3D, with the adversarial patch occupying $20\%$ of the object's bounding box in the front view. The state space for object classification is defined as $[-\frac{\pi}{2}, \frac{\pi}{2}] \times [0, \frac{\pi}{2}]$, encompassing a comprehensive range of viewpoints.}
    \label{fig:visual_classify}
    \vspace{-1em}
\end{figure*}

\begin{table*}[t] 
\caption{The \textbf{Standard accuracy} (\%) and \textbf{attack success rates} (\%) on 3D object classification. $^\dagger$ denotes the methods using adversarial training. \colorbox{lightcyan}{Methods with light blue background} do not require a differentiable environment, while the rest do.} 
\label{tab:object_classification}
\centering
\setlength\tabcolsep{11.0pt}
\renewcommand\arraystretch{1.25}
\begin{tabular}{c|c|ccc|c|cc|c}
\hline
\multirow{2}{*}{Method} & \multirow{2}{*}{Acc. (\%)} & \multicolumn{3}{c|}{White-box} & Transfer-based & \multicolumn{2}{c|}{Query-based} & {Adaptive}  \\ \cline{3-9} 
 & & {MIM} & {EoT} & {MeshAdv} & {Swin-T} & {RGF} & {NAttack} & {BPDA} \\
\hline\hline
Undefended & 88.17 & 100.00 & 93.72 & 96.19 & 58.90 & 68.03 & 96.48 & 100.00 \\
\hline
\rowcolor{lightcyan} JPEG & 83.22 & 99.50 & 92.44 & 94.76 & 43.35 & 15.73 & 49.09 & 100.00 \\
\rowcolor{lightcyan} LGS  & 85.91 & 18.36 & 61.91 & 64.55 & 33.59 & 38.96 & 13.77 & 93.46 \\
{PAD}	& {87.16} & {25.31} & {27.14} & {29.45} & {18.48} & {17.81} & {27.22} & {90.47} \\ 
{ DIFFender}	& {80.20} & {20.61} & {54.71} & {61.40} & {28.24} & {14.15} & {18.41} & {45.08} \\
\hline
SAC$^\dagger$  & 88.00 & 10.30 & 9.53  & 10.49 & 9.72  & 11.15 & 10.20 & 57.01 \\
PZ$^\dagger$   & 88.00 & 5.34  & 5.34  & 7.53  & 4.29  & 9.06  & 4.39  & 80.65 \\
DOA$^\dagger$  & 87.33 & 6.63  & 6.53  & 7.30  & 5.57  & 4.61  & 5.96  & 59.75 \\
\rowcolor{lightcyan} \textbf{\textsc{Rein-EAD}}  & \textbf{88.93} & \textbf{3.21}  & \textbf{4.15}  & \textbf{4.34}  & \textbf{3.87} & \textbf{2.26} & \textbf{3.87}  & \textbf{28.96} \\
\hline
\end{tabular}
\vspace{-1em}
\end{table*}

\begin{figure*}[ht]

    \centering
    \includegraphics[width=0.95\linewidth]{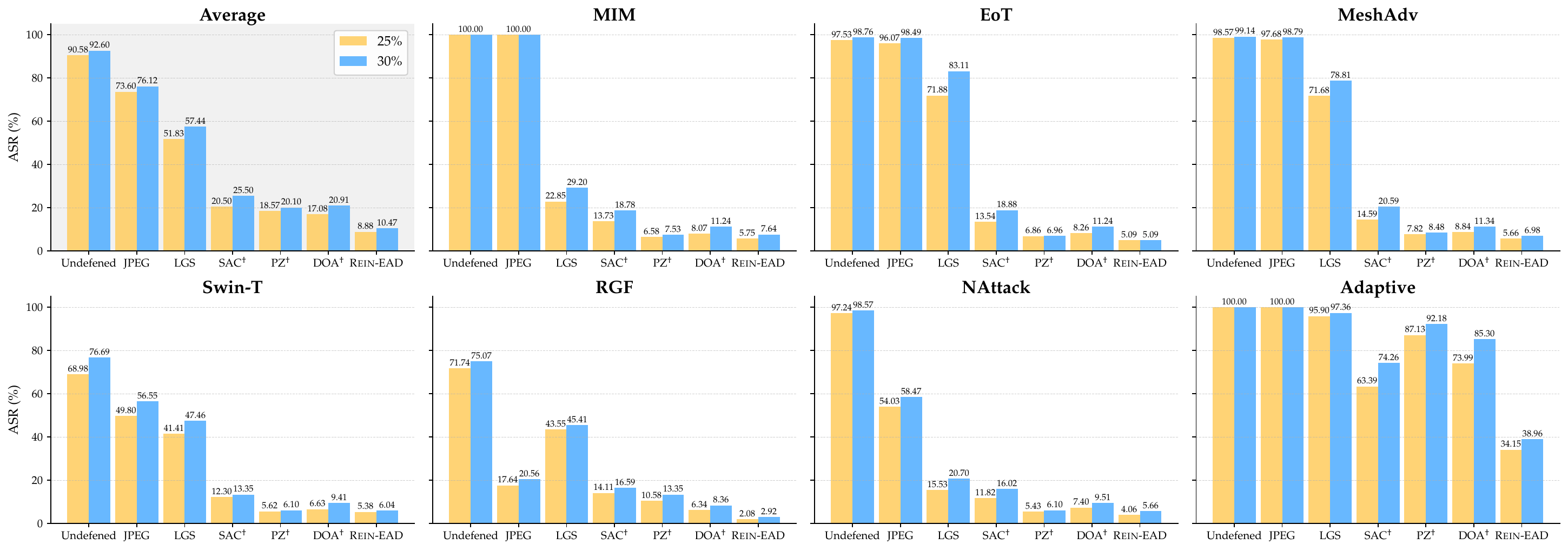}  
 \vspace{-1em}
    \caption{Evaluating generalization on 3D object classification models under attacks with different patch sizes.}
    \label{fig:cls_patch_size_bins}
    \vspace{-1em}
\end{figure*}

\noindent \textbf{Effectiveness of recurrent feedback.} We thoroughly investigate the critical role of recurrent feedback, \textit{i.e.}, reflecting on prior beliefs using a comprehensive fusion model, in achieving robust performance. In Table~\ref{tab:ablation}, even when equipped with only the perception model, \textsc{Rein}-EAD significantly surpasses both the undefended baseline and passive FR model that relies on multi-view ensembles (Random Movement). Notably, the multi-view ensemble model fails to counteract the state-of-the-art 3DAdv. This observation corroborates that \textsc{Rein}-EAD's defensive strength is not merely a function of the vulnerability of adversarial examples to viewpoint transformations. Instead, the superior performance of \textsc{Rein}-EAD can be attributed to its ability to actively explore the environment and dynamically adjust its perception.

\noindent\textbf{Impact of horizon length $H$.}
We conduct an analysis of the impact of horizon length $H$ on the performance of \textsc{Rein}-EAD in Fig.~\ref{fig:sim_step}. By examining the changes in the similarity between face pairs affected by adversarial patches, we consistently demonstrate the effectiveness of \textsc{Rein}-EAD in mitigating the detrimental effects of adversarial attacks. Specifically, for impersonation attacks, the change in similarity implies an increase, while a decrease is observed for dodging attacks. This trend suggests that the accumulation of information from additional viewpoints during the decision process effectively attenuates the issues of information loss and model hallucination engendered by adversarial patches.

\noindent \textbf{Efficiency of learned policy.} We further validate its superiority of the policy empirically by comparing the performance of EAD with two variants: EAD integrated with a random movement policy \textsc{Rand}-EAD and \textsc{Rein}-EAD. All three approaches share identical neural network architecture and parameters. Fig.~\ref{fig:sim_step} illustrates that the \textsc{Rand}-EAD cannot mitigate the adversarial effect with random exploration, even when a greater number of actions are employed. Consequently, the exploration efficiency of the random policy is significantly inferior to \textsc{Rein}-EAD. Moreover, \textsc{Rein}-EAD demonstrates a more stable and continuous deduction of adversarial effect, which can be primarily attributed to its adoption of an accumulative informative policy that consistently reduces perceptual uncertainty, instead of a greedy approach. This strategy effectively avoids the oscillations resulting from temporal inconsistency.

\begin{figure*}[ht]
    \centering
    \includegraphics[width=0.99\linewidth]{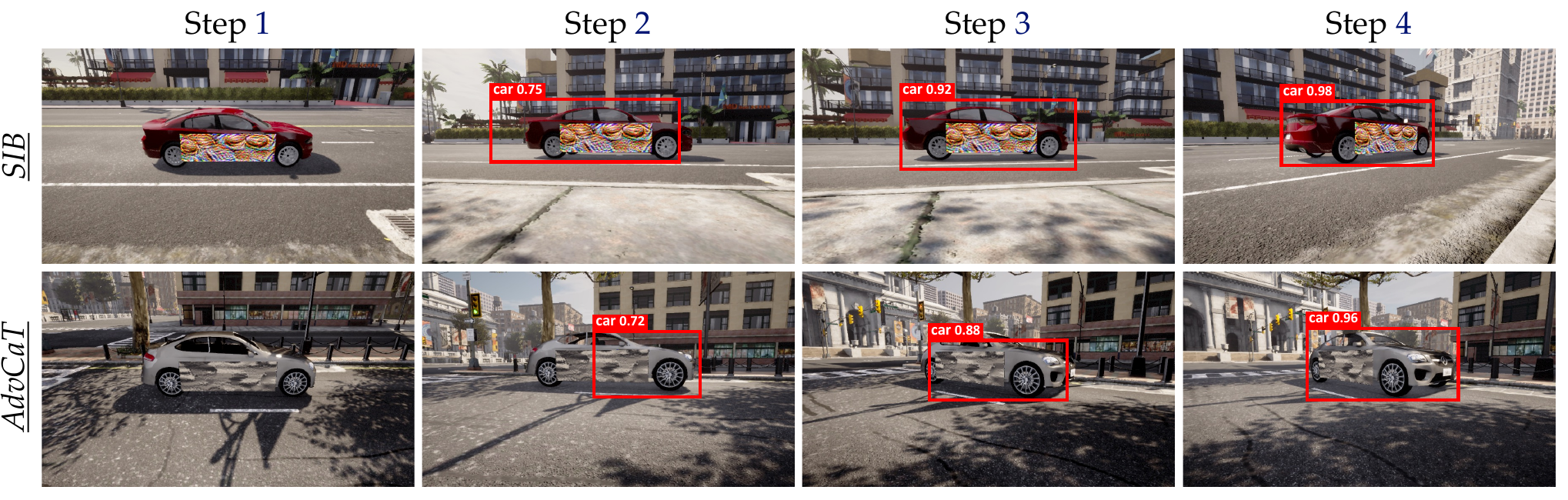}  
 \vspace{-1em}
    \caption{Qualitative results of \textsc{Rein}-EAD on CARLA.  Adversarial patches cover 25\% of the object's front-view bounding box.}
    \label{fig:visual_det}
    \vspace{-1em}
\end{figure*}

\noindent \textbf{Influence of patched data.} For \textsc{Rein}-EAD, the inherent limitations of RL in efficiently exploring the vast patch space are addressed by incorporating the OAPA algorithm.
{ We compare OAPA using default PGD with a variant using FGSM adversarial examples, which are less optimized due to taking only a single gradient step. As shown in Table~\ref{tab:ablation}, OAPA ensures that \textsc{Rein}-EAD achieves superior robustness after substantial training, with PGD outperforming FGSM and others.}

{\noindent \textbf{Stable convergence of policy training.} To address active defense challenges in dynamic 3D environments, \textsc{Rein}-EAD enhances standard PPO with two key stability features: a two-phase training approach (offline perception pretraining followed by joint online training) that accelerates convergence, and integrated supervised learning that functions as regularization to maintain perception capabilities while preventing erratic policy updates. More experiments are presented Appendix {C.9}.
}

{\noindent \textbf{Alternative reward formulations.} We have conducted additional ablation experiments to compare the performance of REIN-EAD with the uncertainty-oriented reward shaping to two alternative reward formulations, named Direct Entropy Deduction and Binary Outcome Reward. More details are presented Appendix {C.10}. As shown in Table~\ref{tab:ablation_reward}, our proposed reward shaping approach outperforms other methods in terms of both clean accuracy and adversarial robustness against patches. It employs a dense formulation that accelerates convergence and guides the model to learn a policy that maximizes information gain towards accurate perception.
}

\vspace{-1em}
\subsection{Evaluation on 3D Object Classification}
\label{subsec:classification}

\subsubsection{Experimental Settings}
\textbf{Application in non-differential environment.} 
The classification task, widely employed and inherently vulnerable to patch attacks, often relies on rendering techniques for 3D dataset synthesis due to the scarcity of 3D data~\cite{saito1990comprehensible}. However, the discrete nature of rasterization renders analytical derivation of action transitions non-differentiable~\cite{kato2020differentiable}, limiting the application of EAD that depends on a differential dynamic model. Existing differentiable rendering frameworks~\cite{liu2019soft,ravi2020pytorch3d,loper2014opendr,rhodin2015versatile} either lack precision or compromise rendering quality, failing to meet the training requirements of EAD. To address this, we introduce \textsc{Rein-EAD}, a novel framework for operating effectively in non-differentiable environments.


\noindent\textbf{Dynamic OmniObject3D.} We leverage the recently proposed OmniObject3D~\cite{wu2023omniobject3d}, the largest real-scanned 3D dataset. OmniObject3D shares numerous common classes with classic 2D datasets (e.g. ImageNet~\cite{deng2009imagenet}), making it particularly suitable for evaluating \textsc{Rein-EAD}. The environment is established by Pytorch3D~\cite{ravi2020pytorch3d} and Gym~\cite{brockman2016openai}, enabling the rendering of objects at specific viewpoints according to the agent's actions. We refer to this environment as dynamic OmniObject3D. Details of the environment establishment can be found in Appendix {D.1}.

\begin{table}[t]
\caption{{The performance of \textsc{Rein}-EAD with different reward shaping.}} 
\label{tab:ablation_reward}
\setlength\tabcolsep{7.0pt}
\renewcommand\arraystretch{1.25}
\centering
\begin{tabular}{>{\centering\arraybackslash}p{5em}|c|cccc}
\hline
\multirow{2}{*}{{Reward}} &  \multirow{2}{*}{{Acc (\%)}} & \multicolumn{4}{c}{{Attack Success Rate (\%)}}  \\ \cline{3-6} 
                                  &    &  {MIM} & {EoT} & {GenAP} & {3DAdv} \\ \hline \hline
Entropy Deduction        & 88.67 & 3.15 & 2.11 & 4.21 & 11.42  \\ 
Outcome Reward & 88.62 & 3.22 & 3.26 & 5.94 & 10.86  \\ 
\textbf{ours} & \textbf{89.03} & \textbf{2.10} & \textbf{3.15} & \textbf{7.37} & \textbf{4.21}  \\ \hline
\end{tabular}
\vspace{-1em}
\end{table}

\begin{table}[t]
\caption{The performance on object detection in EG3D.  $^\dagger$ indicates training with adversarial examples.} 
\label{tab:det_eg3d}
\vspace{-1em}
\setlength\tabcolsep{7.0pt}
\renewcommand\arraystretch{1.2}
\centering
\begin{tabular}{c|ccccc}
\hline
\multirow{2}{*}{{Method}} &  \multicolumn{5}{c}{{Average Precision (\%)}}  \\ \cline{2-6} 
                                  &  {Clean} & {EoT} & {SIB} & {UAP} & {AdvCaT} \\ \hline \hline
    Undefended  & 88.55 & 9.06  & 17.30 & 20.29 & 28.38  \\ \hline
\rowcolor{lightcyan} JPEG        & 88.40 & 11.78 & 12.46 & 13.11 & 30.20 \\
\rowcolor{lightcyan} LGS         & 87.81 & 43.15 & 34.26 & 10.01 & 57.31 \\
SAC         & 88.55 & 67.99 & 69.70 & 71.64 & 32.80 \\
PZ          & 88.55 & 80.58 & 81.32 & 81.87 & 28.65 \\ \hline
SAC$^\dagger$ & 88.55 & 70.10 & 71.08 & 74.06 & 40.67 \\
PZ$^\dagger$  & 88.55 & 85.31 & 85.43 & 83.53 & 43.36 \\
EAD & 92.50 & 91.61 & \textbf{91.47} & 91.02 & 91.34\\
\rowcolor{lightcyan} \textbf{\textsc{Rein-EAD}} & \textbf{94.26} & \textbf{92.09} & 91.45 & \textbf{91.14} & \textbf{92.13} \\ \hline
\end{tabular}
\vspace{-1em}
\end{table}


\noindent\textbf{Adversaries in the texture space.} To address adversarial threats in dynamic environments, we employ Pytorch3D to implement patch attacks on 3D mesh textures through its differential back-propagation pipeline\footnote{Pytorch3D facilitates texture differentiation but does not support action transition differentiation for training EAD.}. This approach enables rendering 3D adversarial objects as 2D images from specified viewpoints, ensuring consistent adversarial appearance across multiple views. As the attack modifies the texture space, the patch’s shape varies with perspective, challenging defense generalization. MeshAdv~\cite{xiao2019meshadv} accounts for expectations across 3D transformations in its differential rendering, suitable for multi-view 3D mesh attacks. The patch affects 20\% of the bounding box area in the object’s front view. Further information is detailed in Appendix {D.3}.

\noindent\textbf{Implementation details.} For the visual backbone, we employ the pretrained Swin Transformer (Swin-S)~\cite{liu2021swinth} on ImageNet and fine-tune it on the dynamic OmniObject3D. The weights of Swin-S are frozen in the subsequent experiments. We implement \textsc{Rein}-EAD following a paradigm similar to that used in the FR system, leveraging the OAPA algorithm for \textsc{Rein}-EAD with patches that occupy 20\% of the object bounding box.
More details are in Appendix {D.2}.

\noindent\textbf{Defense baseline.} We employ the same defense baselines as in the FR task, making necessary adaptations to the parameters to accommodate the classification settings. Notably, SAC, PZ, and DOA involve adversarial training, which is a widely adopted technique for enhancing robustness. More details are shown in Appendix {D.4}.

\subsubsection{Experimental Results}
\textbf{Effectiveness of \textsc{Rein}-EAD.} We evaluate the robustness of \textsc{Rein-EAD} and five baseline defenses under various white-box, black-box and adaptive attacks on the test set of OmniObject3D in Table~\ref{tab:object_classification}. In most cases, JPEG and LGS fail to provide an effective defense, while the other baselines provide protection under white-box and black-box settings. Notably, \textsc{Rein}-EAD significantly reduces the attack success rate of various unseen adversaries, without compromising standard accuracy. In contrast, although SAC$^\dagger$ and PZ$^\dagger$ can purify the adversarial patch with the prior knowledge of the EoT adversary, they are inferior to \textsc{Rein}-EAD due to the loss of masked features. Furthermore, the robustness gap is amplified under a stronger adaptive attack targeting the combination of classifier and defense module. We can see that the performances of passive baselines drop significantly under adaptive attack. In contrast, \textsc{Rein}-EAD achieves the strongest robustness among the baselines. These results demonstrate the effectiveness of our model-free learning method and its applicability in general non-differentiable environments. Fig.~\ref{fig:visual_classify} illustrates that even when initially deceived by adversaries, \textsc{Rein}-EAD can actively explore and observe the environment to correct and refine its predictions. 

\noindent\textbf{Generalization of \textsc{Rein}-EAD.} We further investigate the adaptability of \textsc{Rein}-EAD to various patch sizes. Specifically, the defense methods are trained using patches covering 20\% of the bounding box, and their performance is subsequently assessed with patches sized at 25\% and 30\%. As depicted in Fig.~\ref{fig:cls_patch_size_bins}, \textsc{Rein}-EAD maintains its robustness when encountering patches larger than those used in the training phase. 

\subsection{Evaluation on Object Detection}
\label{subsec:od}
\subsubsection{Experimental Settings}
Object detection in autonomous driving is more challenging than face verification and object classification, as the model must distinguish vehicles from intricate backgrounds and accurately regress bounding boxes in non-differentiable real-world environments. To tackle this, we have designed EAD and \textsc{Rein-EAD} for object detection and evaluate them on EG3D~\cite{chan2022efficient} that supports differentiable learning, and CARLA~\cite{dosovitskiy2017carla} belonging to a photorealistic environment used in autonomous driving research.

\noindent\textbf{Evaluation on differentiable EG3D.}
We adopt a differentiable generative framework by EG3D, enabling the generation of diverse vehicle types from controllable perspectives while ensuring 3D consistency. The differentiable environment allows for the verification of both EAD and \textsc{Rein}-EAD paradigms. More details can be found in Appendix {E.1}.

\begin{table}[t]
\caption{The performance on object detection model in CARLA. $^\dagger$ indicates training with adversarial examples. } 
\label{tab:det_carla}
\setlength\tabcolsep{7.0pt}
\renewcommand\arraystretch{1.25}
\centering
\begin{tabular}{c|ccccc}
\hline
\multirow{2}{*}{{Method}} &  \multicolumn{5}{c}{{Average Precision (\%)}}  \\ \cline{2-6} 
                                  &  {Clean} & {EoT} & {SIB} & {UAP} & {AdvCaT} \\ \hline \hline
            Undefended & 80.97 & 28.47 & 28.61 & 35.85 & 38.87 \\ \hline
\rowcolor{lightcyan} JPEG       & 81.57 & 36.50 & 35.80 & 34.66 & 37.28 \\
\rowcolor{lightcyan} LGS        & 80.32 & 74.73 & 72.64 & 57.76 & 51.34 \\
SAC        & 79.78 & 27.06 & 28.55 & 42.03 & 37.38 \\
PZ         & 80.70 & 62.43 & 59.35 & 49.24 & 37.90 \\ \hline
SAC$^\dagger$ & 79.28 & 31.68 & 32.95 & 30.70 & 39.09 \\
PZ$^\dagger$ & 80.91 & 76.10 & 75.50 & 75.24 & 40.51 \\ 
\rowcolor{lightcyan} \textbf{\textsc{Rein-EAD}} & \textbf{83.15} & \textbf{82.82} & \textbf{81.97} & \textbf{82.12} & \textbf{82.86} \\ \hline
\end{tabular}
\vspace{-1em}
\end{table}

\noindent\textbf{Evaluation on non-differentiable CARLA.} 
We evaluate the \textsc{Rein}-EAD's defense capabilities in CARLA, a practical and non-differentiable autonomous driving simulator. The model's robustness is assessed on 41 vehicles, encompassing all available asset blueprints in CARLA. The details about the CARLA environment are provided in Appendix {E.2}. 

\noindent\textbf{Evaluation metric and adversaries.}
The \textbf{average precision} at intersections over union thresholds from 50\% to 95\% (AP@50:95) is used as the evaluation metric. The adversarial methods include EoT~\cite{athalye2018synthesizing}, SIB~\cite{zhao2019seeing}, UAP~\cite{moosavi2017uap} and AdvCaT~\cite{hu2023advcat}, which aim to launch a hiding attack that makes the vehicle disappear from the detector. SIB impacts both hidden and final layers, perturbing features before \textsc{Rein}-EAD, while AdvCaT generates inconspicuous environmental mosaic camouflage. All attack methods use the expectation over 3D transformations to enhance patch resilience against viewpoint changes. The patch, attached to the vehicle's side, occupies $25\%$ of the initial perspective's bounding box.

\noindent\textbf{Implementation details.} The pretrained YOLOv5n~\cite{jocher2021ultralytics} serves as the visual backbone, combined with a Decision Transformer to implement \textsc{Rein}-EAD. Additional details on \textsc{Rein}-EAD can be found in Appendix {E.3}. The defense baselines follow the same approach as in previous tasks. More details can be found in Appendix {E.5}.


\subsubsection{Experimental Results}
\textbf{Effectiveness on EG3D. }
We evaluate \textsc{Rein}-EAD and other defense baselines on 200 test data samples from EG3D under four white-box attacks designed to deceive the YOLO detector. Table~\ref{tab:det_eg3d} shows that the undefended model and JPEG defense are breached by all attacks. LGS performs poorly under EoT, SIB, and UAP but moderately resists AdvCaT. SAC and PZ defend well against noisy-patterned attacks, with enhanced versions (SAC$^\dagger$ and PZ$^\dagger$) performing slightly better. PZ$^\dagger$ approximates clean average precision but fails against stealthy AdvCaT. \textsc{Rein}-EAD surpasses others in clean and robust performance, significantly outperforming passive baselines under different attacks.


\noindent\textbf{Effectiveness on CARLA.}
We evaluate the performance of \textsc{Rein}-EAD in the photo-realistic CARLA environment in a non-differential and practical setting. Table~\ref{tab:det_carla} reveals a similar trend that aligns with observations across other tasks. Specifically, LGS performs better under EoT and SIB attack, while SAC series degraded significantly due to the severe occlusion of its ‌completion‌ mechanism (visualized in Appendix {E.6}). \textsc{Rein}-EAD achieves the best results in both clean and adversarial conditions, thanks to recurrent temporal feedback. Fig.~\ref{fig:visual_det} illustrates \textsc{Rein}-EAD's defending process under SIB and AdvCaT attacks, showing improved prediction accuracy and bounding box precision with interaction and environment observation. More qualitative results are presented in Appendix {E.6}. {Furthermore, we  discuss some failure cases and analyze their implications in Appendix {E.9}.}

\begin{figure}[t]
    \centering
    \begin{minipage}{0.5\linewidth}
        \centering
        \includegraphics[width=\linewidth]{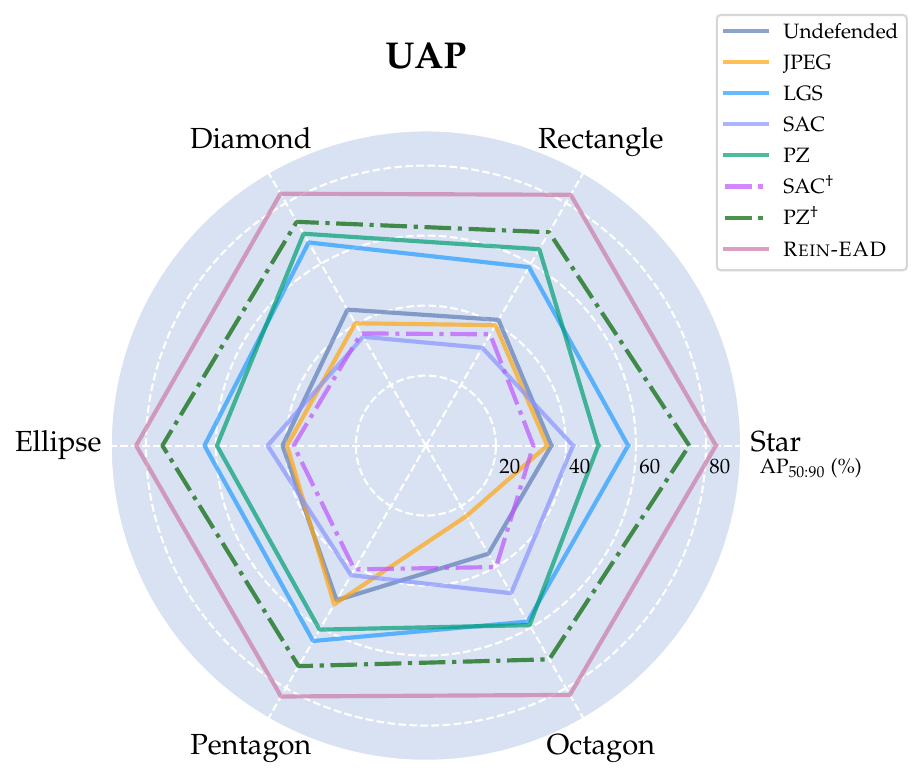}
    \end{minipage}%
    \begin{minipage}{0.5\linewidth}
        \centering
        \includegraphics[width=\linewidth]{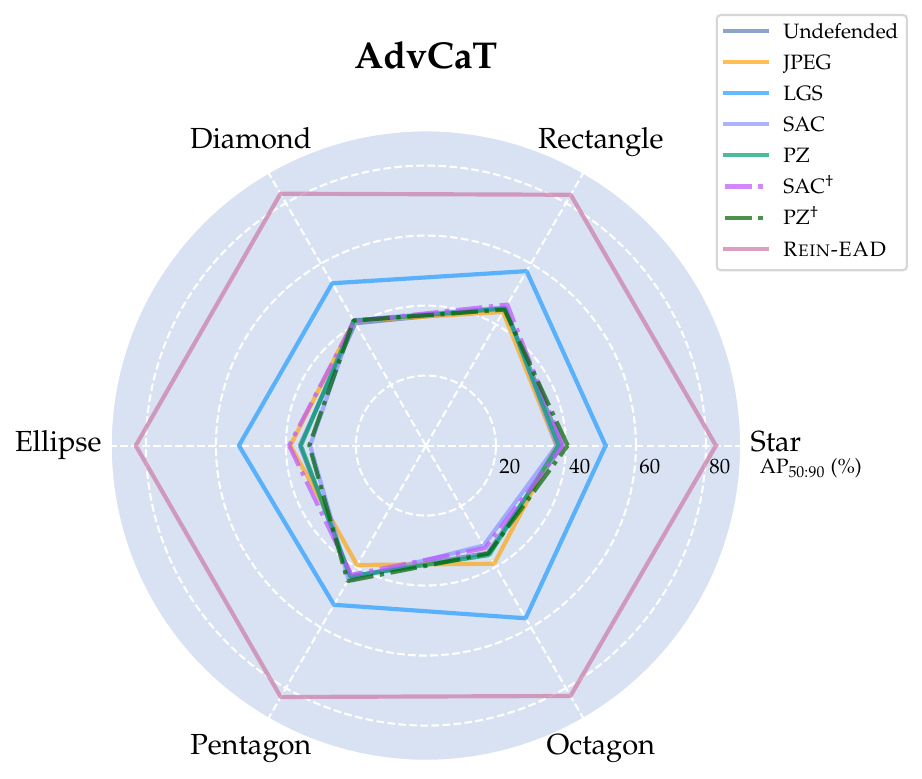}
    \end{minipage}
\caption{Comparative evaluation of object detection under different white-box attacks and patch shapes on CARLA. }
\label{fig:lidar_det}
\vspace{-2em}
\end{figure}

\noindent\textbf{Generalization on patch shapes.}
We evaluate a rectangle-trained \textsc{Rein}-EAD with adversarial patches of $6$ varying shapes while maintaining a fixed patch area occupancy.  As demonstrated in Fig.~\ref{fig:lidar_det}, our model surpasses other methods in both clean and robust accuracy, even when faced with diverse, unencountered adversarial patch shapes. More results are provided in Appendix {E.8}. These findings further underscore the exceptional generalization of \textsc{Rein}-EAD in dealing with unknown adversarial attacks.



\section{Conclusion}
In this paper, we introduce Reinforced Embodied Active Defense (\textsc{Rein}-EAD), a novel proactive defensive framework that effectively mitigates adversarial patch attacks in real-world 3D environments. \textsc{Rein}-EAD leverages exploration and interaction with the environment to contextualize environmental information and refine its understanding of the target object. It accumulates multi-step interactions for temporal consistency that balances immediate prediction accuracy with long-term entropy minimization. Moreover, \textsc{Rein}-EAD involves an uncertainty-oriented reward-shaping mechanism to improve efficiency without requiring differentiable environments. Extensive experiments demonstrate that \textsc{Rein}-EAD significantly enhances robustness and generalization and obtains strong applicability in complex tasks.

\ifCLASSOPTIONcaptionsoff
  \newpage
\fi



\bibliographystyle{IEEEtran}
\bibliography{egbib}
%
\clearpage

\appendices
\renewcommand{\theequation}{\thesection.\arabic{equation}}
\renewcommand{\thefigure}{\thesection.\arabic{figure}}
\renewcommand{\thetable}{\thesection.\arabic{table}}
\setcounter{figure}{0}
\setcounter{table}{0}

\section{Proofs and Additional Theory}

\label{sec:proofs}

\subsection{Proof of Theorem {3.1}}  %

\label{sec:pf_max_mutial_info}

\begin{proof}
    For a series of observations $\{o_1,\cdots o_{t}\}$ and a previously maintained belief $b_{t-1}$ determined by the scene $x$, we expand the left-hand side of Eq.~({6}) as follows:
    \begin{equation}
    \begin{split}
        \mathbb{E}_{x}I(o_t;y \mid b_{t-1}) 
        = \; & \mathbb{E}_{x,y}\log\frac{p(b_{t-1})p(b_{t-1},o_t, y)}{p(b_{t-1}, y) p(b_{t-1}, o_t)}  \\
        = \; & \mathbb{E}_{x,y}\log\frac{p(y \mid b_{t-1},o_{t})}{p(y \mid b_{t-1}) } \label{eqn:mutual_info_conditional}. 
    \end{split}
    \end{equation}

    By introducing the variational distribution $q_{\theta}(y \mid o_1,\cdots o_t)$ as a multiplicative factor in the integrand of Eq.~(\ref{eqn:mutual_info_conditional}), we obtain:
    \begin{equation}
    \begin{split}
         \mathbb{E}_{x} I(o_t;y \mid b_{t-1}) 
         = & \mathbb{E}_{x,y}\log\frac{p(y \mid b_{t-1}, o_{t}) q_{\theta}({y}\mid b_{t-1}  o_{t}) }{p(y \mid b_{t-1}) q_{\theta}({y}\mid b_{t-1},  o_{t})}  \\
         = \; & \mathbb{E}_{x, y} \log\frac{q_{\theta}(y \mid b_{t-1}, o_{t})}{p(y \mid b_{t-1}) }  \\
          + & \mathbb{E}_{x} \KL(p(y \mid b_{t-1}, o_{t}) \Vert q_{\theta}(y \mid b_{t-1}, o_{t})).
    \end{split}
    \end{equation}
    
    The non-negativity property of the KL-divergence allows us to establish a lower bound for the mutual information:
    \begin{equation}
    \begin{split}
        \mathbb{E}_{x} I(o_t;y \mid b_{t-1}) 
        \ge \; & \mathbb{E}_{x,y}\log\frac{q_{\theta}(y \mid b_{t-1}, o_{t})}{p(y \mid b_{t-1}) } \\
        = \; & \mathbb{E}_{x,y}\log q_{\theta}(y \mid b_{t-1}, o_{t}) + \mathcal{H}(y \mid b_{t-1}) \label{eqn:ba_bound},
    \end{split}
    \end{equation}
    where $\mathcal{H}(y \mid b_{t-1})$ represents the conditional entropy of $y$ given the belief $b_{t-1}$, and Eq. (\ref{eqn:ba_bound}) is the well-known Barber and Agakov bound \cite{barber2004algorithm}.
    We proceed by selecting an energy-based variational family that incorporates a \emph{critic} $\mathcal{E}{\theta}(b{t-1}, o_t, y)$ and is scaled by the data density $p(b_{t-1}, o_t)$:
    \begin{equation}
    \begin{split}
        \label{eqn:energy_variation}
        q_{\theta}(y \mid b_{t-1}, o_t) = \frac{p(y \mid b_{t-1})}{Z(b_{t-1}, o_t)}e^{\mathcal{E}_{\theta}(b_{t-1}, o_t, y)},
    \end{split}
    \end{equation}
    where $Z(b_{t-1}, o_t) = \mathbb{E}{y}e^{\mathcal{E}{\theta}(b_{t-1}, o_t, y)}$. Substituting the distribution defined in Eq. (\ref{eqn:energy_variation}) into Eq. (\ref{eqn:ba_bound}) yields:
    \begin{equation}
    \begin{split}
        & \mathbb{E}_{x} I(o_t;y \mid b_{t-1}) \\
        \ge \; & \mathbb{E}_{x,y}\log q_{\theta}(y \mid b_{t-1}, o_{t}) + \mathcal{H}(y \mid b_{t-1})  \\
        = \; &  \mathbb{E}_{x,y}[\mathcal{E}_{\theta}(b_{t-1}, o_t, y)] -  \mathbb{E}_{x}[\log Z(b_{t-1}, o_t)], 
    \end{split}
    \end{equation}
    which represents the unnormalized version of the Barber and Agakov bound. Applying the inequality $\log Z(b_{t-1},o_t) \le \frac{Z(b_{t-1},o_t)}{g(b_{t-1},o_t)} + \log[g(b_{t-1},o_t)] - 1$ for any $g(b_{t-1},o_t) > 0$, with the bound becoming tight when $g(b_{t-1},o_t) = Z(b_{t-1},o_t)$, we arrive at a tractable upper bound known as the tractable unnormalized version of the Barber and Agakov lower bound on mutual information:
    \begin{equation}
    \begin{split}
        &  \mathbb{E}_{x,y}[\mathcal{E}_{\theta}(b_{t-1}, o_t, y)] -  \mathbb{E}_{x}[\log Z(b_{t-1}, o_t)] \\
        \ge \; & \mathbb{E}_{x,y}[\mathcal{E}_{\theta}(b_{t-1}, o_t, y)] -  \mathbb{E}_{x}\left[\frac{\mathbb{E}_y e^{\mathcal{E}_{\theta}(b_{t-1}, o_t, y)}}{g(b_{t-1},o_t)}\right]  \\
        - & \mathbb{E}_{x} \log[g(b_{t-1},o_t)] + 1  \\
        = \; & 1 + \mathbb{E}_{x,y}\left[\log \frac{e^{\mathcal{E}_{\theta}(b_{t-1}, o_t, y)}}{g(b_{t-1},o_t)}\right] - \mathbb{E}_{x}\left[\frac{\mathbb{E}_y e^{\mathcal{E}_{\theta}(b_{t-1}, o_t, y)}}{g(b_{t-1},o_t)} \right]\label{eqn:tuba}.
    \end{split}
    \end{equation}

    To mitigate variance, we leverage multiple samples $\{x^{(j)}, y^{(j)}\}{j=1}^{K}$ from $\mathcal{D}$ to implement a low-variance, high-bias estimation of the mutual information. For an observation trajectory originating from a different scene $(x^{(j)}, y^{(j)}) \; (j \neq i)$, with annotations $\{y^{(j)}\}{j=1, j \neq i}^{K}$ independent of $x^{(i)}$ and $y^{(i)}$, we have:
    \begin{equation}
    \begin{split}
        g(b_{t-1},o_t) = g(b_{t-1},o_t;y^{(1)}, \cdots ,y^{(K)}).
    \end{split}
    \end{equation}
    This enables us to utilize the additional samples $\{x^{(j)}, y^{(j)}\}{j=1}^{K}$ to construct a Monte-Carlo estimate of the function $Z(b{t-1}, o_t)$:

    \begin{equation*}
    \begin{split}
        g(b_{t-1},o_t;y_1, \cdots y_{K}) = &m(b_{t-1},o_t;y^{(1)}, \cdots y^{(K)})\\
        = &\frac{1}{K}\sum_{j=1}^{K} e^{\mathcal{E}_{\theta}(y^{(j)}\mid b_{t-1}, o_{t})}.
    \end{split}
    \end{equation*}

     When estimating the bound over $K$ samples, the last term in Eq. (\ref{eqn:tuba}) reduces to a constant value of $1$:
    \begin{equation}
    \begin{split}
        \label{eqn:constant_term}
       & \mathbb{E}_{x}\left[\frac{\mathbb{E}_{y^{(1)}, \cdots, y^{(K)}} e^{\mathcal{E}_{\theta}(b_{t-1}, o_t, y)}}{m(b_{t-1},o_t;y^{(1)}, \cdots y^{(K)})}\right]  \\
        =  &\mathbb{E}_{x_1} \left[\frac{\frac{1}{K} \sum_{j=1}^{K} e^{\mathcal{E}_{\theta}(b_{t-1}, o_t^{(1)}, y^{(j)})}} {m(b_{t-1},o_t^{(1)};y^{(1)}, \cdots y^{(K)})}\right]
        = 1.
    \end{split}
    \end{equation}
     
    Applying Eq. (\ref{eqn:constant_term}) back to Eq. (\ref{eqn:tuba}) and averaging the bound over $K$ samples (reindexing $x^{(1)}$ as $x^{(j)}$ for each term), we precisely recover the lower bound on mutual information proposed by Oord \etal~\cite{oord2018representation}:
    \begin{equation}
    \begin{split}
        & 1 + \mathbb{E}_{x^{(j)},y^{(j)}}\left[\log \frac{e^{\mathcal{E}_{\theta}(b_{t-1}^{(j)}, y^{(j)})}}{g(b_{t-1}^{(j)};y^{(1)},\cdots, y^{(K)})}\right]\\
        &- \mathbb{E}_{x^{(j)}}\left[\frac{\mathbb{E}_{y^{(j)}} e^{\mathcal{E}_{\theta}(b_{t-1}^{(j)}, o_t^{(j)}, \hat{y}^{(j)})}}{g(b_{t-1}^{(j)},o_t^{(j)};y^{(1)},\cdots, ^{(K)})}\right]  \\
        = \; & \mathbb{E}_{x^{(j)},y^{(j)}}\left[\log \frac{e^{\mathcal{E}_{\theta}(b_{t-1}^{(j)}, o_t^{(j)}, y^{(j)})}}{g(b_{t-1}^{(j)},o_t^{(j)};y^{(1)},\cdots, y^{(K)})}\right]  \\
        = \; & \mathbb{E}_{x^{(j)},y^{(j)}}\left[\log \frac{e^{\mathcal{E}_{\theta}(b_{t-1}^{(j)}, o_t^{(j)}, y^{(j)})}}{\frac{1}{K} \sum_{\hat{y}^{(j)}} e^{\mathcal{E}_{\theta}(b_{t-1}^{(j)},  o_{t}^{(j)}, \hat{y}^{(j)})}}\right]. 
    \end{split}
    \end{equation}

    Multiplying and dividing the integrand in Eq. (\ref{eqn:mutual_info_conditional}) by $\frac{p(y \mid b_{t-1}^{(j)},o_{t-1})}{Z(b_{t-1}^{(j)}, o_t)}$ and extracting $\frac{1}{K}$ from the brackets transforms the equation into:
    \begin{equation}
    \begin{split}
         \mathbb{E}_{x^{(j)},y^{(j)}}\left[\log \frac{q_{\theta}(b_{t-1}^{(j)}, o_t^{(j)}, y^{(j)})}{\sum_{\hat{y}^{(j)}} q_{\theta}(b_{t-1}^{(j)},  o_{t}^{(j)}, \hat{y}^{(j)})}\right] + \log(K).
    \end{split}
    \end{equation}


    


Therefore, we obtain
\begin{equation}
    \begin{aligned}
    &\mathbb{E}_{(x^{(j)},y^{(j)})\sim \mathcal{D}} 
    \left[\frac{1}{K} \sum_{j=1}^K \log \frac{q_{\vtheta}(y^{(j)} \mid  b_{t-1}^{(j)}, o_{t}^{(j)})}{\frac{1}{K}\sum_{k=1}^K  q_{\vtheta}({y^{(k)}} \mid b_{t-1}^{(j)}, o_{t}^{(j)})}\right] \\ 
    \leq  & \mathbb{E}_{x} I(o_t;y \mid b_{t-1}) - \frac{\log(K)}{K}.
    \end{aligned}
\end{equation}

Given a scene annotation $y$, we quantify the uncertainty of annotation $y$ at time step $t$ using the conditional entropy of $y$ given the series of observations $\{b_{t-1}, o_t\}$, denoted as $\mathcal{H}(y \mid b_{t-1}, o_t)$. In the following, we demonstrate that the conditional mutual information $I(o_t;y \mid b_{t-1})$ is equivalent to the decrease in conditional entropy:
\begin{equation*}
    \begin{aligned}
        & I(o_t;y \mid b_{t-1}) \\
        = & I(y;b_{t-1}, o_{t}) - I(y;b_{t-1}) \\  
        = \;&[\mathcal{H}(y) - I(y;b_{t-1})] -  [\mathcal{H}(y) - I(y;b_{t-1}, o_{t})] \\
        = \; & \mathcal{H}(y \mid b_{t-1})-\mathcal{H}(y \mid b_{t-1}, o_{t}).
    \end{aligned}
\end{equation*}



Here the first equality in the derivation follows from the Kolmogorov identities \cite{polyanskiy2014lecture}.

\end{proof}
    
    



Theorem {3.1} enables us to forge a connection between the mutual information in Eq. ({6}) and the greedy informative exploration defined in Definition {3.3}. Consequently, we can infer the relationship between the policy model of EAD, which optimizes the InfoNCE objective in Eq. ({5}), and the principle of greedy informative exploration.

\subsection{Proof of Theorem {3.7}}
\label{subsec:proof_efficay_inequality}

\begin{proof}
    We aim to demonstrate that the accumulative informative policy $\pi^{*}$ results in a greater or equal reduction in entropy of $y$ compared to the greedy informative policy $\pi^g$, under the condition that the belief update function $f_b$ is bijective.






    Recall that the information gain from time $0$ to $H$ under a given policy $\pi$ is defined as the reduction in entropy of $y$:
    \[
    \Delta \mathcal{H}_{\pi} = \mathcal{H}(y) - \mathcal{H}(y \mid b_{H-1}, o_H),
    \]
    where $b_{H-1}$ and $o_H$ are the belief and observation at time $H-1$ and $H$, respectively, under policy $\pi$.

    Specifically, for the two policies:
    \[
    \Delta \mathcal{H}_{\pi^{*}} = \mathcal{H}(y) - \mathcal{H}(y \mid b_{H-1}^{*}, o_H^{*}),
    \]
    \[
    \Delta \mathcal{H}_{\pi^{g}} = \mathcal{H}(y) - \mathcal{H}(y \mid b_{H-1}^{g}, o_H^{g}).
    \]


    

    We first consider the cumulative information gain over each time step. For each time step $t = 1$ to $H$, define the incremental information gain under policy $\pi$ as:
    \begin{align}
        \Delta \mathcal{H}_t^{\pi} = &\mathcal{H}(y \mid b_{t-1}^{\pi}, o_t^{\pi}) - \mathcal{H}(y \mid b_t^{\pi}, o_{t+1}^{\pi}) \nonumber \\ 
        = &\left[\mathcal{H}(y \mid b_{t-1}^{\pi}, o_t^{\pi}) - \mathcal{H}(y \mid b_{t}^{\pi}) \right] \nonumber\\
        + &\left[\mathcal{H}(y \mid b_{t}^{\pi}) - \mathcal{H}(y \mid b_t^{\pi}, o_{t+1}^{\pi}) \right].
    \end{align} 
    Since $f_b$ is bijective, each observation $o_t$ uniquely determines the subsequent belief $b_t$, and vice versa. This bijectivity ensures that the mapping between beliefs and observations preserves information about $y$ without loss:
    \begin{equation}
        \mathcal{H}(y \mid b_{t-1}^{\pi}, o_t^{\pi}) - \mathcal{H}(y \mid b_{t}^{\pi}) = 0.
    \end{equation}
    Thereby, we have
    \begin{equation}
        \Delta \mathcal{H}_{\pi} = \sum_{t=1}^{H} \Delta \mathcal{H}_t^{\pi} = \sum_{t=1}^{H}\left[\mathcal{H}(y \mid b_{t}^{\pi}) - \mathcal{H}(y \mid b_t^{\pi}, o_{t+1}^{\pi})\right].    
    \end{equation}

    
    


    


   To this end, we compare the trajectory information gain between these two policies:
    \begin{equation*}
    \begin{split}
    & \Delta \mathcal{H}_{\pi^{*}} - \Delta \mathcal{H}_{\pi^{g}} \\
    =& \left( \sum_{t=1}^{H} \Delta \left[\mathcal{H}(y \mid b_{t}^*) - \mathcal{H}(y \mid b_t^{*}, o_{t+1}^{*})\right] \right) \\ 
    &- \left( \sum_{t=1}^{H} \Delta\left[\mathcal{H}(y \mid b_{t}^g) - \mathcal{H}(y \mid b_t^{g}, o_{t+1}^{g})\right] \right).    
    \end{split}
    \end{equation*}
    
    
    Since $\pi^{*}$ maximizes $\Delta \mathcal{H}$ considering the $H$-steps trajectory, we have:
    \begin{equation}
    \begin{split}
         &\;\left( \sum_{t=1}^{H} \Delta \left[\mathcal{H}(y \mid b_{t}^*) - \mathcal{H}(y \mid b_t^{*}, o_{t+1}^{*})\right] \right) \\ &- \left( \sum_{t=1}^{H} \Delta\left[\mathcal{H}(y \mid b_{t}^g) - \mathcal{H}(y \mid b_t^{g}, o_{t+1}^{g})\right] \right) \ge 0,
    \end{split}
    \end{equation}
    
    
    

    with equality if and only if for all $t$, $\Delta \mathcal{H}_t^{*} = \Delta \mathcal{H}_t^{g}$, which occurs precisely when the problem exhibits optimal substructure. In such cases, the greedy policy $\pi^g$ inherently accumulates information gains as effectively as the accumulative policy $\pi^{*}$.
\end{proof}

\subsection{Derivation of Reward}
\label{subsec:reward}

Given the definition of the dense reward \( r_t \) from Eq.~({11}), we substitute into the expression for trajectory reward \( \mathcal{R}(\tau)\):
\begin{equation}
\begin{split}
   \mathcal{R}(\tau) = &\sum_{t=1}^{H} \gamma^{t-1} \left[ \mathcal{L}(\hat{y}_{t-1}, y) - \gamma \cdot \mathcal{L}(\hat{y}_{t}, y) \right]\\
   = &\sum_{t=1}^{H} \gamma^{t-1} \mathcal{L}(\hat{y}_{t-1}, y) - \sum_{t=1}^{H} \gamma^{t} \mathcal{L}(\hat{y}_{t}, y). 
\end{split}
\end{equation}






Observe that the second summation now spans from \( t = 1 \) to \( t = H \), which matches the first summation's indexing shifted by one. Therefore, the two summations can be aligned as follows:

\begin{align*}
    \mathcal{R}(\tau) = &\gamma^{0} \mathcal{L}(\hat{y}_0, y) + \gamma^{1} \mathcal{L}(\hat{y}_1, y) + \cdots + \gamma^{H-1} \mathcal{L}(\hat{y}_{H-1}, y) \\
& - \left( \gamma^{1} \mathcal{L}(\hat{y}_1, y) + \gamma^{2} \mathcal{L}(\hat{y}_2, y) + \cdots + \gamma^{H} \mathcal{L}(\hat{y}_H, y) \right).
\end{align*}

All intermediate terms cancel out due to the telescoping nature of the series:

\begin{equation}
    \mathcal{R}(\tau) = \mathcal{L}(\hat{y}_0, y) - \gamma^{H} \mathcal{L}(\hat{y}_H, y).    
\end{equation}

Thus, the cumulative discounted reward $\mathcal{R}(\tau)$ is equivalent to the initial loss minus the discounted final loss. Since the initial loss $\mathcal{L}(\hat{y}_0, y)$  is solely determined by the initial state distribution $\rho_0$  and is independent of the policy, it can be considered a constant during policy optimization. This completes the proof that this form of reward aligns with the objective in Eq.(~{9}).

\section{Experiment Details for Simulation Environment}
\label{sec:sim_env}
\textbf{Environmental dynamics.} We formally define the state $s_t = (h_t, v_t)$ as a combination of the camera's yaw $h_t \in \mathbb{R}$ and pitch $v_t \in \mathbb{R}$ at moment $t$, while the action is defined as a continuous rotation denoted by $a_{t} = (\Delta h, \Delta v)$. Consequently, the transition function is expressed as $T(s_t, a_t, x) = s_t + a_t$, indicating that the next state is obtained by adding the action's rotation to the current state. The observation function is reformulated using a 3D generative model, denoted as $O(s_t,x) = \mathcal{R}(s_t,x)$, where $\mathcal{R}(\cdot)$ represents a renderer (\textit{e.g.}, a 3D generative model or graphic engine) that generates a 2D image observation $o_t$ given the camera parameters determined by the state $s_t$. 

In computational graphics, the detailed formulation for the renderer is presented as:
\begin{equation}
    o_t = \mathcal{R}'(\mE_t, \mI, x),
\end{equation}
where $\mE_t \in \mathbb{R}^{4\times 4}$ represents the camera's extrinsic matrix determined by the state $s_t$, and $\mI \in \mathbb{R}^{3\times 3}$ is the pre-defined camera intrinsic matrix. To utilize the renderer for generating 2D images, we need to calculate the camera's extrinsic matrix $\mE_t$ based on the state $s_t$. Assuming a right-handed coordinate system and column vectors, we have:
\begin{equation}
\mE_t = \begin{bmatrix}
    \mR_t & \mT \\
    \mathbf{0} & 1
\end{bmatrix},
\end{equation}
where $\mR_t \in \mathbb{R}^{3 \times 3}$ is the rotation matrix determined by $s_t$, and $\mT \in \mathbb{R}^{3 \times 1}$ is the invariant translation vector. The rotation matrices for yaw $h_t$ and pitch $v_t$ are given by:

\begin{equation}
\mR^y(h_t) = \begin{bmatrix}
    \cos(h_t) & 0 & \sin(h_t) \\
    0 & 1 & 0 \\
    -\sin(h_t) & 0 & \cos(h_t)
\end{bmatrix},
\end{equation}

\begin{equation}
\mR^x(v_t) = \begin{bmatrix}
    1 & 0 & 0 \\
    0 & \cos(v_t) & -\sin(v_t) \\
    0 & \sin(v_t) & \cos(v_t)
\end{bmatrix}.
\end{equation}

The combined rotation $\mR_t$ is obtained by multiplying the individual rotation matrices: $\mR_t = \mR^y(h_t) \times \mR^x(v_t)$. The complete extrinsic matrix is then constructed as:

\begin{equation}
  \mE_t = \begin{bmatrix}
    \mR^y(h_t) \times \mR^x(v_t) & \mT \\
    \mathbf{0} & 1
\end{bmatrix}.  
\end{equation}

\noindent \textbf{Applying function.} In the experiments, the adversarial patch is attached to a flat surface, such as eyeglasses for face recognition and billboards for object detection. By utilizing the known corner coordinates of the adversarial patch in the world coordinate system, both the extrinsic matrix $\mE_t$ and the intrinsic matrix $\mK$ are employed to render image observations containing the adversarial patch. The projection process of the 3D patch, as described by Zhu \etal~\cite{zhu2023understanding}, is followed to construct the applying function. The projection matrix $\mM_{\text{3d-2d}} \in \mathbb{R}^{4 \times 4}$ is specified as:

\begin{equation}
   \mM_{\text{3d-2d}}  = \begin{bmatrix}
    \mK &  \mathbf{0} \\
    \mathbf{0} & 1
\end{bmatrix} \times \mE_t.
\end{equation}

This process is differentiable, enabling the optimization of the adversarial patches.

In summary, a deterministic environmental model is proposed, which is applicable to all the experimental environments (\textit{e.g.}, EG3D, CARLA):

\begin{align}
\text{State} \quad & s_t = (h_t, v_t) \in \mathbb{R}^2, \nonumber \\
\text{Action} \quad & a_t = (\Delta h, \Delta v) \in \mathbb{R}^2, \nonumber \\
\text{Transition Function} \quad & T(s_t,a_t,x) = s_t + a_t, \nonumber \\
\text{Observation Function} \quad & Z(s_t,x) = \mathcal{R}(s_t,x). \nonumber
\end{align}

The primary distinction between simulations for different tasks lies in the feasible viewpoint regions. These regions are detailed in the implementation sections for each task, specifically in Appendices~\ref{sec:experiment_detail_fr},~\ref{sec:experiment_detail_oc}~\&~\ref{sec:experiment_detail_od}.

\section{Experiment Details for Face Recognition}
\label{sec:experiment_detail_fr}

\subsection{CelebA-3D}
\label{subsec:celeba_3d}

We employ an unofficial implementation of GAN Inversion with EG3D (\url{https://github.com/oneThousand1000/EG3D-projector}), utilizing default parameters to transform 2D images from the CelebA dataset~\cite{liu2018large} into 3D latent representation $w^{+}$. As the 3D generative model prior, we leverage the EG3D models pre-trained on the FFHQ dataset, which are officially released at \url{https://catalog.ngc.nvidia.com/orgs/nvidia/teams/research/models/eg3d}. To reduce computational overhead, we forgo the super-resolution module of EG3D and directly render RGB images with a resolution of $112 \times 112$ pixels using its neural renderer. 

We conduct a comprehensive evaluation of the reconstructed CelebA-3D dataset to assess its quality. Image quality is measured using PSNR, SSIM and LPIPS between the original images and the EG3D-rendered images from the same viewpoint. These metrics provide a multi-faceted assessment of the fidelity of the reconstructed images compared to their 2D counterparts. Furthermore, we introduce a modified identity consistency (ID) metric, slightly deviating from the one presented in~\cite{chan2022efficient}, to evaluate the identity consistency between the reconstructed 3D faces and their original 2D faces. Our ID metric calculates the mean ArcFace ~\cite{deng2018arcface} cosine similarity score between pairs of views of the face rendered from random camera poses and its original image from CelebA. 

\begin{figure}[t]
    \centering
    \includegraphics[width=0.99\linewidth]{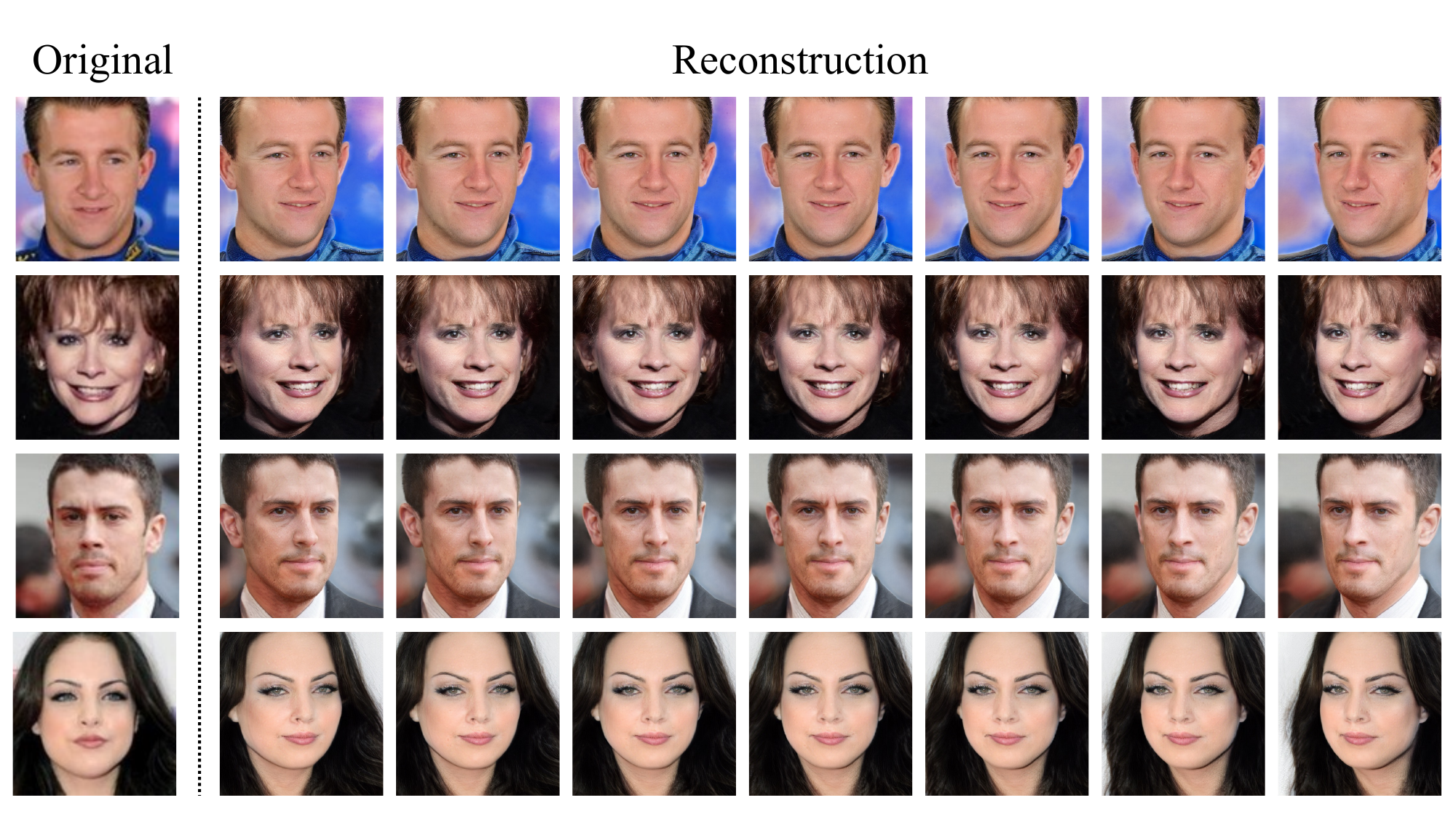}
    
    \caption{The first  column presents the source face from CelebA, and the succeeding columns demonstrate the rendered multiview faces from inverted $w^{+}$ with EG3D. The dimensions of each rendered facial image are $112 \times 112$.}
    \label{fig:celeba_3d_qual}
\end{figure}

The results presented in Table \ref{tab:celeba_3d_quan} demonstrate that the learned 3D prior over FFHQ enables remarkably high-quality single-view geometry recovery. Consequently, our reconstructed CelebA-3D dataset exhibits high image quality and sufficient identity consistency with its original 2D form, making it suitable for subsequent experiments. A selection of reconstructed multi-view faces is shown in Fig. \ref{fig:celeba_3d_qual}.

\begin{table}[t]
    \centering
    \caption{quantitative evaluation for CelebA-3D. The image size is $112 \times 112$. }
    \begin{tabular}{l c c c c}
        \toprule
        			            & {PSNR $\!\uparrow$}        & {SSIM $\!\uparrow$}   & { LPIPS $\!\downarrow$}  & {ID $\!\uparrow$} \\
        \midrule
        CelebA-3D 			& $21.28$              & $.7601$ 	& $.1314$                & $.5771$ \\
	\bottomrule
    \end{tabular}
    \label{tab:celeba_3d_quan}
\end{table}

The CelebA-3D dataset incorporates the rich annotations provided by the widely-used CelebA dataset, which can be accessed at \url{https://mmlab.ie.cuhk.edu.hk/projects/CelebA.html}

\subsection{Details for Attacks}
\label{subsec:fr_attack_details}

\textbf{Impersonation and dodging in adversarial attacks against FR system.} In adversarial attacks targeting FR systems, two primary objectives emerge impersonation and dodging. Impersonation entails the manipulation of an image to deceive FR algorithms into erroneously identifying an individual as a different person. Dodging focuses on preventing accurate identification by the FR system through strategic modifications to the image. These alterations are designed to either prevent the FR system from detecting a face altogether or to obscure the association between the detected face and its corresponding authentic identity within the system's database. The nature of these subtasks underscores the profound challenges they pose to the security and integrity of FR systems.

\noindent \textbf{Attacks in pixel space.} 
The Momentum Iterative Method (MIM)\cite{dong2018boosting} and Expectation over Transformation (EoT)\cite{athalye2018synthesizing} represent state-of-the-art techniques for refining adversarial patches within the RGB pixel space. MIM enhances the generation of adversarial examples by incorporating momentum into the optimization process, thereby facilitating the transferability of these examples across different models and settings. In contrast, EoT employs a diverse array of transformations, such as rotations and variations in illumination, to boost the robustness of attacks under physical conditions. To ensure optimal performance, we adhered to the recommended parameters as detailed in~\cite{xiao2021improving}, setting the number of iterations to $N = 150$, the learning rate to $\alpha = 1.5 / 255$, and the decay factor to $\mu = 1$. These parameters are consistently maintained across all experimental conditions. Furthermore, the sampling frequency for EoT is established at $M = 10$ to balance the computational efficiency and attack effectiveness.


\noindent \textbf{Attacks in latent space of the generative model.} 
GenAP~\cite{xiao2021improving} and Face3DAdv~\cite{yang2024face3dadv} represent pioneering approaches that shift the focus of adversarial patches from the pixel space to the latent space of a Generative Adversarial Network (GAN). By utilizing the generative capabilities, GANs develop adversarial patches that can deceive facial recognition systems, accounting for 3D variations. Additionally, we employ the Adam optimizer~\cite{kingma2014adam} for the latent space of EG3D for patch optimization, with a learning rate of $\eta = 0.01$ and an iteration count of $N = 150$. The sampling frequency is set at $M = 10$.


\subsection{Details for Adaptive Attacks}
\label{subsec:details_for_adaptive_attack}

\noindent \textbf{Adaptive attack for defense baselines.} 
To launch adaptive attacks against parameter-free, purification-based defenses such as JPEG and LGS, we employ Backward Pass Differentiable Approximation (BPDA) as proposed by Athalye \etal~\cite{athalye2018obfuscated}. This method assumes that the output from each defense mechanism closely approximates the original input. For adaptive attacks on SAC and PZ, we use their official implementations~\cite{liu2022segment}, incorporating Straight-Through Estimators (STE)~\cite{bengio2013estimating} for backpropagation through thresholding operations.

\noindent \textbf{Adaptive attack with uniform superset policy.}  In adaptive attacks for \textsc{Rein}-EAD, we leverage uniform superset approximation for the policy model. Thus, we have the surrogate policy
\begin{equation}
    \Tilde{\pi} \coloneqq \mathcal{U}(h_{\text{min}}, h_{\text{max}}) \times \mathcal{U}(v_{\text{min}}, v_{\text{max}}),
\end{equation}
where $a_t \in [h_{\text{min}}, h_{\text{max}}] \times [v_{\text{min}}, v_{\text{max}}]$, and $[h_{\text{min}}, h_{\text{max}}]$, $[v_{\text{min}}, v_{\text{max}}]$ separately denotes the pre-defined feasible region for horizontal rotation (yaw) and vertical rotation (pitch).
The optimization objective is outlined as follows, with a simplified sequential representation for clarity\footnote{The recurrent inference procedure is presented sequentially in this section for simplicity.}:


\begin{equation}
    \begin{aligned}
        \max_{p} \quad & \mathbb{E}_{s_0 \sim \rho_0, a_i \sim \Tilde{\pi} }\mathcal{L}(\overline{y}_{\tau}, y), \\
        \textrm{with} \quad & \{\overline{y}_\tau, b_{\tau}\} = f(\{A(o_i,p;s_0 + \sum_{j=0}^{i-1}a_{j})\}_{i=0}^{\tau};\vtheta)\\
        \textrm{s.t.} \quad & p \in [0, 1]^{H_p \times W_p \times C},
    \end{aligned}
\end{equation}
where $\rho_0$ denotes the distribution of initial state $s_0$, and $\mathcal{L}$ is the task-specific loss function. 

\noindent \textbf{Adaptive Attack Against Sub-Modules.} An end-to-end attack may not always be the most effective strategy, particularly against defenses with complex forward passes. Targeting the weakest component is often sufficient. Therefore, we propose two separate adaptive attacks: one against the perception model and another against the policy model. The attack on the perception model aims to generate an adversarial patch that corrupts the internal belief $b_t$~\cite{sabour2015adversarial}. The optimization objective for this attack is to maximize the Euclidean distance between the corrupted belief ${b}{\tau}$ and the benign belief ${b}{\tau}^{+}$, formulated as follows:

\begin{equation}
    \begin{aligned}
        \max_{p} \quad & \mathbb{E}_{s_0 \sim \rho_0, a_i \sim \Tilde{\pi} } \|b_{\tau} -  b_{\tau}^{+}\|_2^2, \\
        \textrm{with} \quad & \{\overline{y}_\tau, b_{\tau}\} = f(\{A(o_i,p;s_0 + \sum_{j=0}^{i-1}a_{j})\}_{i=1}^{\tau};\vtheta), \\
         \quad & \{\overline{y}_\tau^{+}, b_{\tau}^{+}\} = f(\{o_i\}_{i=0}^{\tau};\vtheta), \\
        \textrm{s.t.} \quad & p \in [0, 1]^{H_p \times W_p \times C}.
    \end{aligned}
\end{equation}

For the attack against the policy model, the goal is to create an adversarial patch that induces the policy model to output a zero action $a_i = \pi(b_i; \vphi) = 0$, thereby keeping the model stationary with an invariant state $s_i = s_1$ and generating erroneous predictions $\overline{y}_{\tau}$. While the original problem can be challenging with policy output as a constraint, we employ Lagrangian relaxation to incorporate the constraint into the objective and address the following problem:

\begin{equation}
    \begin{aligned}
        \max_{p} \quad & \mathbb{E}_{s_0 \sim \rho_0 }\mathcal{L}(\overline{y}_{\tau}, y) + c \cdot \|\pi(\{A(o_i,p;s_i)\}_{i=0}^{\tau}; \vphi)\|_2^2, \\
        \textrm{with}\quad & \{\overline{y}_\tau, b_{\tau}\} = f(\{A(o_i,p;s_i)\}_{i=0}^{\tau};\vtheta) \\
        \textrm{s.t.} \quad & p \in [0, 1]^{H_p \times W_p \times C},
    \end{aligned}
\end{equation}
where $c > 0$ is a constant that yields an adversarial example ensuring the model outputs zero actions.

\noindent \textbf{Adaptive Attack for the Entire Pipeline.} Attacking the model through backpropagation is infeasible due to the rapid consumption of GPU memory as trajectory length increases (\textit{e.g.}, 4 steps require nearly 90 GB of video memory). To mitigate this, we use gradient checkpointing~\cite{chen2016training} to reduce memory consumption. By selectively recomputing parts of the computation graph defined by the $H$-step inference procedure, instead of storing them, this technique effectively reduces memory costs at the expense of additional computation. Using this method, we successfully attack the entire pipeline along a 4-step trajectory using an NVIDIA RTX 3090 Ti, but still fail to extend it to attack 16-step \textsc{Rein}-EAD.

Regarding implementation, we adopt the same hyper-parameters as Face3DAdv and consider the action bounds defined in Appendix~\ref{subsec:fr_ead_details}. For the constant $c$ in the adaptive attack against the policy model, we employ a bisection search to identify the optimal value as per Carlini~\etal~\cite{carlini2017towards}, finding $c = 100$ to be the most effective. Additionally, the evaluation results from these adaptive attacks and analysis are detailed in Appendix~\ref{subsec:more_fr_result} with main results in Table~\ref{tab:more_adaptive_attack}.


\subsection{Details for Defenses}
\label{subsec:fr_defense_details}


\noindent \textbf{JPEG compression.} we set the quality parameter to $75$.

\noindent \textbf{Local gradients smoothing.} We adopt the implementation at \url{https://github.com/fabiobrau/local_gradients_smoothing}, and maintain the default hyper-parameters claimed in \cite{naseer2019local}.

\noindent \textbf{Segment and complete.} We use official implementation at \url{https://github.com/joellliu/SegmentAndComplete}, and retrain the patch segmenter with adversarial patches optimized by EoT~\cite{athalye2018obfuscated} and USAP technique separately. We adopt the same hyper-parameters and training process claimed in~\cite{liu2022segment}, except for the prior patch sizes, which we resize them proportionally to the input image size.

\noindent \textbf{Patchzero.} For Patchzero~\cite{xu2023patchzero}, we directly utilize the trained patch segmenter of SAC, for they share almost the same training pipeline. 

\noindent \textbf{Defense against Occlusion Attacks.} DOA is an adversarial training-based method. We adopt the DOA training paradigm to fine-tune the same pre-trained IResNet-50 and training data as \textsc{Rein}-EAD with the code at \url{https://github.com/P2333/Bag-of-Tricks-for-AT}. As for hyper-parameters, we adopt the default ones in~\cite{wu2019defending} with the training patch size scaled proportionally. 

\subsection{Details for Implementations}
\label{subsec:fr_ead_details} 


\noindent \textbf{Model details.} In the context of face recognition, we implement \textsc{Rein}-EAD model with a composition of a pre-trained face recognition feature extractor and a Decision transformer, specifically, we select  IResNet-50 as the the visual backbone to extract feature. For each time step $t > 0$, we use the IResNet-50 to map the current observation input of dimensions $112 \times 112 \times 3$ into an embedding with a dimensionality of $512$. This embedding is then concatenated with the previously extracted embedding sequence of dimensions $(t-1) \times 512$, thus forming the temporal sequence of observation embeddings ($t \times 512$) for Decision Transformer input. The Decision Transformer subsequently outputs the temporal-fused face embedding for inference as well as the predicted action. For the training process, we further map the fused face embedding into logits using a linear projection layer. For the sake of simplicity, we directly employ the Softmax loss function for model training.

 In experiments, we use pre-trained IResNet-50 with ArcFace margin on MS1MV3~\cite{guo2016ms} from \texttt{InsightFace}~\cite{deng2018arcface}, which is available at \url{https://github.com/deepinsight/insightface/tree/master/model_zoo}.

\noindent \textbf{Curriculum training.}
It's observed that the training suffers from considerable instability when simultaneously training perception and policy models from scratch. A primary concern is that the perception model, in its early training stages, cannot provide accurate supervision signals, leading the policy network to generate irrational actions and hindering the overall learning process. To mitigate this issue, we initially train the perception model independently using frames obtained from a random action policy, namely \emph{offline phase}. Once achieving a stable performance from the perception model, we proceed to the \emph{online phase} and jointly train both the perception and policy networks, employing Algorithm 1,
thereby ensuring their effective coordination and learning. 

Meanwhile, learning offline with pre-collected data in the first phase proves to be significantly more efficient than online learning through interactive data collection from the environment. By dividing the training process into two distinct phases \emph{offline} and \emph{online}, we substantially enhance training efficiency and reduce computational costs.

\textbf{Training details.} To train \textsc{Rein}-EAD for face recognition, we randomly sample images from $2,500$ distinct identities from the training set of CelebA-3D. we adopt the previously demonstrated phased training paradigm with hyper-parameters listed in Table \ref{tab:param-fr}.

\begin{table}[t]
  \renewcommand{\arraystretch}{1.25}
  \centering
  \small
  \caption{Hyper-parameters of \textsc{Rein}-EAD for face recognition.}
  \label{tab:param-fr}
  \begin{tabular}{l l}
  \toprule
    \textbf{Hyper-parameter}          & \textbf{Value}              \\ 
    \midrule
    Lower bound for horizontal rotation ($h_{\text{min}}$) & $-0.35$ \\
    Upper bound for horizontal rotation ($h_{\text{max}})$ & $0.35$ \\
    Lower bound for vertical rotation ($v_{\text{min}}$) & $-0.25$ \\
    Upper bound for vertical rotation ($v_{\text{max}}$) & $0.25$ \\ 
    Ratio of patched data ($r_{\text{patch}}$) & $0.4$ \\
    Training epochs for offline phase ($\mathrm{lr}_{\text{offline}}$) & $50$ \\
    learning rate for offline phase ($\mathrm{lr}_{\text{offline}}$) & $10^{-3}$ \\
    batch size for offline phase ($b_{\text{offline}}$) & $64$ \\
    Learning rate for online phase ($\mathrm{lr}_{\text{online}}$) & $2.5\times 10^{-4}$ \\
    Batch size for online phase ($b_{\text{online}}$) & $256$ \\
    Return attenuation factor ($\gamma$) & $0.95$ \\
    Updates per iteration ($n$) & $2$ \\
    \bottomrule
  \end{tabular}%
\end{table}

\subsection{Computational Overhead}
\label{subsec:overhead}


This section evaluates our method's computational overhead compared to other passive defense baselines in facial recognition systems. Performance evaluation is performed on an NVIDIA GeForce RTX 3090 Ti and an AMD EPYC 7302 16-Core Processor, using a training batch size of 64. SAC and PZ necessitate training a segmenter to identify the patch area, entailing two stages: initial training with pre-generated adversarial images and subsequent self-adversarial training~\cite{liu2022segment,xu2023patchzero}. DOA, an adversarial training-based approach, requires retraining the feature extractor~\cite{wu2019defending}. Additionally, \textsc{Rein}-EAD's training involves offline and online phases, without involving adversarial training.


As indicated in Table \ref{tab:running_time}, although differential rendering imposes significant computational demands during the online training phase, the total training time of our \textsc{Rein}-EAD model is effectively balanced between the pure adversarial training method DOA and the partially adversarial methods like SAC and PZ. This efficiency stems mainly from our unique USAP approach, which bypasses the need to generate adversarial examples, thereby boosting training efficiency.
In terms of model inference, our \textsc{Rein}-EAD, along with PZ and DOA, demonstrates superior speed compared to LGS and SAC. This is attributed to the latter methods requiring CPU-intensive, rule-based image preprocessing, which diminishes their inference efficiency.

\begin{table*}[t] 
		\centering
        \small
        \setlength{\tabcolsep}{3.2pt}
         \renewcommand{\arraystretch}{1.25}
        \caption{Computational overhead comparison of different defense methods in face recognition. We report the training and inference time of defense on a NVIDIA GeForce RTX 3090 Ti and an AMD EPYC 7302 16-Core Processor with the training batch size as $64$.} 
        \label{tab:running_time}
		\begin{tabularx}{\textwidth}{c| c CCCCCC}
            \hline
            
             {Method} & {\# Params (M)} & {Parametric Model} & {Training Epochs} &{Training Time per Batch (s)} &  {Total Training Time (GPU hours)} & {Inference Time per Instance (ms)} \\
             \hline \hline
             JPEG & - & non-parametric & - & - & - & 9.65 \\
             LGS & - & non-parametric & - & - &  - & 26.22  \\
             \hline
             SAC   & \multirow{2}{*}{44.71} & \multirow{2}{*}{segmenter} & \multirow{2}{*}{50 + 10} & \multirow{2}{*}{0.152 / 4.018}  &  \multirow{2}{*}{104} & 26.43 \\
               PZ &  &  &  &  &  & 11.88 \\
             DOA & 43.63 & feature extractor & 100 & 1.732 &  376 & 8.10 \\
             
               \hline
             EAD & 57.30 &  \multirow{2}{*}{\shortstack[c]{Perception \& \\ Policy Model}} & 50 + 50 &  0.595 / 1.021 & 175 & \multirow{2}{*}{11.51} \\ 
             \textsc{Rein}-EAD & 57.52 & & 50 + 40 & 0.595 / 1.096  & 188 &  \\ \hline
           
    		\end{tabularx}
	\end{table*}

Regarding detailed training, the \textsc{Rein}-EAD model was trained following the configuration in Appendix~\ref{subsec:fr_ead_details}. The offline training utilized 4 NVIDIA GeForce RTX 3090 Ti for approximately 2.5 hours (150 minutes). Due to the increased sampling requirements of reinforcement learning, the online training phase was extended to about 12 hours (728 minutes) and utilized eight NVIDIA GeForce RTX 3090 Ti GPUs.

{ Regarding the computational costs of its key components, the end-to-end inference pipeline achieves 11.5 ms per instance on the face recognition task. The perception model accounts for 98.4\% of this processing time (IResNet50 backbone: 8.13 ms; causal transformer: 3.19 ms), while the lightweight policy MLP requires only 0.18 ms. Besides, the Offline Patch Approximation (OAPA) stage occurs before training, and does not add computational burden to either the training or inference phases. The entire training set of approximately 50,000 instances was processed in an acceptable timeframe of around 20 minutes using eight NVIDIA RTX 3090 GPUs via batch processing. The results show that the perception model accounts for the majority of the computational costs in REIN-EAD during inference. These findings suggest that future work on optimizing the computational efficiency of REIN-EAD should focus primarily on the perception model.}



\subsection{More Evaluation Results}
\label{subsec:more_fr_result}

\noindent \textbf{Evaluation with different patch sizes.} 
To further assess the generalizability of the \textsc{Rein}-EAD model across varying patch sizes and attack methods, we conduct experiments featuring both impersonation and dodging attacks. These attacks share similarities with the setup illustrated in Table {1}. Although with different patch sizes, the results in Table~\ref{tab:white_attack} and Table~\ref{tab:white_dod_attack} bear a considerable resemblance to those displayed in Table {1}. This congruence further supports the adaptability of the \textsc{Rein}-EAD model in tackling unseen attack methods and accommodating diverse patch sizes.

\begin{table*}[t]
\caption{The \textbf{white-box attack success rates} (\%) on face recognition models with different patch sizes. $^\dagger$ denotes methods are trained with adversarial examples.}
\setlength\tabcolsep{6.75pt}
\renewcommand\arraystretch{1.25}
\centering
\begin{tabular}{c|cccc|cccc|cccc}
\hline
\multirow{2}{*}{Method} & \multicolumn{4}{c|}{8 \%} & \multicolumn{4}{c|}{10 \%} & \multicolumn{4}{c}{12 \%}  \\ \cline{2-13} 
                                  & MIM & EoT & GenAP & 3DAdv & MIM & EoT & GenAP & 3DAdv & MIM & EoT & GenAP & 3DAdv \\ \hline \hline
\multicolumn{13}{c}{\textit{Impersonation Attack}} \\ \hline
        Undefended                        & 100.0      & 100.0      & 99.00        & 98.00        & 100.0       & 100.0       & 100.0        & 99.00       & 100.0      & 100.0       & 100.0        & 99.00       \\
\hline
JPEG                               & 99.00      & 100.0      & 99.00        & 93.00        & 100.0       & 100.0       & 99.00        & 99.00       & 100.0      & 100.0       & 99.00        & 99.00       \\
LGS                                & 5.10       & 7.21       & 33.67        & 30.61        & 6.19        & 7.29        & 41.23        & 36.08       & 7.21       & 12.37       & 61.85        & 49.48       \\
SAC                                & 6.06       & 9.09       & 67.68        & 64.64        & \textbf{1.01} & 3.03      & 67.34        & 63.26       & 5.05       & 4.08        & 69.70        & 66.32       \\
PZ                                 & 4.17       & 5.21       & 59.38        & 45.83        & 2.08        & 3.13        & 60.63        & 58.51       & 4.17       & \textbf{3.13} & 60.63        & 58.33       \\
\hline
SAC$^\dagger$                      & 3.16       & 3.16       & 18.94        & 22.11        & 2.10        & 3.16        & 21.05        & 16.84       & 3.16       & 4.21        & 15.78        & 18.95       \\
PZ$^\dagger$                       & 3.13 & 3.16   & 19.14        & 27.37        & 2.11        & 3.13        & 20.00        & 30.53       & 5.26       & 5.26        & 18.95        & 28.42       \\
DOA$^\dagger$                      & 95.50      & 89.89      & 96.63        & 89.89        & 95.50       & 93.26       & 100.0        & 96.63       & 94.38      & 93.26       & 100.0        & 100.0       \\
\hline
\textbf{EAD}                & 4.12       & 3.09 & 5.15 & \textbf{7.21} & 3.09     & 2.06 & \textbf{4.17} & \textbf{8.33} & \textbf{3.09} & 5.15      & \textbf{8.33} & \textbf{10.42} \\ 
\textbf{\textsc{Rein}-EAD} & \textbf{2.10} & \textbf{1.06} & \textbf{3.15} & 7.37 & 1.04 &\textbf{ 1.06} & 6.32 & 9.47 & 3.15 & 4.21 & 7.37 & 15.79 \\ \hline
\multicolumn{13}{c}{\textit{Dodging Attack}} \\ \hline
Undefended & 100.0 & 100.0 & 99.00 & 89.00 & 100.0 & 100.0 & 100.0 & 95.00 & 100.0 & 100.0 & 100.0 & 99.00 \\ \hline
 JPEG & 98.00 & 99.00 & 95.00 & 88.00 & 100.0 & 100.0 & 99.00 & 95.00 & 100.0 & 100.0 & 100.0 & 98.00 \\
 LGS & 49.47 & 52.63 & 74.00 & 77.89 & 48.93 & 52.63 & 89.47 & 75.78 & 55.78 & 54.73 & 100.0 & 89.47 \\
 SAC & 73.46 & 73.20 & 92.85 & 78.57 & 80.06 & 78.57 & 92.85 & 91.83 & 76.53 & 77.55 & 92.85 & 92.92 \\
 PZ & 6.89 & 8.04 & 58.44 & 57.14 & 8.04 & 8.04 & 60.52 & 65.78 & 13.79 & 12.64 & 68.49 & 75.71 \\ \hline

         SAC$^\dagger$ & 78.78 & 78.57 & 79.59 & 85.85 & 81.65 & 80.80 & 82.82 & 86.73 & 80.61 & 84.69 & 87.87 & 87.75 \\
         PZ$^\dagger$ & 6.12 & 6.25 & 14.29 & 20.41 & 7.14 & 6.12 & 21.43 & 25.51 & 11.22 & 10.20 & 24.49 & \textbf{30.61} \\

         DOA$^\dagger$ & 75.28 & 67.42 & 87.64 & 95.51 & 78.65 & 75.28 & 97.75 & 98.88 & 80.90 & 82.02 & 94.38 & 100.0 \\ \hline
        \textbf{EAD} & \textbf{0.00} & \textbf{0.00} & \textbf{2.10} & \textbf{13.68} & \textbf{2.11} & \textbf{1.05} & \textbf{6.32} & \textbf{16.84} & \textbf{2.10} & \textbf{3.16} & \textbf{12.64} & 34.84 \\
        \textbf{\textsc{Rein}-EAD} & 1.04 & 2.04 & 5.15 & 13.54 & 1.03 & 2.02 & 8.16 & 14.58 & 6.18 & 5.15 & 13.54 & 34.02 \\ \hline
\end{tabular}
\label{tab:white_attack}
\end{table*}

\begin{table*}[t]
\caption{The \textbf{black-box attack success rates} (\%) on face recognition models with different patch sizes. $^\dagger$ denotes methods are trained with adversarial examples.}
\setlength\tabcolsep{6.5pt}
\renewcommand\arraystretch{1.25}
\centering
\begin{tabular}{c|cccc|cccc|cccc}
\hline
\multirow{2}{*}{Method} & \multicolumn{4}{c|}{8 \%} & \multicolumn{4}{c|}{10 \%} & \multicolumn{4}{c}{12 \%}  \\ \cline{2-13} 
                                  & Cos. & Softmax & NAttack & RGF & Cos. & Softmax & NAttack & RGF & Cos. & Softmax & NAttack & RGF \\ \hline \hline
\multicolumn{13}{c}{\textit{Impersonation Attack}} \\ \hline
Undefended & 28.00 & 23.00 & 100.00 & 100.00 & 41.00 & 36.00 & 100.00 & 100.00 & 55.00 & 45.00 & 100.00 & 100.00 \\ \hline
JPEG & 33.00 & 33.00 & 96.00 & 94.00 & 42.00 & 40.00 & 98.00 & 98.00 & 62.00 & 48.00 & 99.00 & 99.00 \\
LGS & 11.63 & 6.98 & 11.63 & 4.65 & 12.79 & 11.63 & 9.30 & 3.49 & 15.12 & 17.44 & 9.30 & 4.65 \\
SAC & 8.70 & 9.78 & 13.04 & 14.13 & 13.04 & 11.97 & 14.13 & 15.22 & 17.39 & 13.04 & 13.04 & 17.39 \\
PZ  & 6.45 & 9.68 & 4.30 & 3.26 & 5.38 & 7.63 & 4.30 & 4.31 & 8.60 & 5.52 & 4.30 & 4.15 \\ \hline
SAC$^\dagger$ & 11.11 & 12.36 & 12.22 & 14.44 & 10.00 & 19.10 & 13.33 & 15.56 & 12.22 & 17.98 & 15.56 & 14.44 \\ 
PZ$^\dagger$ & 8.24 & 5.00 & 10.58 & 9.41 & 10.59 & 3.75 & 9.41 & 8.25 & 11.77 & \textbf{5.00} & 9.41 & 8.25 \\
DOA$^\dagger$ & 15.73 & 17.97 & 34.83 & 16.86 & 15.73 & 14.61 & 32.58 & 15.72 & 16.85 & 17.98 & 31.46 & 10.11 \\ \hline
\textbf{EAD} & 4.17 & 5.20 & 4.12 & 4.12 & 5.21 & 5.21 & \textbf{3.09} & 3.08 & 7.29 & 5.21 & \textbf{2.06} & \textbf{4.12} \\ 
\textbf{\textsc{Rein}-EAD} & \textbf{2.10} & \textbf{2.08} & \textbf{1.05} & \textbf{2.10} & \textbf{4.21} & \textbf{3.16} & 3.12 & \textbf{2.10} & \textbf{5.26} & 8.33 & 4.21 & 4.21 \\ \hline
\multicolumn{13}{c}{\textit{Dodging Attack}} \\ \hline
Undefended & 44.00 & 35.00 & 96.00 & 96.00 & 53.00 & 43.00 & 100.00 & 99.00 & 68.00 & 61.00 & 100.00 & 100.00 \\ \hline
JPEG & 49.00 & 45.00 & 81.00 & 83.00 & 58.00 & 51.00 & 94.00 & 96.00 & 71.00 & 72.00 & 99.00 & 99.00 \\
LGS & 22.11 & 21.05 & 18.95 & 20.00 & 22.11 & 21.05 & 20.00 & 16.84 & 29.48 & 31.58 & 20.00 & 24.21 \\
SAC & 40.80 & 36.84 & 55.26 & 50.00 & 47.37 & 44.74 & 53.95 & 52.63 & 50.00 & 50.00 & 59.21 & 47.37 \\
PZ & 41.67 & 28.34 & 28.33 & 31.67 & 38.33 & 36.67 & 28.33 & 26.67 & 41.67 & 28.33 & 30.00 & 30.00 \\ \hline
SAC$^\dagger$ & 47.46 & 43.54 & 55.93 & 62.71 & 44.07 & 50.00 & 47.46 & 66.10 & 44.07 & 46.77 & 44.07 & 57.63 \\
PZ$^\dagger$ & 50.88 & 47.69 & 56.14 & 50.87 & 47.37 & 52.30 & 50.88 & 56.14 & 43.86 & 53.84 & 52.63 & 49.12 \\
DOA$^\dagger$ & 30.33 & 31.46 & 53.93 & 28.09 & 30.33 & 31.46 & 61.80 & 25.85 & 31.46 & 34.83 & 62.93 & 23.60 \\ \hline
\textbf{EAD} & 5.26 & 7.36 & 1.05 & \textbf{0.00} & \textbf{4.21} & \textbf{6.31} & \textbf{0.00} & \textbf{0.00} & \textbf{10.52} & 16.84 & \textbf{0.00} & 3.16 \\
\textbf{\textsc{Rein}-EAD} & \textbf{4.17} & \textbf{7.29} & \textbf{1.03} & \textbf{0.00} & 6.25 & 8.42 & 1.03 & 2.06 & 12.50 & \textbf{13.54} & 5.15 & \textbf{3.06} \\ \hline

\end{tabular}
\label{tab:white_dod_attack}
\end{table*}



\noindent \textbf{Evaluation with different adaptive attacks.} As Table \ref{tab:more_adaptive_attack} demonstrates, our original adaptive attack using USP was more effective than tracing the authentic policy of EAD (overall). This may be attributed to vanishing or exploding gradients~\cite{athalye2018obfuscated} that impede optimization. This problem is potentially mitigated by our approach to computing expectations over a uniform policy distribution. In the meantime, The results reaffirm the robustness of \textsc{Rein}-EAD against a spectrum of adaptive attacks. It further shows that \textsc{Rein}-EAD’s defensive capabilities arise from the synergistic integration of its policy and perception models, facilitating strategic observation collection rather than learning a short-cut strategy to neutralize adversarial patches from specific viewpoints.

   

\begin{table}[t]
\noindent \caption{Evaluation of adaptive attacks on EAD and \textsc{Rein}-EAD. Columns with \emph{USP} represent results obtained by optimizing the patch with expected gradients over the Uniform Superset Policy (USP). And \emph{perception} and \emph{policy} separately represent adaptive attacks against a single sub-module. And \emph{overall} denotes attacking the model by following the gradients for along the overall (4 steps) trajectory with gradient-checkpointing.} 
\setlength\tabcolsep{6.75pt}
\renewcommand\arraystretch{1.25}
\centering
\begin{tabular}{c|cccc}
\hline
\multirow{2}{*}{Method} &  \multicolumn{4}{c}{Attack Success Rate (\%)}  \\ \cline{2-5} 
                                  &  USP & Perception & Policy & Overall  \\ \hline \hline
  \multicolumn{5}{c}{\textit{Impersonation Attack}} \\ \hline
  EAD & 8.33 & 1.04 & \textbf{9.38} & 7.29 \\
  \textsc{Rein}-EAD & \textbf{4.21} & 2.11 & 2.17 & - \\ \hline
   \multicolumn{5}{c}{\textit{Dodging Attack}} \\ \hline
     EAD & \textbf{22.11} & 10.11 & 16.84 & 15.79 \\
  \textsc{Rein}-EAD & \textbf{8.16} & 3.06 & 2.06 & - \\ \hline
\end{tabular}

\label{tab:more_adaptive_attack}
\end{table}

\subsection{More qualitative results}

\begin{figure}[t]
    \centering
 \includegraphics[width=0.99\linewidth]{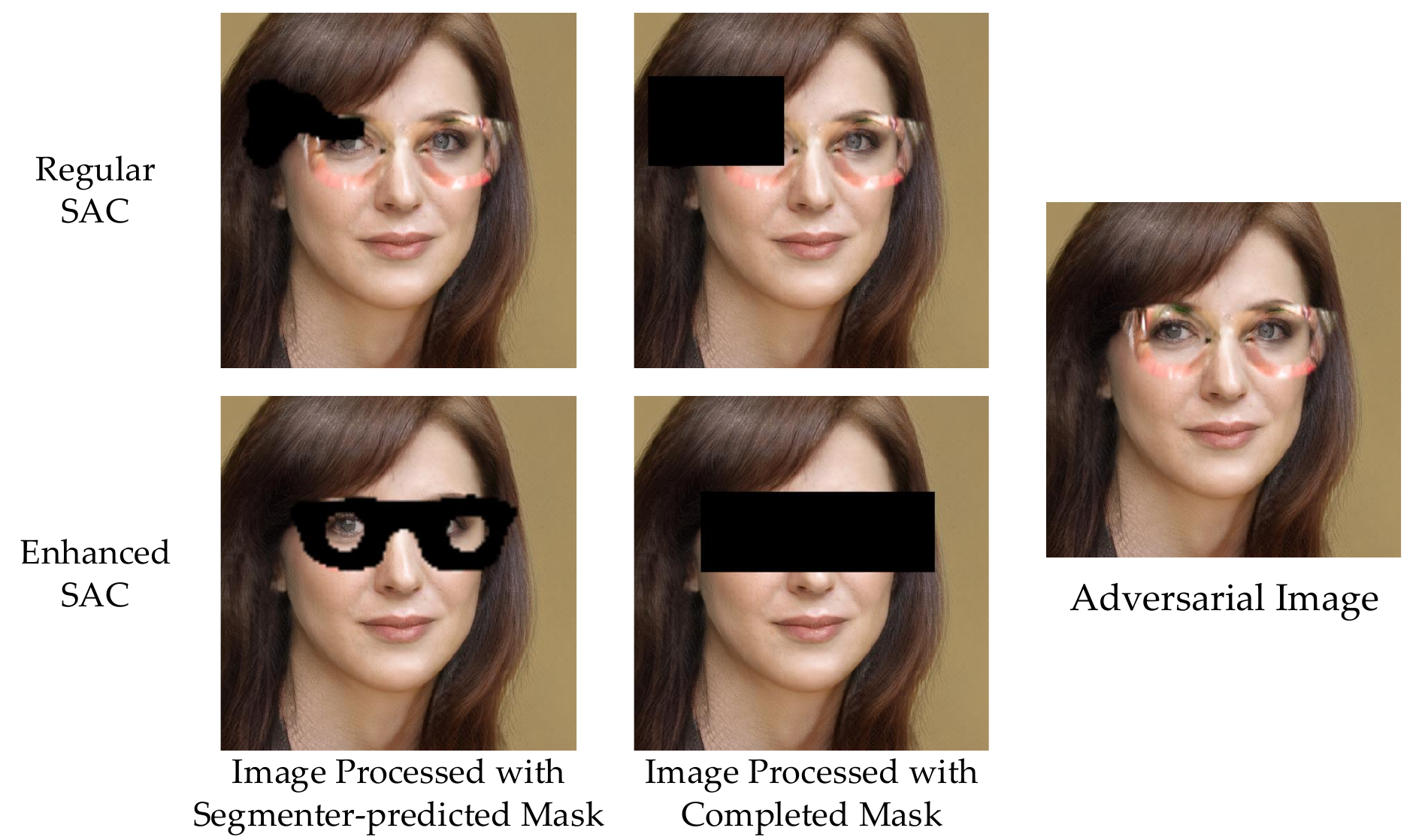}
    \caption{Qualitative results of SAC trained with different data. The first column presents the adversarial image processed by regular SAC which is trained with a patch filled with Gaussian noise, while the subsequent column demonstrates the one processed by enhanced SAC. The adversarial patches are generated with 3DAdv and occupy $8\%$ of the image.}
    \label{fig:vis_sac_demo}
\end{figure}

\noindent \textbf{Qualitative comparison of different versions of SAC.} SAC is a preprocessing-based method that adopts a segmentation model to detect patch areas, followed by a ``shape completion'' technique to extend the predicted area into a larger square, and remove the suspicious area~\cite{liu2022segment}. As shown in Fig.~\ref{fig:vis_sac_demo}, the enhanced SAC, while exhibiting superior segmentation performance in scenarios like face recognition, inadvertently increases the likelihood of masking critical facial features such as eyes and noses. This leads to a reduced ability of the face recognition model to correctly identify individuals, thus impacting its performance in dodging attacks.  

\subsection{{Stable Convergence of Policy Training}}
\label{subsec:convergence}

\begin{figure}[t]
    \centering
 \includegraphics[width=0.99\linewidth]{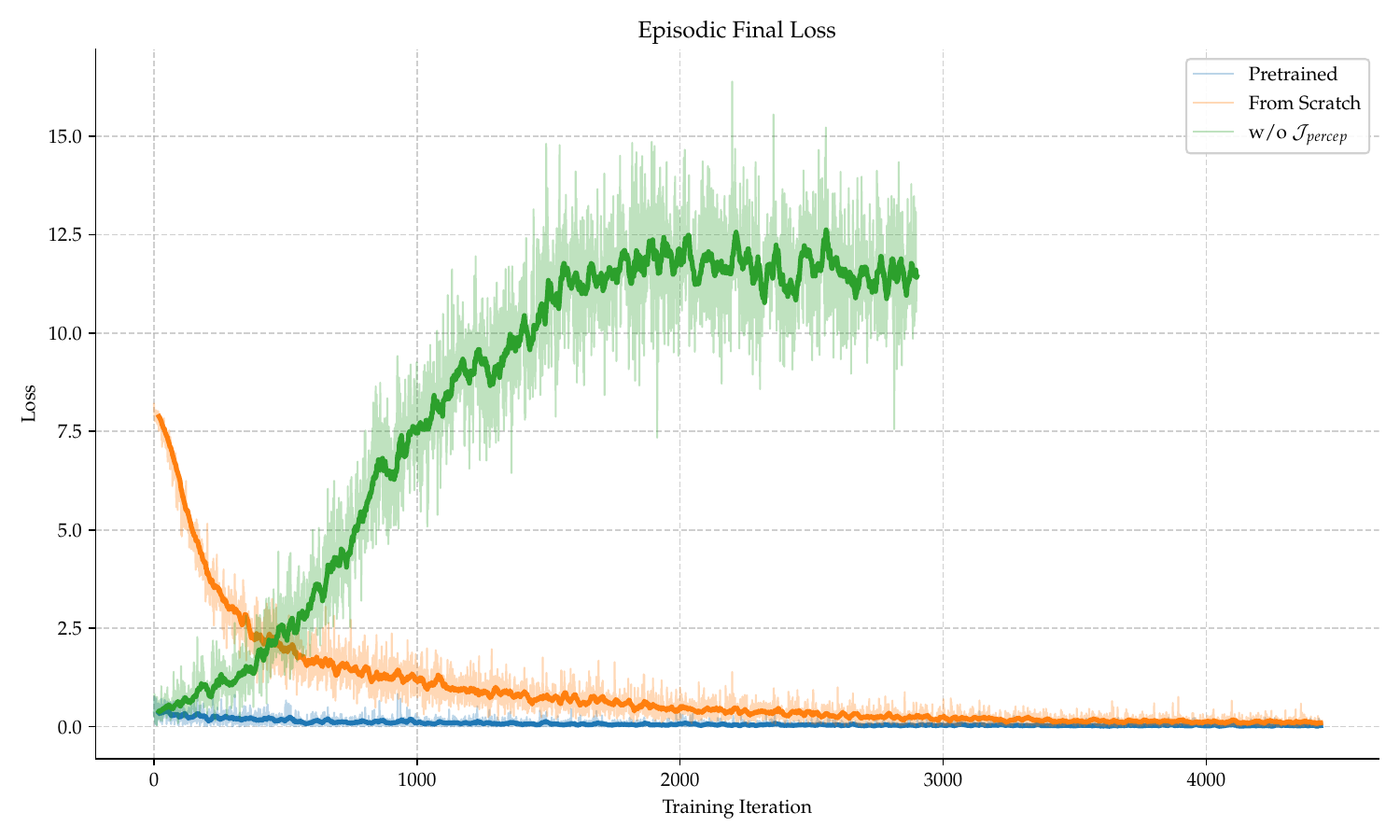}
    \caption{{Episodic Final Loss across training iterations for three model configurations.}}
    \label{fig:convergence}
\end{figure}

{ \textsc{Rein}-EAD has incorporated several key design choices into the standard PPO algorithm for promoting stable convergence during policy training. First, to enhance training stability and efficiency, we avoid training perception and policy models simultaneously from the outset. Early perception models provide unreliable guidance, leading to erratic policy actions and hampering progress. Our solution involves two stages: an initial offline phase where the perception model is trained independently on data from a random policy until stable, followed by an online phase where both networks are trained jointly. This separation enables smoother learning and is significantly more resource-efficient, as the offline pre-training is faster and less computationally demanding than immediate online learning through interaction. As illustrated in Fig.~\ref{fig:convergence}, the pre-trained EAD model ($Pretrained$) demonstrates a better starting point than the model without pre-training ($From Scratch$) and reaches a $10^{-2}$ loss level when trained with the same iterations.
Second, alongside the reinforcement learning objective, we incorporate supervised learning signals derived from ground-truth perceptual annotations. Given that active defense naturally extends passive perception, it is intuitive to retain the original supervised learning objective when training the perception model alongside the policy model. Furthermore, this supervised signal acts as a form of regularization, sustaining the EAD model with a considerable perception capability and preventing excessively large or detrimental changes driven solely by the potentially noisy RL reward signal. It guides the policy towards known effective behaviors, further enhancing stability. As shown by Fig.~\ref{fig:convergence}, the un-regularized curve ( $w/o\;
 J_{percep}$) rapidly increases, fluctuates wildly, and fails to converge.
}

\subsection{Alternative Reward Shaping} 
    
    {\noindent \textbf{Direct Entropy Deduction.} To explicitly encourage directly reducing the uncertainty in predicting the target, we define the reward as the weighted reduction of the entropy term at every step:
    \begin{equation}
    \hat{r}_t
    \;=\;
    \mathcal{H}(\hat{y}_{t-1} \mid b_{t-1}) - \gamma \cdot \mathcal{H}(\hat{y}_{t} \mid b_{t-1}, o_t).
    \label{eq:entropy-deduction}
    \end{equation}
    }

    {\noindent \textbf{Binary Outcome Reward.} As a sparse and unbiased baseline, the binary outcome reward indicates whether the agent finishes the task successfully. Under the case of active perception, we define the binary outcome reward as the predicted probability of the target exceeding the pre-defined threshold. Formally, it's defined as 
    \begin{equation}
    \hat{r}_t
    \; = \; \mathbb{I} (\hat{y}_t = y),
    \label{eq:binary-reward-indicator}
    \end{equation}
    where $\mathbb{I}(\cdot)$ is the indicator function. When the perception model uses probabilistic modeling, which maps from previous belief $b_{t-1}$ and observation $o_t$ to a distribution over target $y$, we regard the episode as successful once the posterior probability for the target exceeds a confidence threshold $\kappa$. 
    The reward is therefore
    \begin{equation}
    \hat{r}_t
    \;=\;
    \mathbb{I} \bigl(f(y \mid b_{t-1}, o_{t}; \vtheta) > \kappa \bigr).
    \label{eq:binary-reward-indicator-2}
    \end{equation}
    In implementation, we select $\kappa = 0.95$ for face recognition during training.
    }

\begin{figure}[t]
    \centering
 \includegraphics[width=0.99\linewidth]{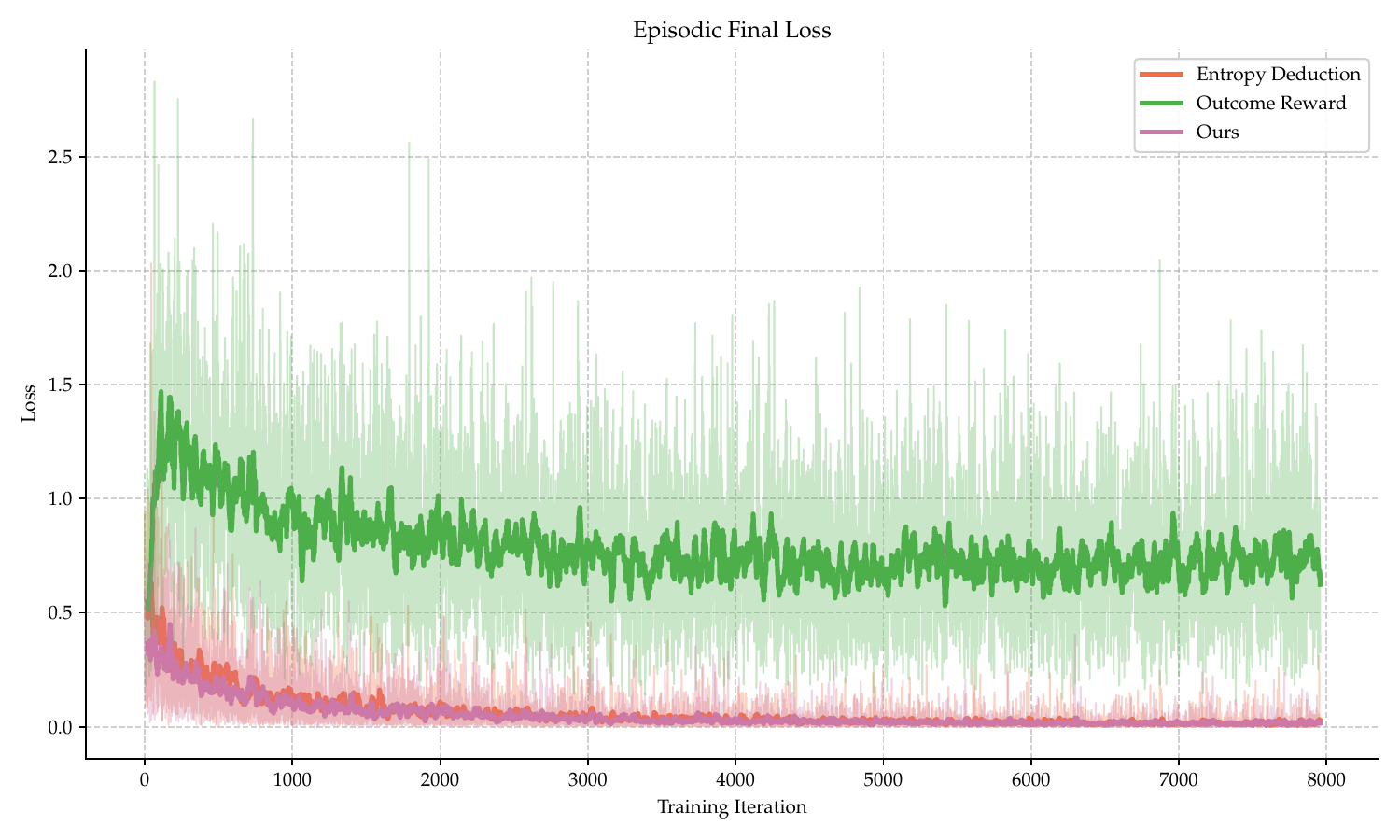}
    \caption{{The loss convergence of REIN-EAD with different
rewards.}}
    \label{fig:rewards}
\end{figure}

    {\noindent\textbf{Evaluation Results.} As shown in Table~\ref{tab:ablation_reward}, our proposed reward shaping approach outperforms other methods in terms of both clean accuracy and adversarial robustness against patches. The Direct Entropy Deduction encourages the policy to select actions that reduce uncertainty; however, as it does not leverage ground-truth labels, it may inadvertently promote confident yet incorrect predictions. Consequently, this method is vulnerable to stronger adversarial attacks, such as 3DAdv, resulting in a higher attack success rate. Meanwhile, the Binary Outcome Reward, due to its sparsity, requires more iterations to converge, leading to comparatively weaker performance given the same number of training iterations. Furthermore, because it lacks uncertainty-informed guidance, its capability for proactive information-seeking to counter adversarial patches is limited. In contrast, our reward shaping employs a dense formulation that accelerates convergence and guides the model to unbiasedly learn a policy that maximizes information gain towards accurate perception. Furthermore, we present the loss convergence behavior of REIN-EAD under different reward designs as shown in Fig.~\ref{fig:rewards}. The Binary Outcome Reward method demonstrates significant instability and slow, suboptimal convergence, ultimately plateauing at a high loss value. In contrast, both the Direct Entropy Deduction method and our proposed reward shaping technique achieve substantially faster and more stable convergence, with the loss for both methods approaching near-zero values within approximately 4,000 training iterations. 
    }

\begin{table}[t]
\caption{{The performance of \textsc{Rein}-EAD with different reward shaping.}} 
\label{tab:ablation_reward}
\setlength\tabcolsep{7.0pt}
\renewcommand\arraystretch{1.25}
\centering
\begin{tabular}{>{\centering\arraybackslash}p{5em}|c|cccc}
\hline
\multirow{2}{*}{{Reward}} &  \multirow{2}{*}{{Acc (\%)}} & \multicolumn{4}{c}{{Attack Success Rate (\%)}}  \\ \cline{3-6} 
                                  &    &  {MIM} & {EoT} & {GenAP} & {3DAdv} \\ \hline \hline
Entropy Deduction        & 88.67 & 3.15 & 2.11 & 4.21 & 11.42  \\ 
Outcome Reward & 88.62 & 3.22 & 3.26 & 5.94 & 10.86  \\ 
\textbf{ours} & \textbf{89.03} & \textbf{2.10} & \textbf{3.15} & \textbf{7.37} & \textbf{4.21}  \\ \hline
\end{tabular}
\vspace{-1em}
\end{table}

\section{Experiment details for object classification}
\label{sec:experiment_detail_oc}
\subsection{Details on Dynamic OmniObject3D} 
\label{sec:dynamic_omni}
The original version of OmniObject3D is accessible at \url{https://omniobject3d.github.io}. We preprocess the dataset since it is raw and does not differentiate any splits for classification tasks. For the dataset following a long tail distribution, we drop categories with less than $10$ instances, which do not contribute much to our experiments, and split the training and test data in each class at a ratio of $4$ to $1$. The final dataset has $176$ classes, with $4409$ objects for training and $1192$ objects for testing. We use Pytorch3D (\url{https://github.com/facebookresearch/pytorch3d}) as the simulation engine since it provides efficient API for batch rendering and a differential pipeline for implementing adversarial attacks on the texture of objects. The meshes of the original OmniObject3D dataset have millions of faces, and the scale of them varies significantly. To lower computational overhead and facilitate batch rendering, we use Pymeshlab (\url{https://pymeshlab.readthedocs.io}) to process the meshes, simplifying the face number to $10,000$ and normalizing the scale without compromising any rendering quality. The image is rendered with a perspective camera of FoV $60$ and resolution $256\times256$ by a hard Phong shader. We use Gym~\cite{brockman2016openai} to warp Pytorch3D as the environment for \textsc{Rein}-EAD.

\subsection{Details for implementation}
\label{sec:oc_imp_detail}
\textbf{Model details.} For the experiment conducted on dynamic OmniObject3D, we implement \textsc{Rein}-EAD for classification with a combination of Swin Transformer and Decision Transformer. We use the pretained Swin-Small Transformer from PyTorch Image Models (\url{https://github.com/huggingface/pytorch-image-models}) and finetune it with a head of 176 classes on the training set of dynamic OmniObject3D. To implement \textsc{Rein}-EAD, We replace the head of the Swin Transformer with a temporal-fusing module and a Decision Transformer. At each step $t>0$, the feature embedding of length $768$ extracted by the Swin-Small backbone is fed to the temporal-fusing module where it concatenates with the previous extracted observation sequence of dimension $(t-1)\times768$ to form a temporal sequence of visual features. This sequence is then temporal-fused by the Decision Transformer, which outputs a refined feature embedding. Finally, a feature decoder, an action decoder and a value decoder implemented by shallow MLP are employed to decode the embedding into refined label, predicted action and value respectively. The output of the action decoder is a predicted view angle. In the training stage, we use it as the mean to sample from a multivariate Gaussian distribution with fixed variance for RL exploration, while in testing, we use it directly as the actual action value.

\noindent\textbf{Training details.} We adopt a similar two-phase training paradigm as \textsc{Rein}-EAD for face recognition. In the offline phase, we freeze the Swin-Small backbone and train the Decision Transformer and feature decoder with a random action policy to collect observations. In the online phase, we employ Algorithm 1 to incorporate the training of the policy network. We approximate the surrogate set of patches with PGD, named OAPA, which attacks the Swin-Small backbone in single view. The learning rate and a number of iterations are $\alpha=8/255$, $N=30$, consistent with DOA$^\dagger$. Note that the training of OAPA for \textsc{Rein}-EAD is offline and only done once, while DOA$^\dagger$ generates a surrogate patch in every iteration. The hyper-parameters of \textsc{Rein}-EAD for object classification are shown in Table~\ref{tab:param_classify}.

\begin{table}[!tbp]
  \renewcommand{\arraystretch}{1.25}
  \centering
  \small
  \caption{Hyper-parameters of \textsc{Rein}-EAD for object classification.}
  \label{tab:param_classify}
  \begin{tabular}{l l}
  \toprule
    \textbf{Hyper-parameter}          & \textbf{Value}              \\ 
    \midrule
    Lower bound for horizontal rotation ($h_{\text{min}}$) & $-90^\circ$ \\
    Lower bound for horizontal rotation ($h_{\text{max}}$) & $90^\circ$ \\
    Lower bound for vertical rotation ($v_{\text{min}}$) & $0^\circ$ \\
    Upper bound for vertical rotation ($v_{\text{max}}$) & $90^\circ$ \\ 
    Ratio of patched data ($r_{\text{patch}}$) & $0.8$ \\
    Training epochs for offline phase ($N_{\text{offline}}$) & $100$ \\
    Learning rate for offline phase ($\mathrm{lr}_{\text{offline}}$) & $2\times10^{-4}$ \\
    Batch size for offline phase ($b_{\text{offline}}$) & $128$ \\
    Total Episodes for online phase  & $150,000$ \\
    Learning rate for online phase ($\mathrm{lr}_{\text{online}}$) & $1\times 10^{-4}$ \\
    Batch size for online phase ($b_{\text{online}}$) & $128$ \\
    Return attenuation factor ($\gamma$) & $0.95$ \\
    Updates per iteration ($n$) & $2$ \\
    \bottomrule
  \end{tabular}%
\end{table}

\subsection{Details for attack}
\label{sec:classify_attack}
\textbf{Attack in texture space for 3D environment.} We implement adversarial attacks to mislead the classifier to output a wrong label. The attack methods on the classification task are similar to those in FR system, but are implemented in texture space. We leverage the traceable rendering computing graph of Pytorch3D to generate the adversarial patch directly on the masked texture of the object, which fits the non-planar surface of the object of diverse categories and ensures multi-view consistency. Since the shape of objects varies significantly in the context of the classification task, to ensure that the patch can be fully attached to the object, we set the patch size to $20\%$ of the bounding box of the object and locate it on the center of the bounding box. The adversarial sample for the robustness test is generated on test datasets that are unseen for both \textsc{Rein}-EAD and other defense baselines.

\noindent \textbf{White-box Attack.} We use MIM \cite{dong2018boosting} as a single-view white box adversary, which is enhanced by momentum item. The decay factor of MIM is set at $\mu=1.0$, consistent with FR attack. The multi-view threats are established by EoT\cite{athalye2018synthesizing} and MeshAdv\cite{xiao2019meshadv}. EoT adopts a 2D batch data augmentation that includes shifting, rotating and flipping the rendered image and averages the gradient in image space. In contrast, MeshAdv employs a batch of 3D transformations across the action range of \textsc{Rein}-EAD and backpropagates the gradient through the differential rendering pipeline to texture space, making it more robust to view change. For the white-box attacks, we set the learning rate and the number of iterations at $N=100$ and $\alpha = 8/255$. The sampling frequency for EoT and MeshAdv is established at $M = 128$. 

\noindent \textbf{Black-box Attack.} For query-based attack, we use RGF \cite{ghadimi2013stochastic} and N attack \cite{li2019nattack}. We set the maximum number of queries at $N=10000$ and sampling frequency at $M = 100$. The learning rates for RGF and N attack are $\alpha = 0.05$ and $\alpha = 0.1$ respectively. For transfer-based adversaries, We utilize MeshAdv with parameters the same as the white box version and fine-tune a Swin-Tiny Transformer as the surrogate model to launch the transfer attack. 

\noindent \textbf{Adaptive attack.} Following the face recognition setting, we adaptively attack JPEG and LGS with BPDA~\cite{athalye2018obfuscated} technique, SAC$^\dagger$ and PZ$^\dagger$ with STE~\cite{bengio2013estimating} technique. For \textsc{Rein}-EAD, we launch the adaptive attack with a uniform superset policy which is provided to be most effective in the FR task. 

\subsection{Details for defense}
\label{sec:oc_defense_detail}
The implementation of JPEG\cite{dziugaite2016study} and LGS\cite{naseer2019local} are consistent with the FR task. For SAC$^\dagger$\cite{liu2022segment} and PZ$^\dagger$\cite{xu2023patchzero}, we retrain the patch segmenter on the training set of dynamic OmniObject3D with adversarial patches optimized by EoT\cite{athalye2018synthesizing} using the same code with FR task. For DOA$^\dagger$\cite{wu2019defending}, we follow its training paradigm to fine-tune the same Swin-Small Transformer backbone used by \textsc{Rein}-EAD on the training set of dynamic OmniObject3D. Specifically, we utilize PGD with learning rate $\alpha=8/255$ and number of iterations $N=30$ for adversarial training and search the patch location using the gradient-based method with the top candidate number $C=10$ as described in their paper. The patches used for training SAC$^\dagger$, PZ$^\dagger$ and DOA$^\dagger$ occupy $20\%$ of the bounding box of the object, which is the same with \textsc{Rein}-EAD.

\section{Experiment details for object detection}
\label{sec:experiment_detail_od}

\subsection{Details on EG3D} 
\label{sec:detail_eg3d}
For object detection, we use a pre-trained EG3D model on ShapeNet Cars at \url{https://catalog.ngc.nvidia.com/orgs/nvidia/teams/research/models/eg3d} to generate multi-view car images with the resolution of $256\times256$. We generate 1000 cars with different appearances and split the data into $800$ training data and $200$ test data. The latent seeds are recorded to ensure that the identity of each car remains consistent in all the experiments. Since the background of the generated image is blank, we are able to annotate the bounding box automatically. The online environment for \textsc{Rein}-EAD is wrapped by Gym~\cite{brockman2016openai}.

\subsection{Details on CARLA}
\label{sec:detail_carla}
We utilize CARLA 0.9.14~\cite{dosovitskiy2017carla} for a more complex object detection experiment. The training data covers all 41 different vehicle blueprints provided by CARLA. For each blueprint we generate vehicles of different color versions and collect multi-view samples in the different backgrounds on Town10 as the offline datasets for training model and adversarial samples. The training and testing procedures for \textsc{Rein}-EAD are conducted online in CARLA, which allows the agent to explore every possible view within the action range. During the test experiment, we use all 41 vehicles and place them in locations different from the training set. We use the Python API provided by CARLA to automatically collect the data at a resolution of $256\times444$ and annotate the label. We warp CARLA with Gym~\cite{brockman2016openai} as an online environment for \textsc{Rein}-EAD.

\subsection{Details of implementations}
\label{sec:detail_det_implement}
\textbf{Model details.} For the object detection task on EG3D, we implement \textsc{Rein}-EAD with a combination of YOLOv5n and Decision Transformer. We use the pretained YOLOv5n from the official implementation (\url{https://github.com/ultralytics/yolov5}) and fine-tune it as a single class detection model on the training set of each environment respectively. For each time step $t > 0$, given the current observation of dimensions $256\times256\times3$ as input, the feature maps of dimension $32\times32\times64$ output by the second Cross Stage Partial Networks are utilized. We find feature maps at this level to be computationally efficient for \textsc{Rein}-EAD as the later stages of YOLOv5n form a large-scale concatenated feature pyramid. These maps are then reshaped into a sequence with dimensions $64\times1024$. To concatenate it with the previous extracted observation sequence $(t-1) \times 64 \times 1024$, we have a temporal sequence of visual features as the input of Decision Transformer, and it outputs the temporal-fused visual feature sequence $64\times1024$ and predicted action. For \textsc{Rein}-EAD, an extra value decoder is utilized to estimate the advantage value. To predict the bounding boxes and objectness score which is required in object detection, we reshape the temporal-fused visual sequence back to its original shape and utilize it as input for the later stage in YOLOv5n.

\noindent\textbf{Training details. } We adopt a similar training paradigm as \textsc{Rein}-EAD for previous tasks and set the
hyper-parameters of \textsc{Rein}-EAD for object detection as Table~\ref{tab:param_det_EG3D} and Table~\ref{tab:param_det_CARLA}

\begin{table}[t]
  \renewcommand{\arraystretch}{1.25}
  \centering
  \small
  \caption{Hyper-parameters of \textsc{Rein}-EAD for object detection on EG3D.}
  \label{tab:param_det_EG3D}
  \begin{tabular}{l l}
  \toprule
    \textbf{Hyper-parameter}          & \textbf{Value}              \\ 
    \midrule
    Lower bound for horizontal rotation ($h_{\text{min}}$) & $-60^\circ$ \\
    Upper bound for horizontal rotation ($h_{\text{max}})$ & $60^\circ$ \\
    Lower bound for vertical rotation ($v_{\text{min}}$) & $0^\circ$ \\
    Upper bound for vertical rotation ($v_{\text{max}}$) & $30^\circ$ \\
    Ratio of patched data ($r_{\text{patch}}$) & $0.4$ \\
    Training epochs for offline phase ($N_{\text{offline}}$) & $50$ \\
    Learning rate for offline phase ($\mathrm{lr}_{\text{offline}}$) & 
    $2\times10^{-4}$ \\
    Batch size for offline phase ($b_{\text{offline}}$) & $128$ \\
    Total episodes for online phase ($N_{\text{online}}$) & $10,000$ \\
    Learning rate for online phase ($\mathrm{lr}_{\text{online}}$) & $1\times 10^{-4}$ \\
    Batch size for online phase ($b_{\text{online}}$) & $64$ \\
    Return attenuation factor ($\gamma$) & $0.95$ \\
    Updates per iteration ($n$) & $2$ \\
    \bottomrule
  \end{tabular}%
\end{table}

\begin{table}[t]
  \renewcommand{\arraystretch}{1.25}
  \centering
  \small
  \caption{Hyper-parameters of \textsc{Rein}-EAD for object detection on CARLA.}
  \label{tab:param_det_CARLA}
  \begin{tabular}{l l}
  \toprule
    \textbf{Hyper-parameter}          & \textbf{Value}              \\ 
    \midrule
    Lower bound for horizontal rotation ($h_{\text{min}}$) & $-60^\circ$ \\
    Upper bound for horizontal rotation ($h_{\text{max}})$ & $60^\circ$ \\
    Lower bound for vertical rotation ($v_{\text{min}}$) & $5^\circ$ \\
    Upper bound for vertical rotation ($v_{\text{max}}$) & $35^\circ$ \\ 
    Ratio of patched data ($r_{\text{patch}}$) & $0.4$ \\
    Training epochs for offline phase ($\mathrm{lr}_{\text{offline}}$) & $50$ \\
    Learning rate for offline phase ($\mathrm{lr}_{\text{offline}}$) & $2\times10^{-4}$ \\
    Batch size for offline phase ($b_{\text{offline}}$) & $128$ \\
    Total Episodes for online phase  & $10,000$ \\
    Learning rate for online phase ($\mathrm{lr}_{\text{online}}$) & $1\times 10^{-4}$ \\
    Batch size for online phase ($b_{\text{online}}$) & $64$ \\
    Return attenuation factor ($\gamma$) & $0.95$ \\
    Updates per iteration ($n$) & $2$ \\
    \bottomrule
  \end{tabular}%
\end{table}

\begin{figure*}
    \centering
    \includegraphics[width=0.99\linewidth]{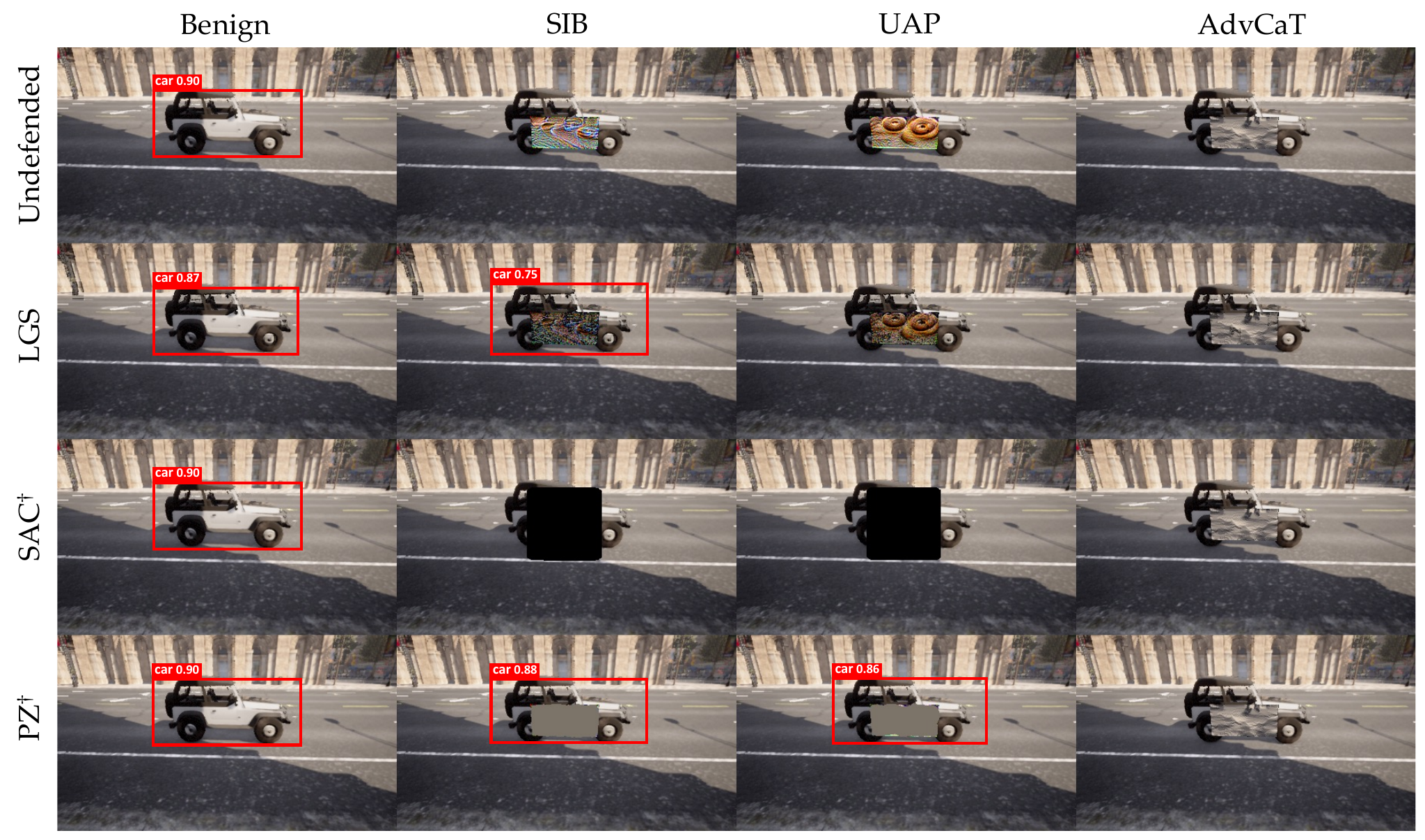}
    \caption{Visualization of defense baseline on CARLA. $^\dagger$ denotes methods are trained with adversarial examples.}
    \label{fig:det_baseline}
\end{figure*}

\subsection{Details for attack}
\textbf{Multi-view hiding attack for vehicle object detection.} 
In the context of a single-class vehicle object detection task, the goal of the adversary is to place a patch on a vehicle and make it disappear from the object detector. The patch is placed on the side of the vehicle and occupies $25\%$ of the bounding box. We follow the adversarial loss designed for YOLO~\cite{thys2019fooling} to minimize the objectness score to achieve a hiding attack. To enhance the multi-view robustness, we train adversarial patches with a batch of images with different views in each iteration. The sampling frequency is set at $100$. 

\noindent \textbf{Attack in pixel space.} EoT is used as a baseline multi-view adversary, which incorporates the expectation of view transformation described above. To achieve a more generalized attack, we utilize Universal Adversarial Perturbations (UAP)\cite{moosavi2017uap} to generate a single adversarial patch for all the vehicles in the dataset, which is able to hide any vehicles from the detector. The learning rate and the number of iterations for both methods are set at $N=500$ and $\alpha = 8/255$ . 

\noindent \textbf{Attack in the hidden layer.} As the EAD module is plugged at the middle stage of YOLOv5n, we implement SIB \cite{zhao2019seeing} , which attacks the feature in the hidden layer. We set the feature to be perturbed as the input of the EAD module and maximize the difference between the adversarial feature and the original feature with their feature-interference reinforcement loss. The coefficients for objectness loss and feature loss are $\mu_{1}=0.5$ and $\mu_{2}=0.5$ respectively. The learning rate and the number of iterations are consistent with EoT and UAP.

\noindent \textbf{Attack in latent space.} To achieve an inconspicuous patch attack for both human and detector, we utilized adversarial camouflage textures (AdvCaT)~\cite{hu2023advcat}, which adopts Voronoi diagram and Gumbel-softmax trick to generate the semantic polygon camouflage with control points and latent seeds. We use K-means with 4 clusters to extract the base colors for camouflage from the environment. We set the learning rate for control points and latent seeds at $\alpha_{1}=0.0005$ and $\alpha_{2}=0.005$ respectively. The number of iterations is $N=200$. 

\begin{figure}[htp]
    \centering
    \includegraphics[width=0.99\linewidth]{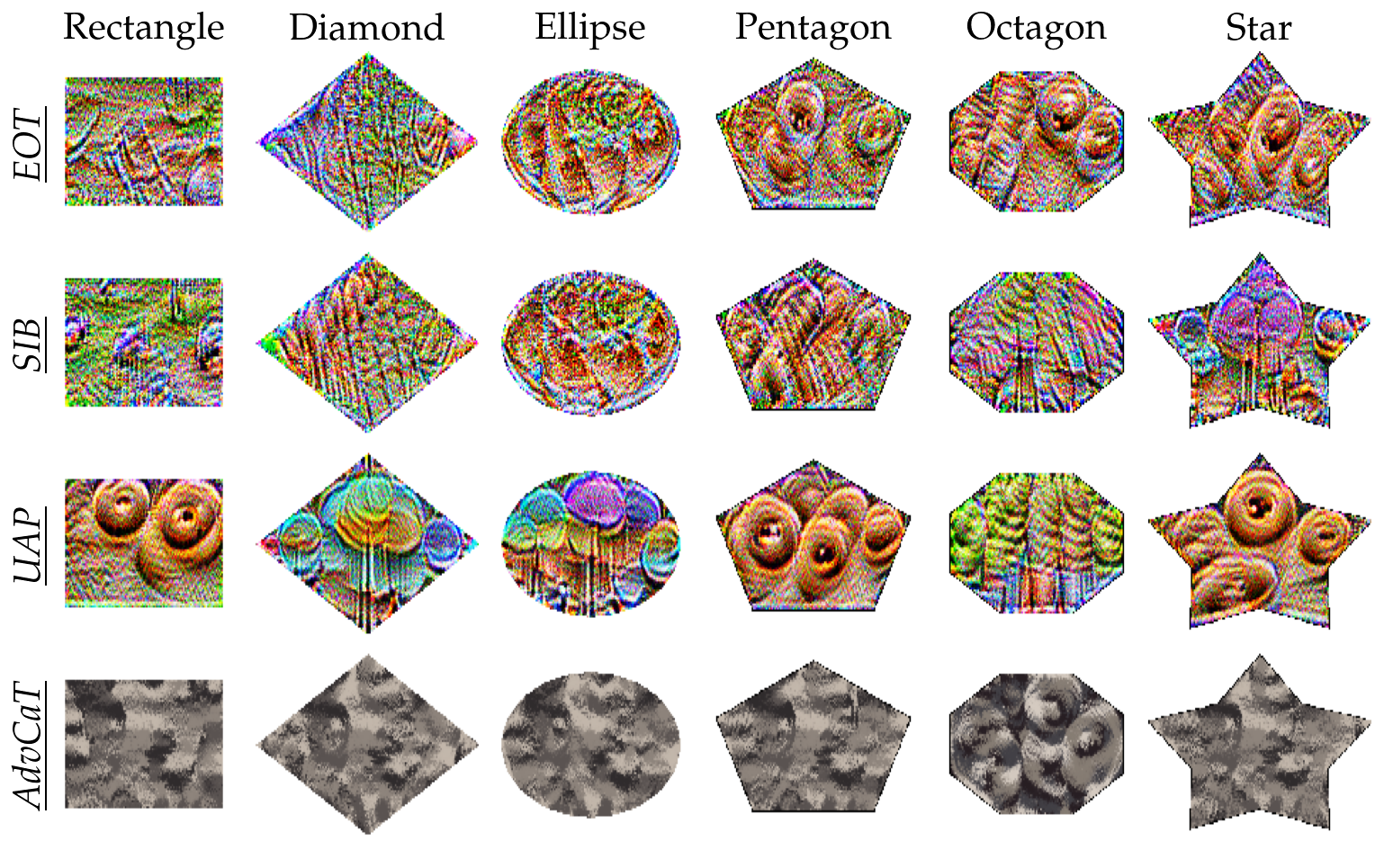}  
    \caption{Visualization of the generated patch under different attacks and shapes on CARLA. }
    \label{fig:det_patch_shape}
\end{figure}

\subsection{Details for defense}
\label{sec:det_defense_detail}
The implementation of JPEG~\cite{dziugaite2016study} and LGS~\cite{naseer2019local} are consistent with the FR task. For SAC$^\dagger$~\cite{liu2022segment} and PZ$^\dagger$~\cite{xu2023patchzero}, we retrain the patch segmenter on the training set of the corresponding environment with adversarial patches optimized by EoT using the same code with FR task. The patches used for training SAC$^\dagger$ and PZ$^\dagger$ occupy $25\%$ of the bounding box of the vehicle, which is the same as those used for training \textsc{Rein}-EAD. 

\subsection{Qualitative comparison of defense baselines in CARLA.}
\label{sec:det_qualitative_baseline}
We present the qualitative defense result for baseline methods in Fig.~\ref{fig:det_baseline}. The smoothing mechanism of LGS is relatively weak and unable to completely eliminate adversarial noise. In contrast, the segmentation backbone of SAC$^\dagger$ and PZ$^\dagger$ can perfectly distinguish and remove the patch attacks with noisy patterns, but fail to detect the environmental mosaic AdvCaT. Additionally, the completion mechanism of SAC$^\dagger$ leads to excessive occlusion, which complicates subsequent detection tasks.

\subsection{Visualization of patches on CARLA}
\label{sec:vis_patch_shapes}
We generate patches with the shapes of diamond, ellipse, pentagon, octagon and star. They are unseen for both \textsc{Rein}-EAD and other defense baselines. A visualization of these patches are shown in Fig.~\ref{fig:det_patch_shape}.

\subsection{More results on patch shapes}
\label{sec:more_patch_shapes}
\begin{figure}[htp]
    \centering
    \begin{minipage}{0.5\linewidth}
        \centering
        \includegraphics[width=\linewidth]{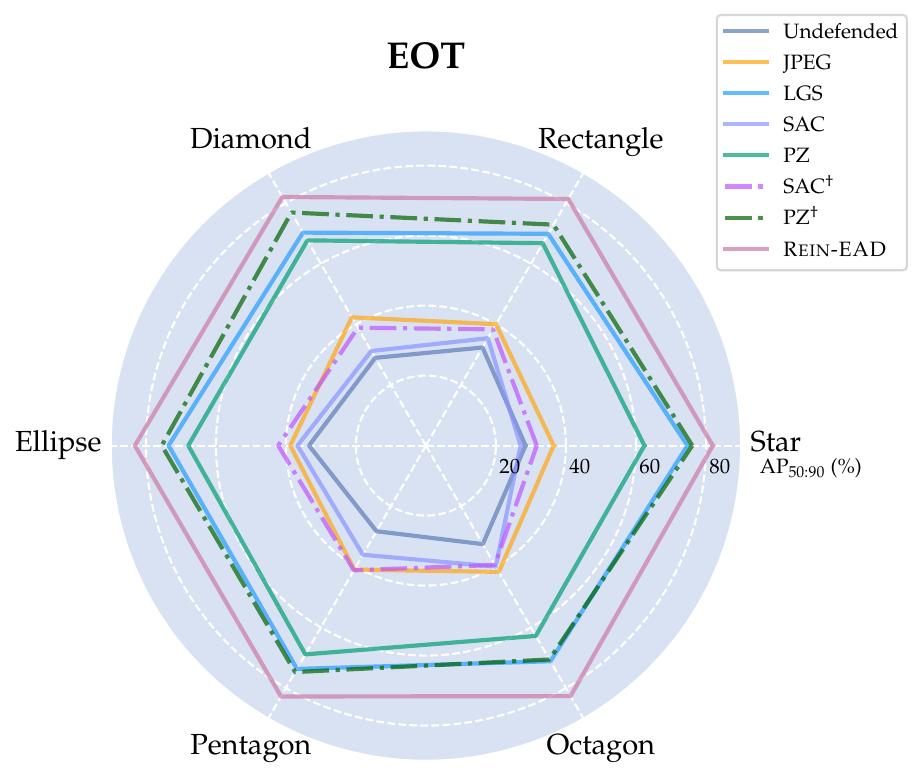}
    \end{minipage}%
    \begin{minipage}{0.5\linewidth}
        \centering
        \includegraphics[width=\linewidth]{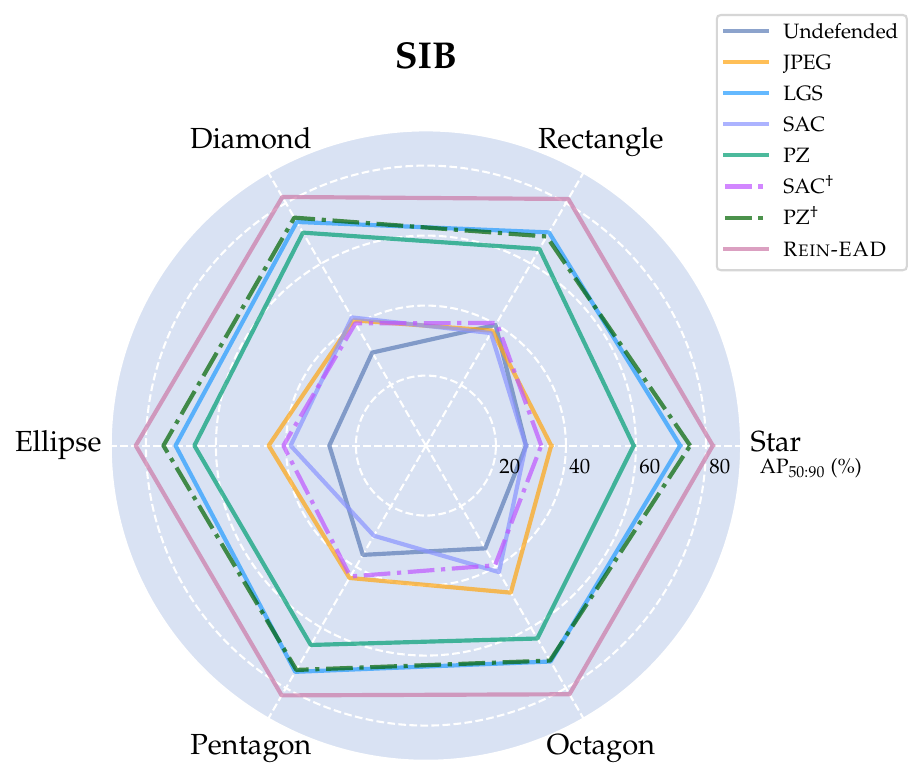}
    \end{minipage}
\caption{Comparative evaluation of object detection defense methods under EoT and SIB attack with different patch shapes on CARLA.}
\label{fig:more_lidar_det}
\end{figure}

More comparative evaluations for EoT and SIB attacks with different patch shapes are illustrated in Fig.~\ref{fig:more_lidar_det}.

\begin{figure*}[htp]
    \centering
    \includegraphics[width=0.99\linewidth]{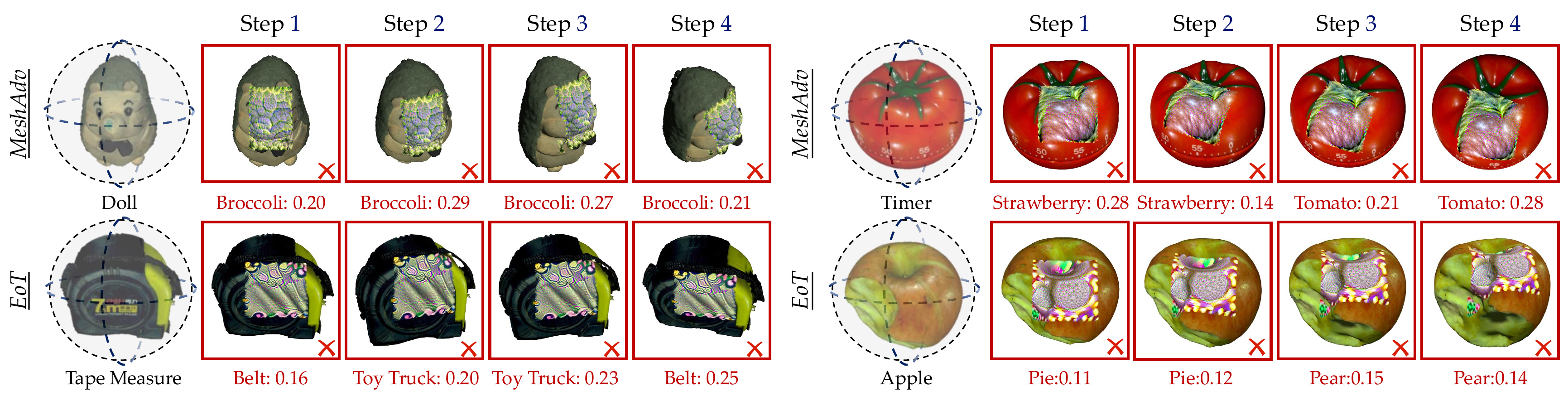}  
\caption{{Visualization of failure cases by adopting REIN-EAD on dynamic OmniObject3D, with the adversarial patch occupying 20\% of the object’s bounding box in the front view.}}
\label{fig:ead_fail}
\end{figure*}
\subsection{Failure cases}
{The failure cases reveal several notable limitations of REIN-EAD. First, the system is vulnerable to strategically positioned adversarial patches that occlude critical object features. As illustrated in Fig.~\ref{fig:ead_fail}, when an adversarial patch obscures the facial region of a green doll, the model consistently misclassifies the object as broccoli. This suggests that the occlusion of key discriminative features compromises the model's ability to acquire adequate information for accurate recognition. Second, the framework exhibits degraded performance when simultaneously subjected to adversarial attacks and natural out-of-distribution interference. The multi-step exploration mechanism, while generally effective, fails to resolve conflicting signals in such compound uncertainty scenarios. A representative example, shown in Fig.~\ref{fig:ead_fail}, demonstrates persistent prediction uncertainty when processing a partially consumed apple, where bite-induced shape deformation interacts with adversarial perturbations. These limitations highlight potential areas for future research. Potential avenues include developing patch placement strategies that prioritize important visual areas and incorporating more diverse 3D object datasets for training to enhance the model's robustness to occlusion and out-of-distribution samples.}

\end{document}

%% file: math_commands.tex
\usepackage{amsmath,amsfonts,bm}

\def\eqref#1{equation~\ref{#1}}

\def\1{\bm{1}}

\def\vtheta{{\bm{\theta}}}
\def\vphi{{\bm{\phi}}}
\def\va{{\bm{a}}}

\def\mE{{\bm{E}}}

\def\mI{{\bm{I}}}

\def\mK{{\bm{K}}}

\def\mM{{\bm{M}}}

\def\mR{{\bm{R}}}

\def\mT{{\bm{T}}}

\DeclareMathAlphabet{\mathsfit}{\encodingdefault}{\sfdefault}{m}{sl}
\SetMathAlphabet{\mathsfit}{bold}{\encodingdefault}{\sfdefault}{bx}{n}

\def\gP{{\mathcal{P}}}

\newcommand{\KL}{D_{\mathrm{KL}}}

\DeclareMathOperator*{\argmax}{arg\,max}

\usepackage{xspace}
\makeatletter
\DeclareRobustCommand\onedot{\futurelet\@let@token\@onedot}
\def\@onedot{\ifx\@let@token.\else.\null\fi\xspace}

\def\wrt{\emph{w.r.t}\onedot}

\def\etal{\emph{et al}\onedot}